\documentclass[10pt,twocolumn,letterpaper]{article}

\usepackage[pagenumbers]{iccv} %

\definecolor{vblue}{rgb}{0.21,0.49,0.74}
\usepackage[pagebackref,breaklinks,colorlinks,allcolors=vblue]{hyperref}
\colorlet{lgreen}{green!10}
\colorlet{lblue}{blue!10}
\colorlet{lred}{red!10}

\usepackage{algorithm}
\usepackage{algorithmic}
\newtheorem{remark}{Remark}

\usepackage{tikz}
\NewDocumentCommand{\progressbar}{m O{2} O{blue!70}}{%
  \begin{tikzpicture}
    \fill[gray!30] (0,0) rectangle (#2,0.3); %
    \fill[#3] (0,0) rectangle ({#1/100*#2},0.3); %
  \end{tikzpicture}%
}

\def\paperID{xxxx} %
\def\confName{xxxx}
\def\confYear{xxxx}

\title{LatentLLM: Attention-Aware Joint Tensor Compression}

\author{ Toshiaki Koike-Akino, Xiangyu Chen, Jing Liu, Ye Wang, Pu (Perry) Wang, Matthew Brand   \\
{\small  Mitsubishi Electric Research Laboratories (MERL), Cambridge, MA 02139, USA}\\
{\tt\small \{koike, xiachen, jiliu, yewang, pwang, brand \}@merl.com}
}

\begin{document}
\maketitle
\begin{abstract}
    Modern foundation models such as large language models (LLMs) and large multi-modal models (LMMs) require a massive amount of computational and memory resources.
We propose a new framework to convert such LLMs/LMMs into a reduced-dimension latent structure.
Our method extends a local activation-aware tensor decomposition to a global attention-aware joint tensor decomposition.
Our framework can significantly improve the model accuracy over the existing model compression methods when reducing the latent dimension to realize computationally/memory-efficient LLMs/LLMs.
We show the benefit on several benchmark including multi-modal reasoning tasks.

\end{abstract}

\section{Introduction}
\label{intro}

Large language models (LLMs)~\cite{touvron2023llama, achiam2023gpt} and large multi-modal models (LMMs)~\cite{liu2023llava} have shown excellent performance across a variety
of general tasks~\cite{wei2022emergent,katz2024gpt, bubeck2023sparks}. 
Nonetheless, these models having billions of parameters demand significant computational resources~\cite{schwartz2020green}. 
Towards increasing the accessibility and sustainability of LLMs/LMMs, extensive efforts have been devoted to model compression~\cite{xu2023survey, zhu2024survey, bai2024beyond}: e.g., partial activation~\cite{jiang2024mixtral, lin2024moe}, pruning~\cite{frantar2023sparsegpt, sun2023simple, bai2024sparsellm, hassibi1993optimal},
quantization~\cite{frantar2022gptq, lin2024awq, wang2024q}, knowledge distillation~\cite{hsieh2023distilling, deepseekai2025deepseekr1incentivizingreasoningcapability, hwang2024pc}, and low-rank factorization~\cite{yuan2023asvd, liu2024deepseek, hwang2024pc, saxena2024eigen}. 

More recently, the reduced-dimension LLM DeepSeek-V3~\cite{liu2024deepseek} has attracted much attention for its high efficiency and performance.
It employs a low-rank architecture called multi-head latent attention (MLA) to compress the standard multi-head attention (MHA), realizing an efficient KV cache~\cite{chang2024palu, saxena2024eigen}, accelerated training, and high-performance inference.
In this paper, we provide a novel solution to convert a pretrained LLM/LMM built with MHA into a compressed LLM/LMM with a type of MLA.
Our approach is motivated by a global compression framework introduced in SparseLLM~\cite{bai2024sparsellm} and Q-VLM~\cite{wang2024q}.
Although the original method was designed for pruning/quantization, we adopt it for tensor rank reduction.
We further extended it to the joint compression of MHA, while the original SparseLLM was for compressing the multi-layer perceptron (MLP) part.
Our derived solution is based on a high-order tensor-rank decomposition to jointly factorize multiple linear layers.

The contributions of our paper are summarized below.
\begin{itemize}

    \item We propose a novel low-rank decomposition approach called LatentLLM to compress LLMs/LMMs.

    \item We discuss an optimal pre-conditioning for activation-aware SVD.

    \item We reveal that a choice of junction matrix can significantly reduce the model size.

    \item We then introduce an attention-aware joint SVD framework to compress multiple weights at the same time.

    \item Our experiments validate that our LatentLLM approach can improve the performance of LLM/LMM compression over existing methods.

    \item The latent LLaVa with our method offers a significant advantage in multi-modal reasoning capability.
\end{itemize}

\section{Related work}

\paragraph{Model compression}
The field of model compression for LLMs/LMMs has seen a surge of innovative techniques aimed at mitigating the substantial computation and memory requirements~\cite{zhu2024survey, yuan2024llm}. 
Various methods have emerged to address this challenge, each taking a unique approach to reduce the memory footprint
of LLMs/LMMs. 
These methods primarily fall into four categories: weight quantization~\cite{lin2024awq, frantar2022gptq, wang2024q}, network pruning~\cite{lecun1989optimal, hassibi1993optimal, frantar2023sparsegpt, bai2024sparsellm},
knowledge distillation~\cite{hsieh2023distilling, deepseekai2025deepseekr1incentivizingreasoningcapability, hwang2024pc}, and low-rank factorization~\cite{yuan2023asvd, liu2024deepseek, hwang2024pc, saxena2024eigen, saha2024compressing}. 

Among these methods, weight quantization has gained significant traction in the context of large foundation models due to its effectiveness. 
However, all four compression techniques are orthogonal and can be applied together. 
Recognizing this, we introduce a novel low-rank decomposition method which jointly compresses multiple layers of an LLM/LMM in a training-free manner.

\paragraph{Low-rank decomposition}

In the realm of low-rank decomposition~\cite{schotthofer2022low} for
neural network compression, existing methods typically involve decomposing weight
matrices of pre-trained networks using techniques like Singular Value Decomposition (SVD) or tensor decomposition, followed by fine-tuning the factorized network \cite{denton2014exploiting, sainath2013low}. 
LoSparse~\cite{li2023losparse} uses low-rank approximation plus a sparse matrix to compress the weight matrix in transformers.
Similarly, CALDERA~\cite{saha2024compressing} uses low-rank approximation plus a quantized matrix.
ASVD~\cite{yuan2023asvd} significantly improves the low-rank decomposition by dealing with activation statistics.
It was applied to SVD-LLM~\cite{wang2024svd} and Palu~\cite{chang2024palu}.
DeepSeek-V3~\cite{liu2024deepseek} employs the similar latent reduction via MLA to make MHA efficient and capable.
Eigen attention~\cite{saxena2024eigen} is highly related to MLA.

\section{LatentLLM: Tensor compression}

\subsection{Reduced-dimension LLM/LMM}

\begin{figure}[tb]
\begin{center}
\centerline{
\includegraphics[width=\linewidth]{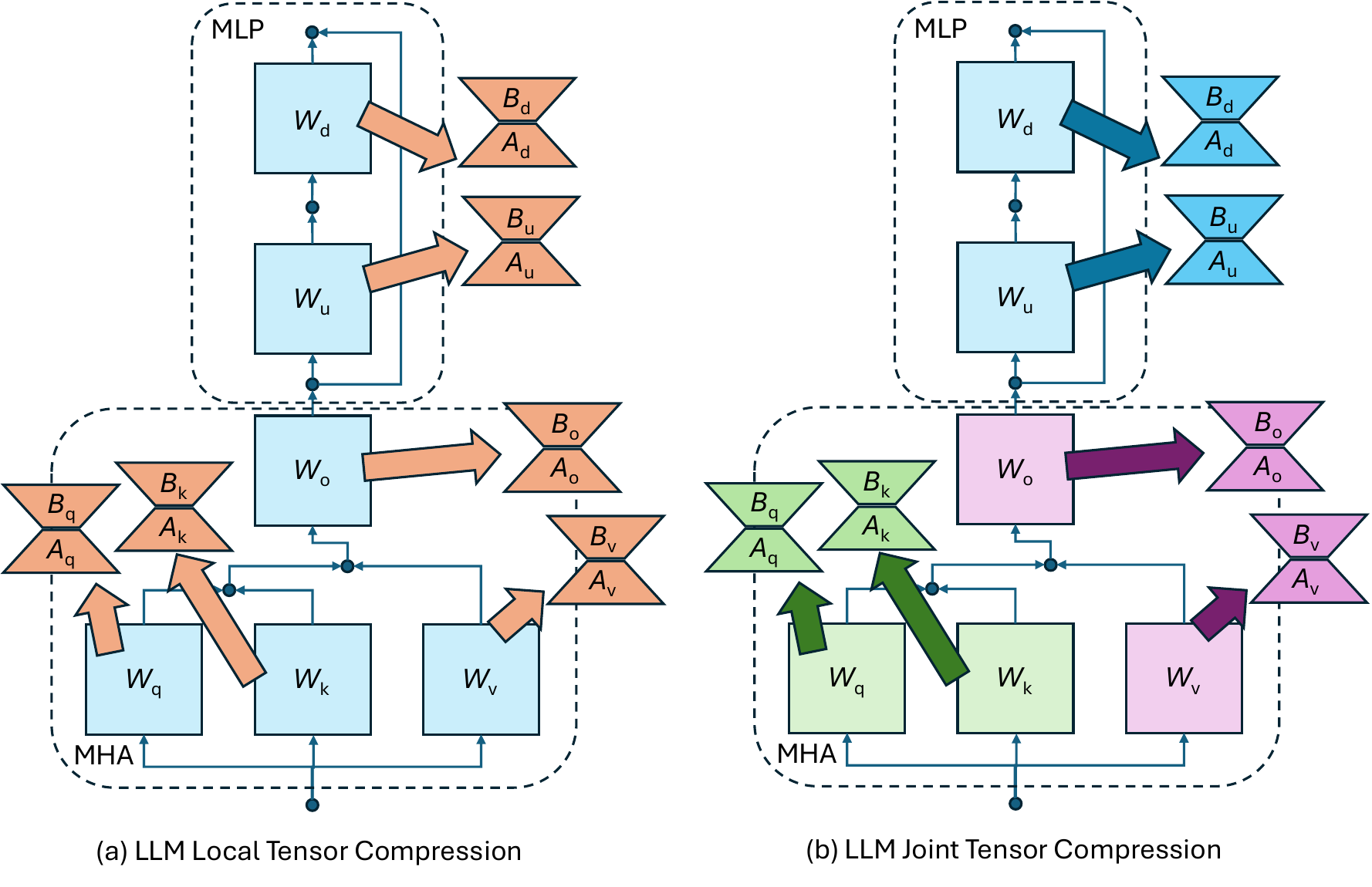}
}

\caption{Reduced-dimension LLM/LMM with low-rank tensor decomposition.
(a) each linear modules are locally compressed by activation-aware tensor decomposition.
(b) multiple linear modules are globally compressed by attention-aware tensor decomposition.
}
\label{fig:llm}
\end{center}
\vskip -0.2in
\end{figure}

\cref{fig:llm} illustrates the basic transformer architecture consisting of MHA and MLP, used in some LLMs/LMMs. 
For MLP, there are up and down projections, whereas MHA has query/key/value/output projections.
By transforming those dense weight matrices into low-rank decompositions, we can realize an efficient latent LLM/LMM having potential benefits: (i) fewer-parameter model size; (ii) KV cache reduction; (iii) accelerated processing; (iv) lower-power consumption.
In fact, some recent LLM models such as DeepSeek-V3~\cite{liu2024deepseek} demonstrated efficiency and high-performance with MLA. 
We focus on compressing a pre-trained LLM/LMM by converting MHA into a type of MLA in a zero-shot fashion, i.e., without any fine-tuning.

Most existing compression methods are based on a local loss minimization to approximate each weight individually.
Motivated by recent work towards global optimization with SparseLLM~\cite{bai2024sparsellm} and Q-VLM~\cite{wang2024q}, we propose a joint tensor compression framework that we call ``LatentLLM.''
Specifically, we derive a mathematical solution to jointly decompose a pair of query and value projections, a pair of value and output projections, and a pair of up and down projections to compress LLMs/LMMs. 

Before describing our solution, we first address activation-aware compression.
We provide some new insights on the choice of pre-conditioner and junction matrix below.

\subsection{Activation-aware SVD: Pre-conditioning}

\begin{table*}[t]
\centering
\caption{Variants of pre-conditioning matrices $P$ for activation-aware distillation.}
\label{tab:cond}
\small
\begin{tabular}{lll}
  \toprule
  Conditioning $P$ & Expression & Reference \\
  \midrule
  Identity & $I$ & Plain SVD~\cite{sainath2013low, denton2014exploiting} 
  \\
  Diagonal Hessian & 
  $\mathsf{diag}[(XX^\top + \lambda I)^{-1}]^\frac{-1}{2}$
  &
  OBS~\cite{hassibi1993optimal}; GPTQ~\cite{frantar2022gptq}; SparseGPT~\cite{frantar2023sparsegpt}
  \\
  Diagonal $\ell_1$-norm & 
  $\mathsf{diag}
  \big[\sum_j |X_{1,j}|,\ldots, \sum_j |X_{d,j}|
  \big]^\alpha$
  &
  ASVD~\cite{yuan2023asvd}; AWQ~\cite{lin2024awq}
  \\
  Diagonal $\ell_2$-norm & $\mathsf{diag}[XX^\top]^\frac{1}{2}$
  & WandA~\cite{sun2023simple}
  \\
  Covariance & $XX^\top + \lambda I$ & 
  CorDA~\cite{yang2024corda}
  \\
  Root-Covariance & $(XX^\top+\lambda I)^\frac{1}{2}$
  &
  LatentLLM (Ours)
  \\
  \bottomrule
\end{tabular}
\end{table*}

A pioneering work by ASVD~\cite{yuan2023asvd} introduced a way to compress a layer depending on the activation statistics.
Consider a pretrained-weight $W\in\mathbb{R}^{d'\times d}$ to compress with a lower-rank decomposition $\hat{W}=BA$ for compression matrix $A\in\mathbb{R}^{r\times d}$ and decompression matrix $B\in\mathbb{R}^{d'\times r}$.
Using the input activation $X\in\mathbb{R}^{d\times l}$ ($l$ is the calibration sample length),
ASVD aims to minimize the activation loss:
\begin{align}
   \mathcal{L}_1 &=
   \mathbb{E}_X \big\| WX - \hat{W}X \big\|^2
   =
   \mathbb{E}_X \big\| WX - BAX \big\|^2,
\end{align}
instead of the na\"{i}ve weight-based loss:
\begin{align}
    \mathcal{L}_0 &=
    \big\| W - \hat{W} \big\|^2
    =
    \big\| W - BA \big\|^2.
\end{align}
It is well-known that the optimal solution to minimize $\mathcal{L}_0$ can be given by the plain SVD of $W$.
To minimize $\mathcal{L}_1$, ASVD introduced a pre-conditioning matrix $P\in\mathbb{R}^{d\times d}$ to whiten the statistical impact of the activation $X$.
Specifically, ASVD uses the low-rank matrices given by whitened SVD:
\begin{align}
    BAP = \mathsf{svd}_{r}[WP],
    \label{eq:asvd}
\end{align}
where $\mathsf{svd}_r[\cdot]$ denotes the rank-$r$ truncated SVD.

Although ASVD originally suggested a diagonal $\ell_1$-norm pre-conditioning, 
the optimal pre-conditioning matrix $P$ can be given by reformulating $\mathcal{L}_1$ as follows: 
\begin{align}
   \mathcal{L}_1 &=
   \mathrm{tr}\big[ 
   (W-BA) \mathbb{E}_X[X X^\top] (W-BA)^\top
   \big]
   \\
   &=
   \big\|
   (W-BA)C^\frac{1}{2}
   \big\|^2
   = 
   \big\|
   WC^\frac{1}{2}-BAC^\frac{1}{2}
   \big\|^2
   ,
   \label{eq:l1_mod}
\end{align}
where $C=\mathbb{E}_X[XX^\top]\in\mathbb{R}^{d\times d}$ is a covariance (precisely, auto-correlation) of input activation.
Hence, the above loss can be minimized by the SVD: $BAC^\frac{1}{2}=\mathsf{svd}_r[WC^\frac{1}{2}]$.
Accordingly, it is found that the optimal pre-conditioner is the square-root covariance: $P=C^\frac{1}{2}$.
Given the finite calibration data $X$, we can estimate the covariance as $C=XX^\top + \lambda I$, where the damping factor $\lambda\in\mathbb{R}_+$ corresponds to the shrunk  estimator~\cite{ledoit2004well}.

\vskip 1em
\begin{remark}
Different pre-conditioning methods were introduced in several techniques including pruning and quantization, as listed in \cref{tab:cond}.
As those variants are sub-optimal, we use the optimal root covariance: $P=C^\frac{1}{2}$.
See more discussion in \cref{sec:cond}.
\end{remark}

\begin{remark}
In the presence of a bias term, the optimal solution is modified accordingly (from auto-correlation to covariance). 
See \cref{sec:bias1}.
\end{remark}

\begin{remark}
Scaling the covariance has no impact in the performance, and it can be often normalized as $C=(XX^\top+\lambda I)/l$.
\end{remark}

\subsection{Junction matrix for model compression}

In fact, the solution of \cref{eq:asvd} does not have a unique decomposition into low-rank matrices $B$ and $A$.
The truncated SVD is written as
\begin{align}
    U S V &= \mathsf{svd}_r[WP],
\end{align}
where $U\in\mathbb{R}^{d'\times r}$, $S\in\mathbb{R}^{r\times r}$, and $V\in\mathbb{R}^{r\times d}$ are the left singular unitary matrix, singular-value diagonal matrix, and right singular unitary matrix, respectively.
The decompression and compression matrices $B$ and $A$ can be expressed:
\begin{align}
    B &= USJ, \qquad
    A = J^+VP^+,
\end{align}
where $J\in\mathbb{R}^{r\times r}$ is a junction matrix and $[\cdot]^+$ denotes the pseudo inverse.
Choosing any junction matrix that satisfies $SJJ^+=S$ has no impact on the loss.
Hence, there is few literature discussing the choice of $J$.
Typically, one may use $J=I$ to put singular-values into the decompression matrix; $J=S^+$ to put it into the compression matrix; or $J=[S^\frac{1}{2}]^+$ to split it across both matrices equally.

However, a certain choice of $J$ has a noticeable advantage to reduce the number of parameters and floating-point operations (FLOPs).
We can write the whitened right-singular matrix $VP^+$ as two sub-blocks:
\begin{align}
    VP^+ &= 
    \begin{bmatrix}
    V_1 &
    V_2
    \end{bmatrix},
\end{align}
for $V_1\in\mathbb{R}^{r\times r}$ and $V_2\in\mathbb{R}^{r \times (d-r)}$.
When we use $J=V_1$, the compression matrix $A$ will contain an identity block as long as $V_1$ is non-singular:
\begin{align}
    A &= J^+ VP^+ =
    V_1^+ \begin{bmatrix}
    V_1 &
    V_2
    \end{bmatrix} 
    = 
    \begin{bmatrix}
    I &
    V_1^+ V_2
    \end{bmatrix}.
\end{align}
This can greatly reduce the number of parameters from $r(d'+d)$ to $r(d'+d)-r^2$, as well as the FLOPs, because no computation is needed for the identity projection.

For example, when the hidden size is $d=d'$, even if we compress it by $25\%$, i.e., the latent size is $r=0.75 d$, the total number of parameters will be $r(d'+d)=1.5d^2$, which is $50\%$ more than the original $d^2$.
This increased FLOPs hinders the low-rank compression of LLMs, even with the KV cache benefit~\cite{liu2024deepseek, yuan2023asvd, chang2024palu}. 
Nevertheless, with our identity block form, we can always reduce the number of parameters regardless of the latent size, i.e., $r(d'+d)-r^2<d'd$ for $r<\min(d',d)$.
For the above example of $25\%$ latent compression, we can achieve $r(d'+d)-r^2=(15/16)d^2<d^2$.

\begin{figure}[tb]
\begin{center}
\includegraphics[width=\linewidth]{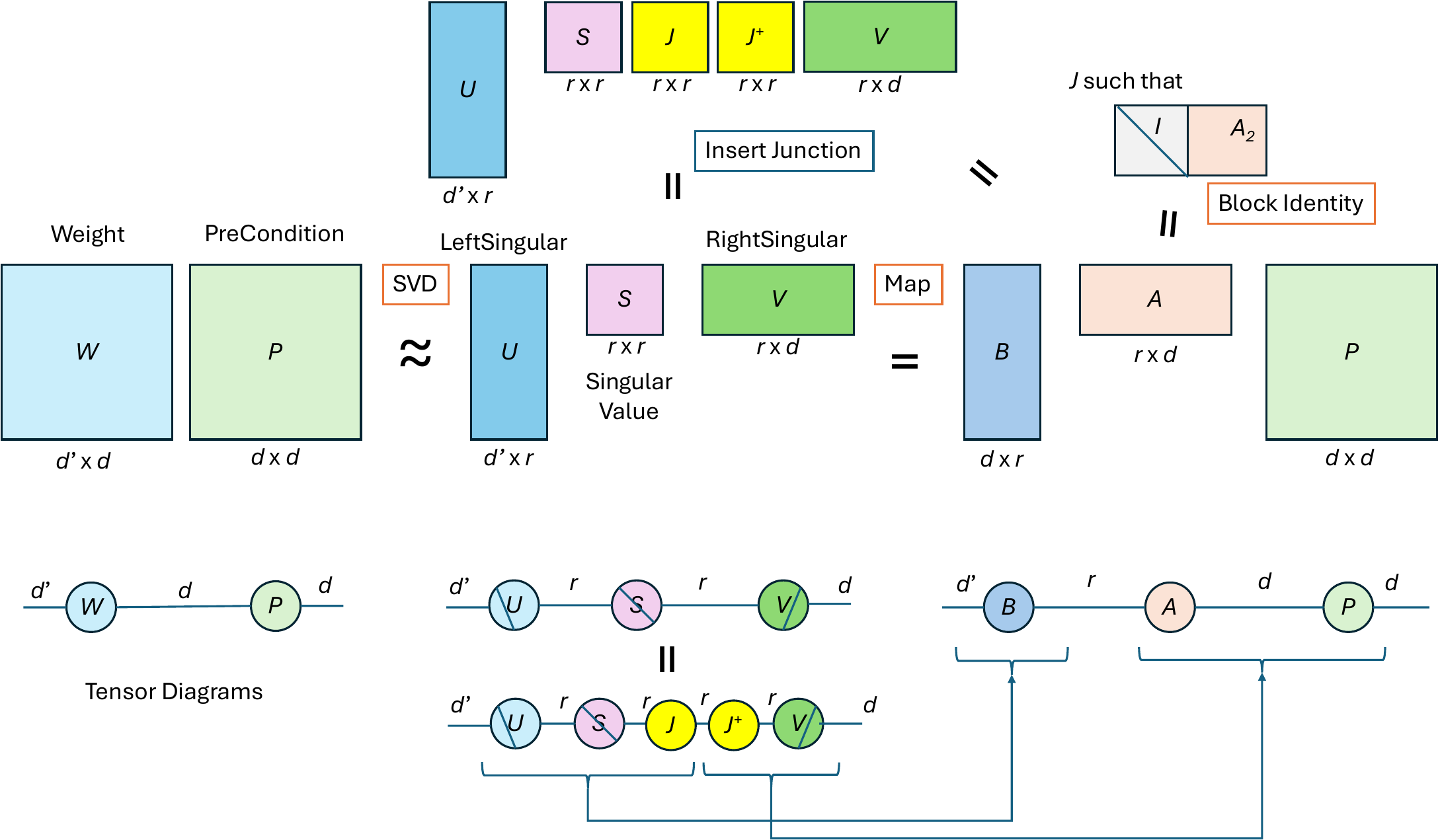}

\caption{Activation-aware compression with pre-conditioning and junction matrix.
The junction matrix $J$ can be adjusted such that $A$ or $B$ is block identity to save the number of parameters and inference computation.
}
\label{fig:junc}
\end{center}
\vskip -0.2in
\end{figure}

\cref{fig:junc} depicts the role of the pre-conditionning and junction matrices for the activation-aware compression.
We also illustrate the tensor diagrams to understand the flexibility of tensor mapping.

\vskip 1em
\begin{remark}
Pivoting columns can solve the case when the left-most sub-block $V_1$ is singular.  
The pivoting does not require any FLOPs in inference while additional memory is required to record the permutation index.
\end{remark}
\begin{remark}
$J=V_1$ is not the only useful choice: 
(i) we can transform $B$ into a block-identity form in a similar manner; or (ii) we can nullify the upper triangular of both $B$ and $A$ like LU factorization. 
See more discussion in \cref{sec:junc}.
\end{remark}

\section{LatentLLM: Joint tensor compression}

The SVD described above is optimal in the sense that the local error  is minimized for the single tensor compression, whereas it does not guarantee global optimality.
Motivated by the global loss minimization framework introduced by SparseLLM~\cite{bai2024sparsellm}, we propose a joint tensor compression technique which factorizes multiple tensors in adjacent modules concurrently.

\subsection{Multi-head latent attention: Joint QK SVD}

First, we consider a joint compression of query (Q) and key (K) projections in MHA to convert into MLA.
The attention map is the dot product of query and key features: 
\begin{align}
    M_i &= X^\top W_{\mathrm{q},i}^\top W_{\mathrm{k},i} X,
\end{align}
where $M_i\in\mathbb{R}^{l\times l}$ is the $i$th head attention map before softmax operation, $W_{\mathrm{q},i}\in\mathbb{R}^{d_\mathrm{h}\times d}$ is the $i$th head query projection matrix, and $W_{\mathrm{k},i}\in\mathbb{R}^{d_\mathrm{h}\times d}$ is the $i$th head key projection matrix.
Here, $d_\mathrm{h}$ is the head dimension, which is often $d_\mathrm{h}=d/h$ for the number of heads $h$.

Rather than individually compressing Q and K projections, we consider minimizing the attention map error:
\begin{align}
    \mathcal{L}_2 &=
    \sum_{i=1}^h \big\| 
    M_i - \hat{M}_i
    \big\|^2,
\end{align}
where $\hat{M}_i$ is the $i$th head latent attention with the low-rank compression:
\begin{align}
    \hat{M}_i &=
    X^\top 
    A_\mathrm{q}^\top B_{\mathrm{q},i}^\top
    B_{\mathrm{k},i}
    A_\mathrm{k}
    X,
\end{align}
where $A_\mathrm{q}\in\mathbb{R}^{r_\mathrm{q}\times d}$ is the Q latent compression matrix, $A_\mathrm{k}\in\mathbb{R}^{r_\mathrm{k}\times d}$ is the K latent compression matrix,
$B_{\mathrm{q},i}\in\mathbb{R}^{d_\mathrm{h} \times r_\mathrm{q}}$ is the $i$th head Q latent decompression matrix, and $B_{\mathrm{k},i}\in\mathbb{R}^{d_\mathrm{h} \times r_\mathrm{k}}$ is the $i$th head K latent decompression matrix, respectively.
Here, $r_\mathrm{q}$ and $r_\mathrm{k}$ are the latent dimensions for Q and K.

\begin{figure}[tb]
\begin{center}
\includegraphics[width=\linewidth]{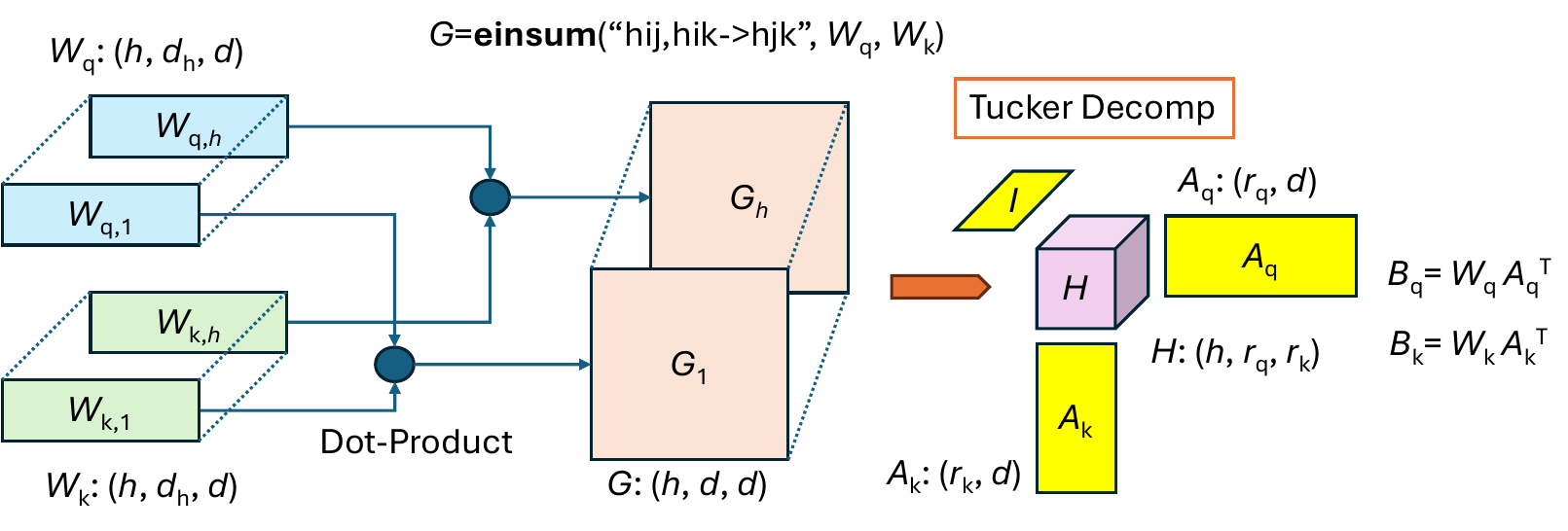}

\caption{Tucker decomposition for joint QK compression.
The compression matrices $A_\mathrm{q}$ and $A_\mathrm{k}$ correspond to the Tucker tensor planes, while $H=A_\mathrm{q} G A_\mathrm{k}^\top$ is the Tucker tensor core.
For simplicity, we omit junction matrices and pre-conditioning matrix.
}
\label{fig:tucker}
\end{center}
\vskip -0.2in
\end{figure}

Similar to \cref{eq:l1_mod}, we can write 
\begin{align}
    \mathcal{L}_2 &=
    \sum_{i=1}^h 
    \big\|
    \underbrace{
    C^\frac{1}{2}
    W_{\mathrm{q},i}^\top W_{\mathrm{k},i} C^\frac{1}{2}
    }_{G_i\in\mathbb{R}^{d\times d}}
    - 
    \underbrace{
    C^\frac{1}{2}
    A_\mathrm{q}^\top
    }_{A_\mathrm{q}'^\top}
    \underbrace{
    B_{\mathrm{q},i}^\top
    B_{\mathrm{k},i}
    }_{H_i\in\mathbb{R}^{r_\mathrm{q}\times r_\mathrm{k}}}
    \underbrace{
    A_\mathrm{k}
    C^\frac{1}{2}
    }_{A_\mathrm{k}'}
    \big\|^2
    \notag
    \\
    &=
    \sum_{i=1}^h
    \big\| 
    G_i - A_\mathrm{q}'^\top H_i A_\mathrm{k}'
    \big\|^2
    .
    \label{eq:l2_hosvd}
\end{align}
This is known as a high-order SVD (HOSVD) problem to decompose for the 3-mode tensor $G\in\mathbb{R}^{h\times d\times d}$, whose $i$th slice is $G_i$.
$A_\mathrm{q}'$ corresponds to the 2nd tensor plane, $A_\mathrm{k}'$ is the 3rd tensor plane, and $H\in\mathbb{R}^{h\times r_\mathrm{q}\times r_\mathrm{k}}$, whose $i$th slice is $H_i$, is the tensor core.
This is illustrated in \cref{fig:tucker}.

This Tucker tensor decomposition is typically solved by alternating SVD over each slice sequentially.
\cref{alg:jsvd} shows the pseudo-code of the joint SVD compression for QK latent projections.
See the detailed derivations of the joint SVD algorithm in \cref{sec:att}.
Here, we generalize the pre-conditioning matrix $P$, as not necessarily the optimal $C^\frac{1}{2}$.
In addition, we explicitly denoted any arbitrary junction matrices that do not change the error.
Note that there are additional junction matrices per heads $J_i\in\mathbb{R}^{d_\mathrm{h}\times d_\mathrm{h}}$ as well as individual Q/K junctions $J_\mathrm{q}\in\mathbb{R}^{r_\mathrm{k}\times r_\mathrm{k}}$ and $J_\mathrm{k}\in\mathbb{R}^{r_\mathrm{k}\times r_\mathrm{k}}$.
This suggests that we can further reduce the number of parameters by transforming into the block identity form per head.
The total number of parameters will be $(r_\mathrm{q}+r_\mathrm{k})(d+d_\mathrm{h}h)-r_\mathrm{q}^2-r_\mathrm{k}^2-d_\mathrm{h}^2h$, reduced from the original weights $2dd_\mathrm{h}h$.

\vskip 1em
\begin{remark}
This conversion from MHA to MLA with joint QK SVD can be readily applied to grouped query attention (GQA). 
See~\cref{sec:gqa}.
\end{remark}

\begin{remark}
Our joint QK SVD can be extended with most positional encoding methods. 
See~\cref{sec:pe}.
\end{remark}

\begin{remark}
A na\"{i}ve way of joint SVD for QKV projections is discussed in \cref{sec:qkv}.
We found this to be worse than the joint QK compression described here.
\end{remark}

\begin{remark}
In the presence of bias terms, the solution is modified slightly. 
See \cref{sec:bias2}.
\end{remark}

\begin{remark}
Attention-aware pruning~\cite{liang2024beyond} is related to our method, while our derivation provides an optimal tensor rank decomposition and only requires preconditioning matrices.
\end{remark}

\begin{algorithm}[tb]
   \caption{Joint SVD for QK Projections in MHA}
   \label{alg:jsvd}
\begin{algorithmic}
   \STATE {\bfseries Input:} Pre-conditioning $P\in\mathbb{R}^{d\times d}$, query projection heads $W_{\mathrm{q},i}\in\mathbb{R}^{d_\mathrm{h}\times d}$, key projection heads $W_{\mathrm{k},i}\in\mathbb{R}^{d_\mathrm{h}\times d}$, number of heads $h$, rank $r_\mathrm{q}, r_\mathrm{k}$, iteration $N$
   \STATE {\bfseries Initialize:} 
   \STATE $W_{\mathrm{q},i}= W_{\mathrm{q},i} P$ for $i \in \{1,\ldots, h\}$
   \STATE $W_{\mathrm{k},i}= W_{\mathrm{k},i} P$  for $i \in \{1,\ldots, h\}$
   \STATE $G_i = W_{\mathrm{q},i}^\top W_{\mathrm{k},i}$  for $i \in \{1,\ldots, h\}$

   \STATE $A_\mathrm{q} = \mathsf{RightSingular}_{r_\mathrm{q}}\big[
   \sum_{i=1}^h G_i G_i^\top \big]$
   \FOR{$n=1$ {\bfseries to} $N$}
   \STATE $A_\mathrm{k} = \mathsf{RightSingular}_{r_\mathrm{k}}\big[ \sum_{i=1}^h G_i^\top A_\mathrm{q}^\top A_\mathrm{q} G_i \big]$
   \STATE $A_\mathrm{q} = \mathsf{RightSingular}_{r_\mathrm{q}}\big[ \sum_{i=1}^h G_i A_\mathrm{k} A_\mathrm{k}^\top G_i^\top \big]$
   \ENDFOR
   \STATE {\bfseries Output:} 
   \STATE $B_{\mathrm{q},i}=J_i^\top W_{\mathrm{q},i} A_\mathrm{q}^\top J_\mathrm{q}$ for $i \in \{1,\ldots, h\}$
   \STATE $B_{\mathrm{k},i}=J_i^+ W_{\mathrm{k},i}A_\mathrm{k}^\top J_\mathrm{k}$ for $i \in \{1,\ldots, h\}$
   \STATE $A_\mathrm{q}=J_\mathrm{q}^+ A_\mathrm{q}P^+$
   \STATE $A_\mathrm{k}=J_\mathrm{k}^+ A_\mathrm{k}P^+$
\end{algorithmic}
\end{algorithm}

\subsection{Multi-head latent attention: Joint VO SVD}

Next, we discuss the joint SVD for value (V) and output (O) projections in MHA.
The MHA output can be written as
\begin{align}
    Y' = \sum_{i=1}^h
    W_{\mathrm{o},i} W_{\mathrm{v},i}
    X
    \,
    \mathsf{softmax}[M_i^\top]
    ,
\end{align}
where $W_{\mathrm{o},i}\in\mathbb{R}^{d'\times d_\mathrm{h}}$ is the $i$th head output projection, and $W_{\mathrm{v},i}\in\mathbb{R}^{d_\mathrm{h}\times d}$ is the $i$th head value value projection.
For any arbitrary attention map, we may consider minimizing the loss:
\begin{align}
    \mathcal{L}_3 &=
    \sum_{i=1}^h
    \big\| W_{\mathrm{o},i} W_{\mathrm{v},i} X -
    \hat{W}_{\mathrm{o},i} \hat{W}_{\mathrm{v},i} X
    \big\|^2,
\end{align}
for the low-rank compression: $\hat{W}_{\mathrm{o},i}=B_\mathrm{o}A_{\mathrm{o},i}\in\mathbb{R}^{d'\times d_\mathrm{h}}$ and $\hat{W}_{\mathrm{v},i}=B_{\mathrm{v},i}A_\mathrm{v}\in\mathbb{R}^{d_\mathrm{h}\times d}$ with $B_\mathrm{o}\in\mathbb{R}^{d'\times r_\mathrm{o}}$, $A_{\mathrm{o},i}\in\mathbb{R}^{r_\mathrm{o}\times d_\mathrm{h}}$, 
$B_{\mathrm{v},i}\in\mathbb{R}^{d_\mathrm{h}\times r_\mathrm{v}}$, and $A_{\mathrm{v}}\in\mathbb{R}^{r_\mathrm{v}\times d}$.
The MLA output is thus given as
\begin{align}
    \hat{Y}' &=
    \sum_{i=1}^h
    B_{\mathrm{o}} A_{\mathrm{o},i}
    B_{\mathrm{v},i} A_\mathrm{v}
    X
    \,
    \mathsf{softmax}[M_i].
    \label{eq:vo}
\end{align}
Interestingly, this is also formulated in a similar manner of \cref{eq:l2_hosvd}, and it can be solved by the joint SVD algorithm.

Note that the MLA computations can be more efficient depending on tensor contraction ordering. 
Specifically, \cref{eq:vo} requires complexity of $\mathcal{O}[ldr_\mathrm{v}+hd_\mathrm{d}lr_\mathrm{v}+hd_\mathrm{h}l^2+hd_\mathrm{h}lr_\mathrm{o}+hdlr_\mathrm{o}]$ in the order:
\begin{align}
    \hat{Y}' &=
    \sum_{i=1}^h    
    \Bigg(
    B_{\mathrm{o}} 
    \bigg(
    A_{\mathrm{o},i}
    \Big(
    \big(
    B_{\mathrm{v},i} 
    (A_\mathrm{v}
    X
    )
    \big)
    \,
    \mathsf{softmax}[M_i]
    \Big)
    \bigg)
    \Bigg).
    \label{eq:vo2}
\end{align}
It can be reduced by computing in another order:
\begin{align}
    \hat{Y}' &=
    B_{\mathrm{o}} 
    \sum_{i=1}^h    
    \bigg(
    A_{\mathrm{o},i}
    \Big(
    B_{\mathrm{v},i} 
    \big(
    (A_\mathrm{v}
    X
    )
    \,
    \mathsf{softmax}[M_i]
    \big)
    \Big)
    \bigg)
    ,
    \label{eq:vo3}
\end{align}
which requires complexity of $\mathcal{O}[ldr_\mathrm{v}+ r_\mathrm{v}l^2 + hd_\mathrm{h}lr_\mathrm{v}+hd_\mathrm{h}lr_\mathrm{o}+dlr_\mathrm{o}]$.
The reduction is $\mathcal{O}[(d-r_\mathrm{v})l^2 + (h-1)dlr_\mathrm{o}]$.
If $hr_\mathrm{o}<r_\mathrm{v}$, then the attention weighting should be done on output compression $A_{\mathrm{o},i}$.

\vskip 1em
\begin{remark}
Because the above loss does not deal with the nonlinear attention map, it was found that the joint VO compression was not effective over the split V/O compression.
Nonetheless, we derived several different alternatives.
See \cref{sec:vo}.
\end{remark}

\begin{table*}[t]
\centering
\caption{Perplexity ($\downarrow$) of OPT models with different SVD compression methods for $10$--$40\%$ size reduction. 
Asterisk ``*'' indicates the better performance than the original un-compressed LLM.}
\label{tab:perp_opt}
\small
\setlength{\tabcolsep}{2pt}
\begin{tabular}{l rrr rrr rrr rrr}
  \toprule
  Compression & \multicolumn{3}{c}{10\%} & \multicolumn{3}{c}{20\%} & \multicolumn{3}{c}{30\%} 
  & \multicolumn{3}{c}{40\%} 
  \\
  \cmidrule(lr){2-4}
  \cmidrule(lr){5-7}
  \cmidrule(lr){8-10}  
  \cmidrule(lr){11-13}  
  Dataset & WT2 & PTB & C4 & WT2 & PTB & C4 & WT2 & PTB & C4 & WT2 & PTB & C4 \\
  \midrule
  \multicolumn{13}{c}{OPT-125M (WT2: 27.7, PTB: 39.0, C4: 26.6)}
  \\
  \midrule
  Plain SVD (Identity) & 
  393.8 & 608.8 & 274.6 &
  668.9 & 1098.0 & 559.0 &
  1298.3 & 1888.7 & 806.5 &
  3306.5 & 2985.9 & 1637.0 
  \\
  ASVD (Hessian)  & 
  57.8 & 92.8 & 45.0 &
  106.9 & 169.8 & 79.9 &
  288.1 & 530.4 & 215.0 &
  838.9 & 1581.9 & 608.2 
  \\
  ASVD ($\ell_2$-norm)  & 
  49.7 & 74.7 & 42.2 &
  87.3 & 126.8 & 72.0 &
  256.0 & 282.1 & 188.3 &
  906.9 & 864.3 & 528.4 
  \\
  ASVD (Cov) &
  87.5 & 121.5 & 67.6 &
  115.7 & 157.0 & 83.1 &
  163.1 & 242.8 & 109.9 &
  248.3 & 390.6 & 158.4 
  \\
  ASVD (RootCov) &
  40.5 & 64.4 & 34.5 &
  54.8 & 86.8 & 42.7 &
  88.8 & 148.9 & 61.5 &
  177.5 & 306.7 & 116.8 
  \\
  \rowcolor{lgreen}
  LatentLLM (RootCov) &
  \bf{29.0} & \bf{42.3} & \bf{27.6} &
  \bf{32.9} & \bf{50.9} & \bf{30.4} &
  \bf{43.4} & \bf{68.7} & \bf{37.4} &
  \bf{73.4} & \bf{68.7} & \bf{37.4} 
  \\
  \bottomrule
  \toprule
  \multicolumn{13}{c}{OPT-350M (WT2: 22.0, PTB: 31.1, C4: 22.6)}
  \\
  \midrule
  Plain SVD (Identity) &
  112.3 & 130.8 & 82.8 &
  211.3 & 226.8 & 151.5 &
  378.1 & 392.0 & 258.7 &
  705.5 & 635.5 & 509.8 
  \\
  ASVD (Hessian) &
  64.0 & 89.1 & 50.9 &
  104.6 & 134.4 & 80.3 &
  202.1 & 212.0 & 145.9 &
  557.3 & 558.6 & 371.6 
  \\
  ASVD ($\ell_2$-norm) &
  40.0 & 59.9 & 36.6 &
  59.4 & 78.0 & 49.8 &
  117.5 & 134.2 & 86.9 &
  308.7 & 283.9 & 201.1
  \\
  ASVD (Cov) &
  78.0 & 90.6 & 61.7 &
  100.8 & 111.0 & 72.7 &
  311.2 & 356.8 & 129.4 &
  1485.3 & 922.7 & 548.2 
  \\
  ASVD (RootCov) & 
  30.8 & 42.2 & 28.5 &
  39.0 & 51.4 & 33.6 &
  71.6 & 86.1 & 49.5 &
  118.5 & 132.1 & 73.0 
  \\
  \rowcolor{lgreen}
  LatentLLM (RootCov) &
  \bf{23.1} & \bf{33.3} & \bf{23.6} &
  \bf{25.9} & \bf{37.0} & \bf{25.8} &
  \bf{32.9} & \bf{45.0} & \bf{30.6} &
  \bf{51.3} & \bf{63.4} & \bf{42.5}
  \\
  \bottomrule
  \toprule
  \multicolumn{13}{c}{OPT-1.3B (WT2: 14.6, PTB: 20.3, C4: 16.1)}
  \\
  \midrule
  Plain SVD (Identity) &
  9428.1 & 10670.8 & 4865.4 &
  16461.2 & 20589.0 & 11039.8 &
  18105.3 & 17360.8 & 12565.2 &
  22155.9 & 15820.3 & 16566.2 
  \\
  ASVD (Hessian) &
  23.8 & 40.6 & 24.9 &
  63.0 & 173.7 & 52.8 &
  825.8 & 927.9 & 385.0 &
  4912.3 & 3086.3 & 2138.9 
  \\
  ASVD ($\ell_2$-norm) & 
  20.3 & 32.3 & 21.6 &
  28.7 & 60.2 & 27.7 &
  74.5 & 217.4 & 58.5 &
  592.4 & 1072.0 & 336.7
  \\
  ASVD (Cov) &
  29750.9 & 31499.1 & 18646.3 &
  19716.9 & 21757.2 & 14967.2 &
  21738.3 & 24300.2 & 16428.7 &
  22776.5 & 23591.7 & 14922.1 
  \\
  ASVD (RootCov) &
  17.7 & 27.9 & 18.9 &
  21.9 & 35.3 & 22.2 &
  33.9 & 55.8 & 29.7 &
  75.0 & 107.9 & 51.1
  \\
  \rowcolor{lgreen}
  LatentLLM (RootCov) &
  *\bf{14.5} & \bf{21.5} & \bf{16.6} &
  \bf{15.8} & \bf{24.3} & \bf{17.8} &
  \bf{20.2} & \bf{31.6} & \bf{21.3} &
  \bf{34.1} & \bf{58.1} & \bf{30.6}
  \\
  \bottomrule
  \toprule
  \multicolumn{13}{c}{OPT-2.7B (WT2: 12.5, PTB: 18.0, C4: 14.3)}
  \\
  \midrule
  Plain SVD (Identity) &
  1922.0 & 2250.3 & 900.7 &
  7446.2 & 7042.4 & 5113.6 &
  11253.8 & 10109.6 & 7742.6 &
  26177.5 & 29321.3 & 17035.3
  \\
  ASVD (Hessian) & 
  18.2 & 31.9 & 20.0 &
  31.6 & 96.9 & 28.0 &
  216.2 & 852.3 & 74.8 &
  2714.9 & 2894.0 & 626.0 
  \\
  ASVD ($\ell_2$-norm) &
  16.9 & 27.1 & 18.7 &
  23.1 & 44.6 & 23.4 &
  53.0 & 190.7 & 43.2 &
  524.3 & 981.5 & 229.3
  \\
  ASVD (Cov) &
  16419.9 & 15136.0 & 10680.6 &
  15495.8 & 14896.4 & 10891.6 &
  17392.3 & 15994.8 & 11926.0 &
  17976.5 & 16298.1 & 11566.8
  \\
  ASVD (RootCov) &
  14.5 & 22.1 & 16.5 &
  17.1 & 26.7 & 18.8 &
  24.1 & 36.3 & 23.7 &
  48.4 & 66.5 & 37.1 
  \\
  \rowcolor{lgreen}
  LatentLLM (RootCov) &
  *\bf{12.3} & \bf{18.8} & \bf{14.7} &
  \bf{13.6} & \bf{20.6} & \bf{15.7} &
  \bf{16.5} & \bf{24.3} & \bf{18.1} &
  \bf{24.5} & \bf{36.0} & \bf{24.2}
  \\
  \bottomrule
  \toprule
  \multicolumn{13}{c}{OPT-6.7B (WT2: 10.9, PTB: 15.8, C4: 12.7)}
  \\
  \midrule
  Plain SVD (Identity) &
  14839.0 & 28665.9 & 22936.1 &
  67517.7 & 116974.8 & 110860.5 &
  123286.4 & 213333.5 & 190378.4 &
  27304.0 & 31719.7 & 24071.3
  \\
  ASVD (Hessian) & 
  14.3 & 22.0 & 16.6 &
  17.3 & 27.3 & 20.1 &
  26.0 & 51.0 & 28.8 &
  73.3 & 252.2 & 67.6 
  \\
  ASVD ($\ell_2$-norm) &
  12.6 & 19.6 & 15.1 &
  14.6 & 23.0 & 17.2 &
  18.7 & 32.1 & 21.4 &
  30.6 & 73.2 & 33.7
  \\
  ASVD (Cov) &
  9111.6 & 9171.3 & 7220.2 &
  9842.6 & 9465.6 & 7175.0 &
  11848.0 & 10046.0 & 6973.6 &
  8514.7 & 7931.2 & 6660.3
  \\
  ASVD (RootCov) &
  11.8 & 17.7 & 14.2 &
  13.5 & 19.5 & 15.4 &
  17.0 & 23.9 & 17.8 &
  27.2 & 36.1 & 24.0 
  \\
  \rowcolor{lgreen}
  LatentLLM (RootCov) &
  *\bf{10.7} & \bf{16.1} & \bf{13.0} &
  \bf{11.5} & \bf{17.4} & \bf{13.7} &
  \bf{13.5} & \bf{19.2} & \bf{15.3} &
  \bf{18.0} & \bf{24.2} & \bf{18.4}
  \\
  \bottomrule
  \toprule
  \multicolumn{13}{c}{OPT-13B (WT2: 10.1, PTB: 14.5, C4: 12.1)}
  \\
  \midrule
  Plain SVD (Identity) &
  892.2 & 1003.5 & 789.3 &
  2157.4 & 2068.3 & 1716.1 &
  3612.9 & 3381.8 & 2806.9 &
  5838.7 & 5069.1 & 4292.5
  \\
  ASVD (Hessian) & 
  12.5 & 18.6 & 14.3 &
  14.6 & 22.0 & 15.8 &
  19.1 & 29.7 & 18.7 &
  29.5 & 48.9 & 25.1 
  \\
  ASVD ($\ell_2$-norm) &
  11.2 & 16.8 & 13.4 &
  12.2 & 18.6 & 14.4 &
  14.0 & 21.9 & 16.3 &
  18.2 & 29.0 & 20.3
  \\
  ASVD (Cov) &
  13999.3 & 11053.5 & 8991.2 &
  10250.7 & 8883.2 & 6556.4 &
  12885.3 & 11756.7 & 7658.0 &
  12625.5 & 10709.9 & 7972.3
  \\
  ASVD (RootCov) &
  10.9 & 15.9 & 13.1 &
  11.9 & 17.0 & 13.9 &
  14.3 & 20.0 & 15.3 &
  20.2 & 24.1 & 18.3 
  \\
  \rowcolor{lgreen}
  LatentLLM (RootCov) &
  \bf{10.2} & \bf{14.8} & \bf{12.4} &
  \bf{10.7} & \bf{15.4} & \bf{13.0} &
  \bf{12.0} & \bf{16.7} & \bf{13.9} &
  \bf{14.8} & \bf{19.2} & \bf{15.8}
  \\
  \bottomrule
\end{tabular}
\end{table*}

\subsection{Latent MLP: Joint UD SVD}

Finally, we address the joint compression of MLP layers which consists of up (U) projection and down (D) projection in typical LLMs/LMMs.
Although the global optimization is generally difficult due to the nonlinear activations in the MLP layer,
SparseLLM~\cite{bai2024sparsellm} provides an elegant way to approximate MLP layer.
The key idea is to minimize the MLP loss in a decoupled manner by introducing auxiliary variables.
Our LatentLLM exploits the same philosophy to compress MLP layers.

Consider a 2-layer MLP:
\begin{align}
    Z = W_\mathrm{u} X, \quad
    Z' = \sigma(Z), \quad
    Y &= W_\mathrm{d} Z',
\end{align}
where $W_\mathrm{u}\in\mathbb{R}^{d_\mathrm{i}\times d}$ is the up projection matrix, $W_\mathrm{d}\in\mathbb{R}^{d\times d_\mathrm{i}}$ is the down projection matrix, and $d_\mathrm{i}$ is the intermediate size which is typically four-fold of hidden size: $d_\mathrm{i}=4d$.
The intermediate variables $Z,Z'\in\mathbb{R}^{d_\mathrm{i}\times l}$ are auxiliary factors to be optimized.

Specifically, we consider minimizing the decoupled loss:
\begin{align}
    \mathcal{L}_4 &=
    \alpha \| W_\mathrm{u}X - Z \|^2
    +
    \beta \| Z'-\sigma(Z) \|^2
    +
    \gamma \| W_\mathrm{d} Z' - Y\|^2
    ,
\end{align}
for auxiliary variables $Z$ and $Z'$, given calibration input $X$ and output $Y$.

Following SparseLLM~\cite{bai2024sparsellm}, the optimal $Z'$ can be obtained given the other parameters fixed:
\begin{align}
    Z' &=
    \big(
    \gamma W_\mathrm{d}^\top W_\mathrm{d} + \beta I
    \big)^+ 
    \big(
    \beta \sigma(Z) + \gamma W_\mathrm{d}^\top Y
    \big).
\end{align}
The optimal $Z$ can be also obtained in closed-form for the case of ReLU:
\begin{align}
    Z_- &=
    W_\mathrm{u} X,\quad
    Z_+ =
    \frac{1}{\alpha+\beta} 
    (\alpha Z_- 
    +\beta Z'),
\end{align}
depending on $Z$'s element-wise sign.

This approach can be used for low-rank approximation.
Given $Z$, we can optimize low-rank matrix $\hat{W}_\mathrm{u}=B_\mathrm{u} A_\mathrm{u}$ by SVD of $ZX^+ C^\frac{1}{2}$, where $ZX^+$ corresponds to the effective weight matrix to map $X$ onto $Z$.
Given $Z'$, we approximate $\hat{W}_\mathrm{d}=B_\mathrm{d}A_\mathrm{d}$ by SVD of $Y Z'^+ C_\mathrm{d}^\frac{1}{2}$, given correlation $C_\mathrm{d}=Z' Z'^\top$.
This alternating solution is iterated over a few rounds.
For detail, see \cref{sec:mlp}.

\section{Experiments}

\paragraph{Experiments setup} 

We conduct experiments for LLM and LMM benchmarks to evaluate the effectiveness of our method.
Our experiments are based on the same setting of SparseLLM~\cite{bai2024sparsellm} and their code base\footnote{\url{https://github.com/BaiTheBest/SparseLLM}}.
We implemented LatentLLM in PyTorch~\cite{paszke2019pytorch} and used the
HuggingFace Transformers library \cite{wolf2019huggingface} for handling models and datasets. 
All experiments are conducted on NVIDIA A40 GPUs. 

For LLM calibration data, we follow the same setup of~\cite{frantar2023sparsegpt} and use 64 samples of 2048-token segments, randomly chosen from the first shard of the
C4~\cite{raffel2020exploring} dataset. 
This dataset represents generic text data crawled from the internet and ensures
our experiments are zero-shot as no task-specific data is seen during compression. 
We followed existing
work~\cite{sun2023simple} and compressed all linear layers in MLP and MHA in LLMs to the target compression ratio.
For LMM calibration data, we use 64 samples, randomly chosen from the train split of the ScienceQA~\cite{lu2022learn} multi-modal question answering dataset.

\begin{figure*}[tb]
\begin{center}

\begin{subfigure}[b]{0.32\linewidth}
\includegraphics[width=\linewidth,trim=10 0 40 30,clip]{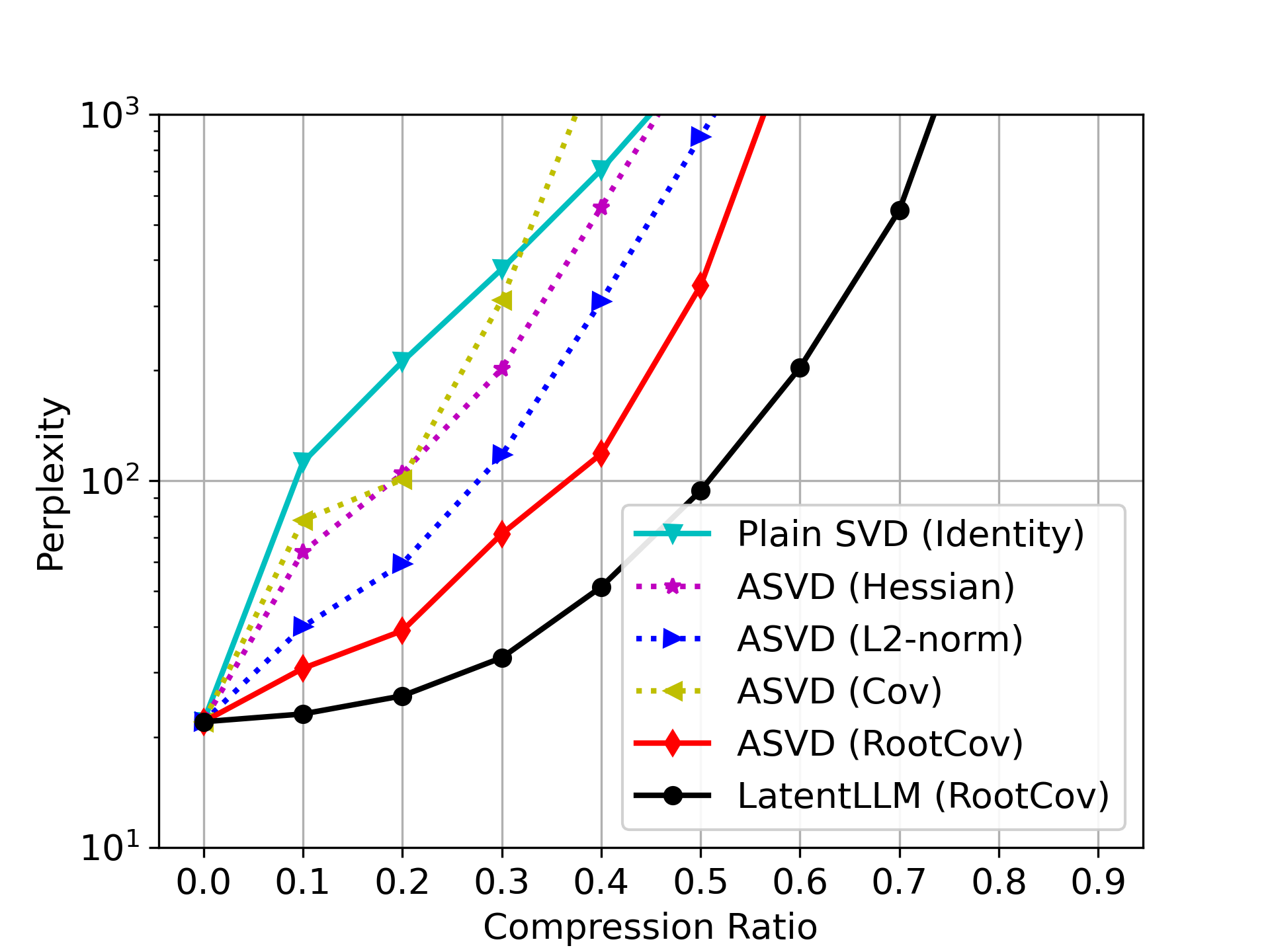}
\caption{WT2 (OPT-350M)}
\end{subfigure}
\begin{subfigure}[b]{0.32\linewidth}
\includegraphics[width=\linewidth,trim=10 0 40 30,clip]{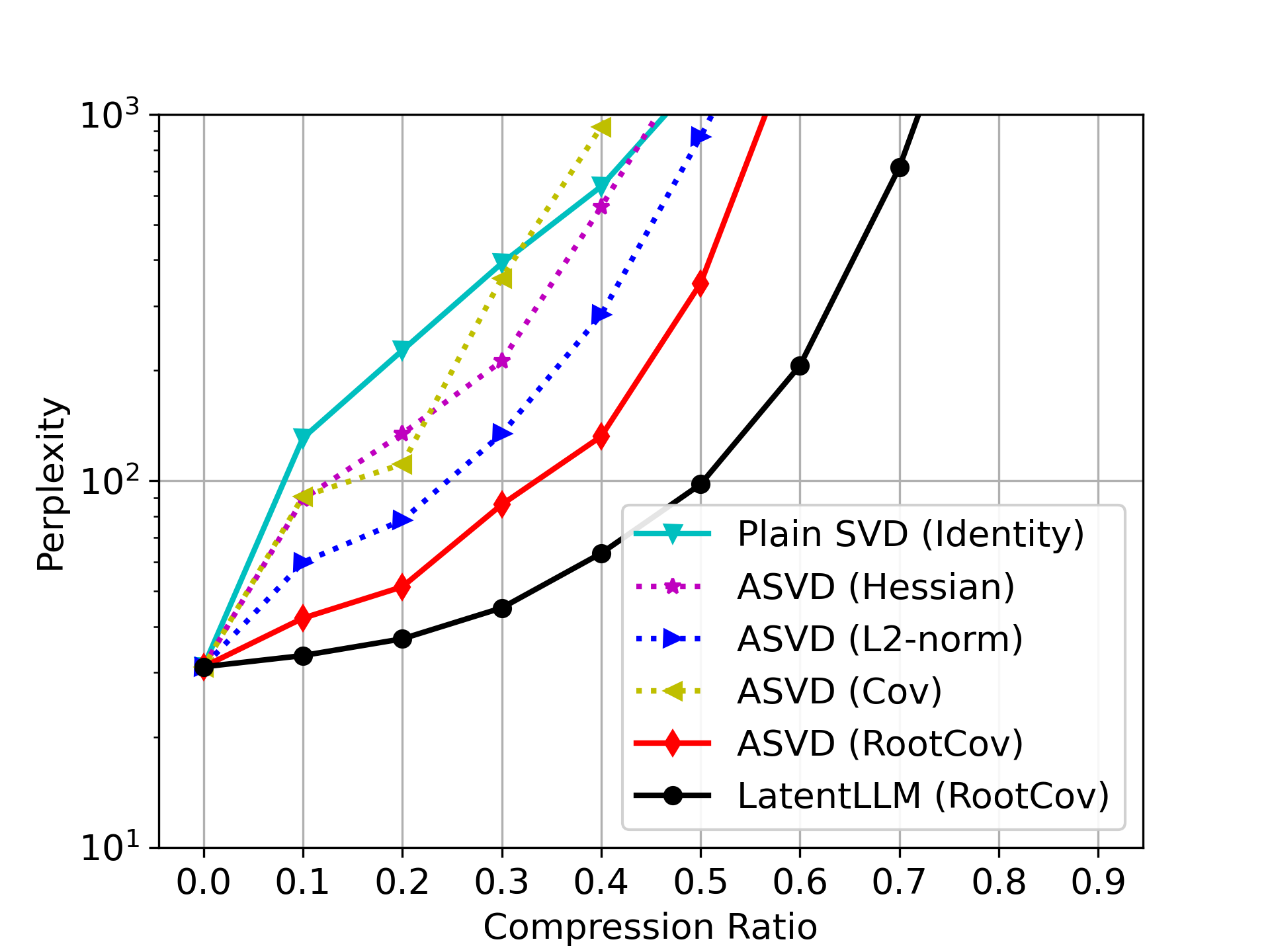}
\caption{PTB (OPT-350M)}
\end{subfigure}
\begin{subfigure}[b]{0.32\linewidth}
\includegraphics[width=\linewidth,trim=10 0 40 30,clip]{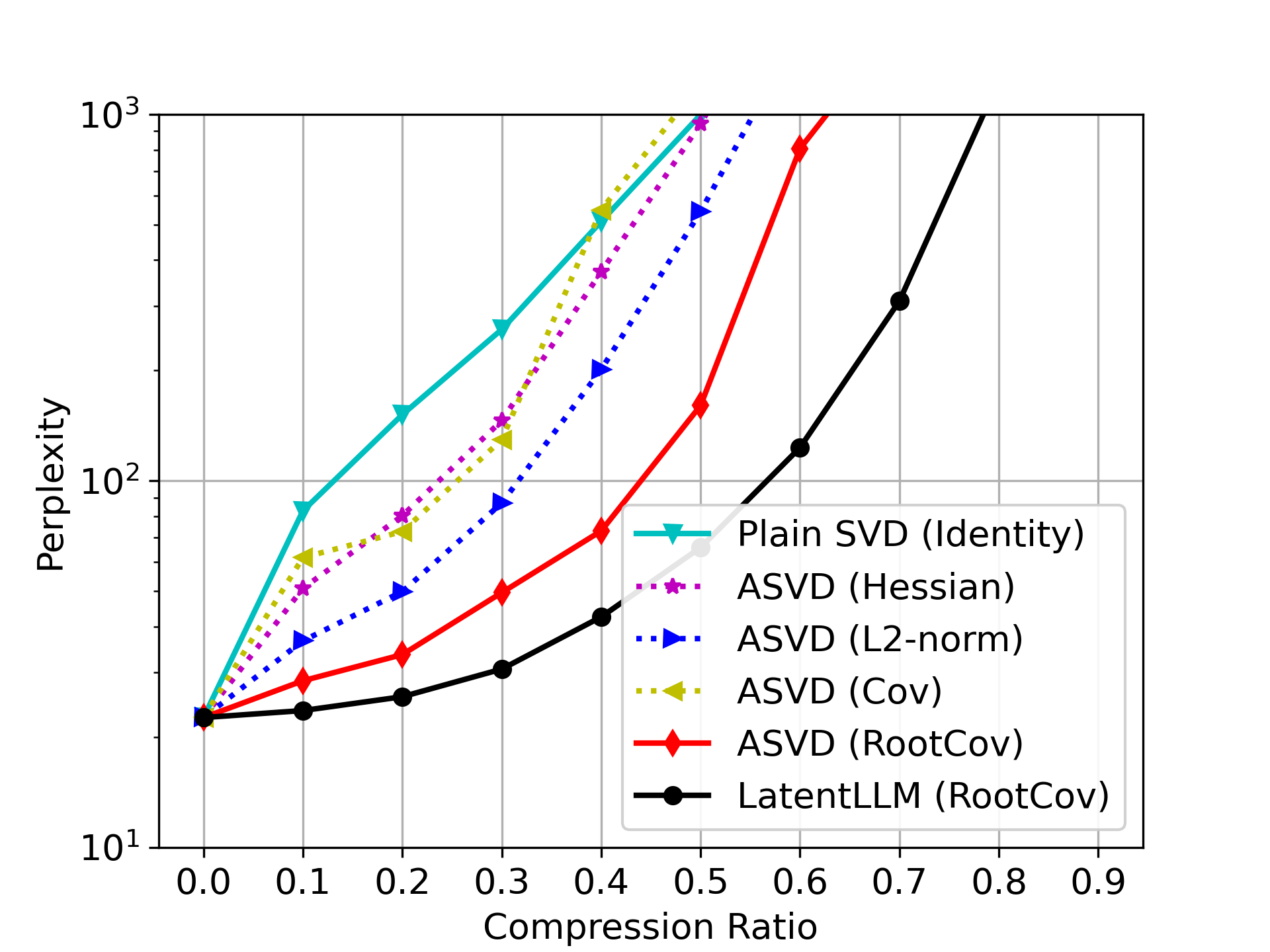}
\caption{C4 (OPT-350M)}
\end{subfigure}

\begin{subfigure}[b]{0.32\linewidth}
\includegraphics[width=\linewidth,,trim=10 0 40 30,clip]{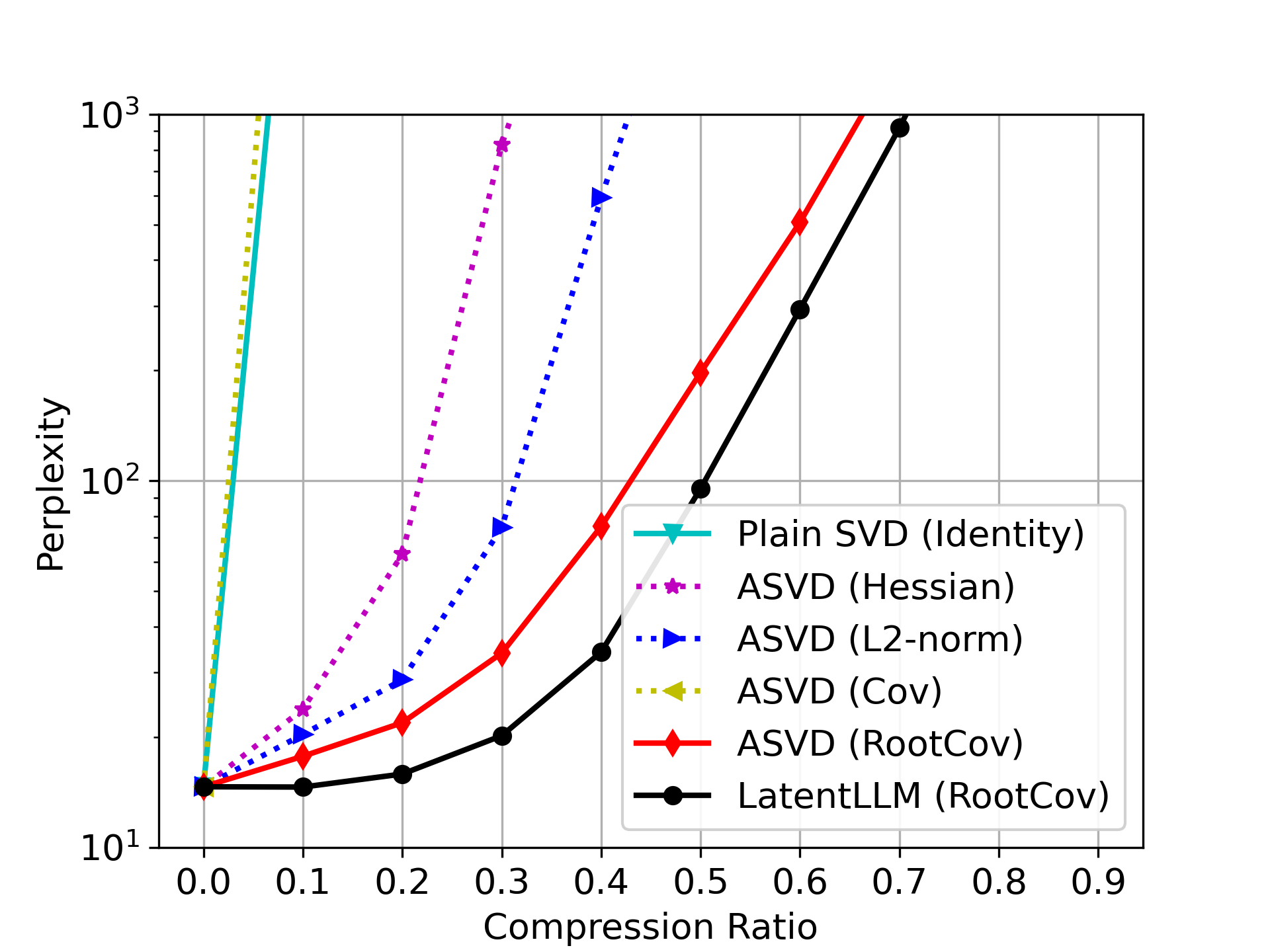}
\caption{WT2 (OPT-1.3B)}
\end{subfigure}
\begin{subfigure}[b]{0.32\linewidth}
\includegraphics[width=\linewidth,,trim=10 0 40 30,clip]{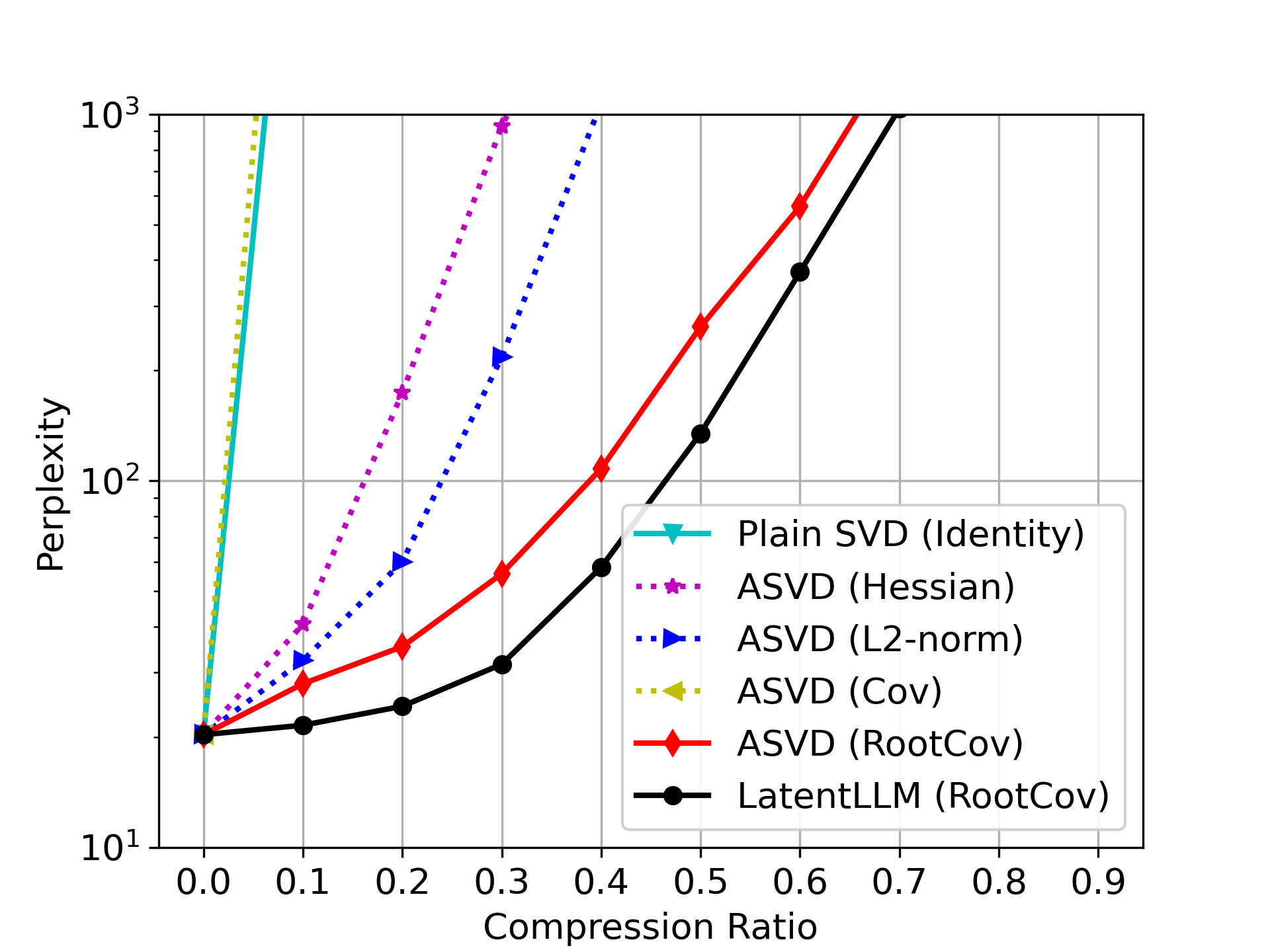}
\caption{PTB (OPT-1.3B)}
\end{subfigure}
\begin{subfigure}[b]{0.32\linewidth}
\includegraphics[width=\linewidth,,trim=10 0 40 30,clip]{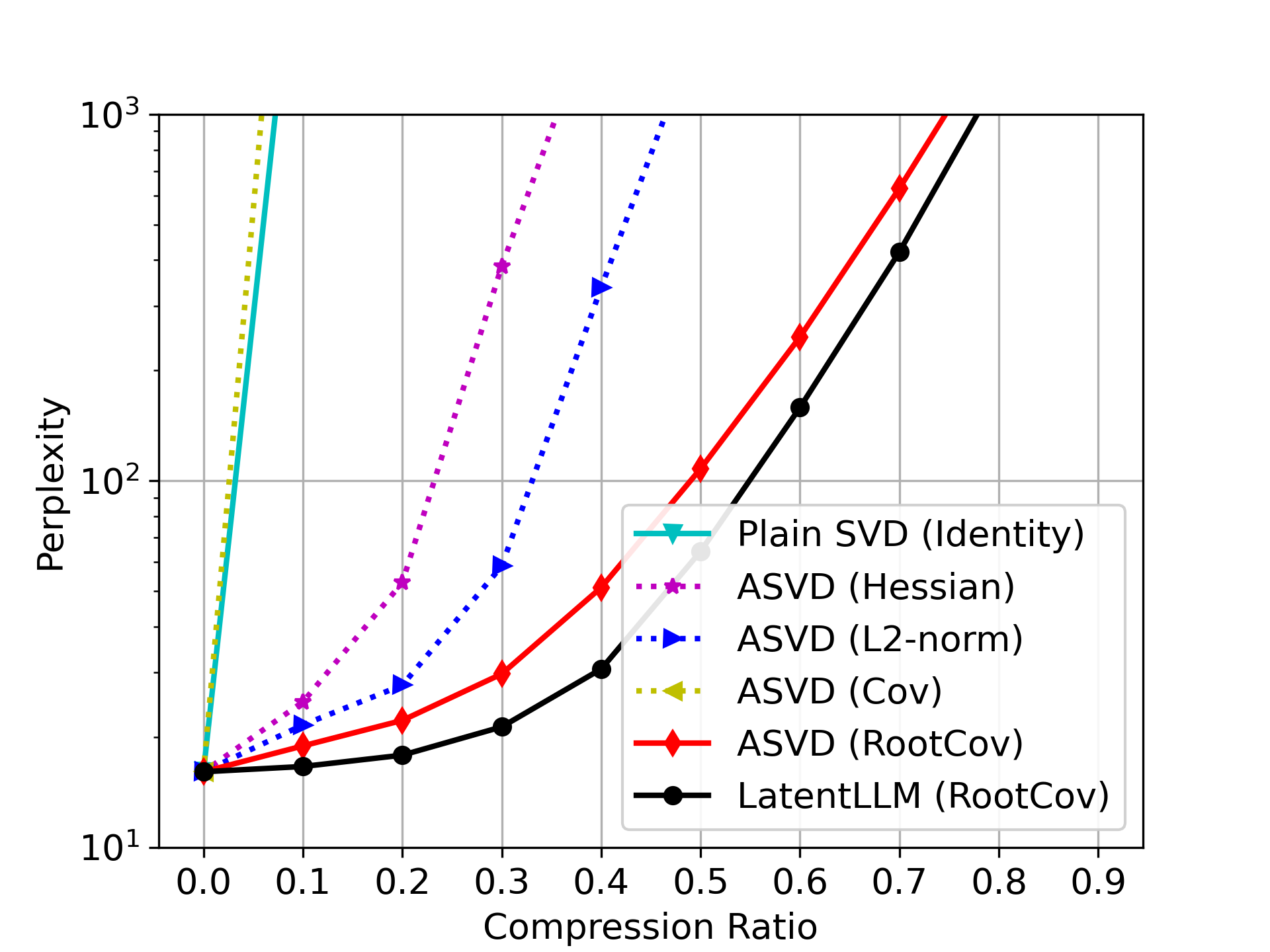}
\caption{C4 (OPT-1.3B)}
\end{subfigure}

\begin{subfigure}[b]{0.32\linewidth}
\includegraphics[width=\linewidth,,trim=10 0 40 30,clip]{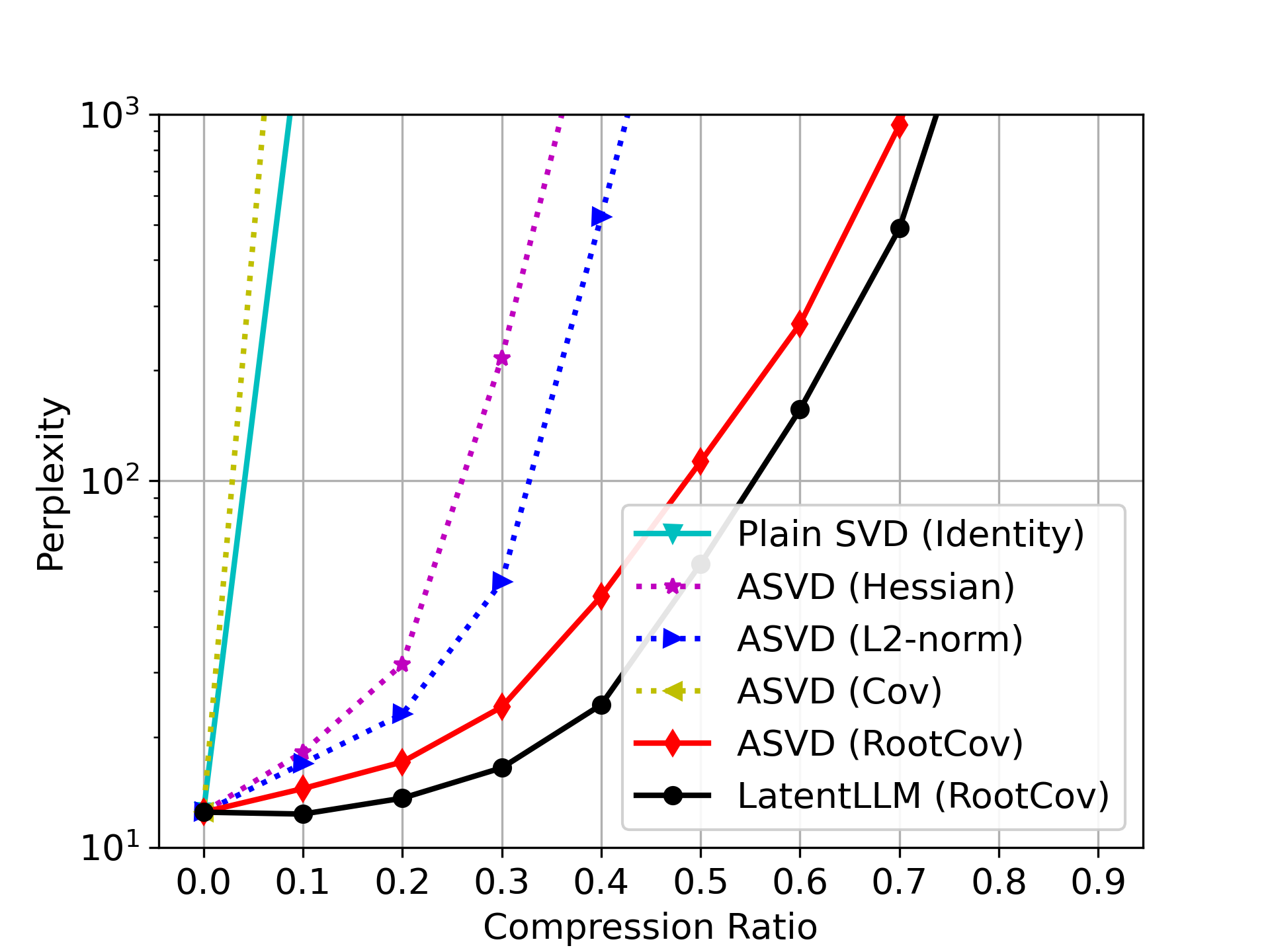}
\caption{WT2 (OPT-2.7B)}
\end{subfigure}
\begin{subfigure}[b]{0.32\linewidth}
\includegraphics[width=\linewidth,,trim=10 0 40 30,clip]{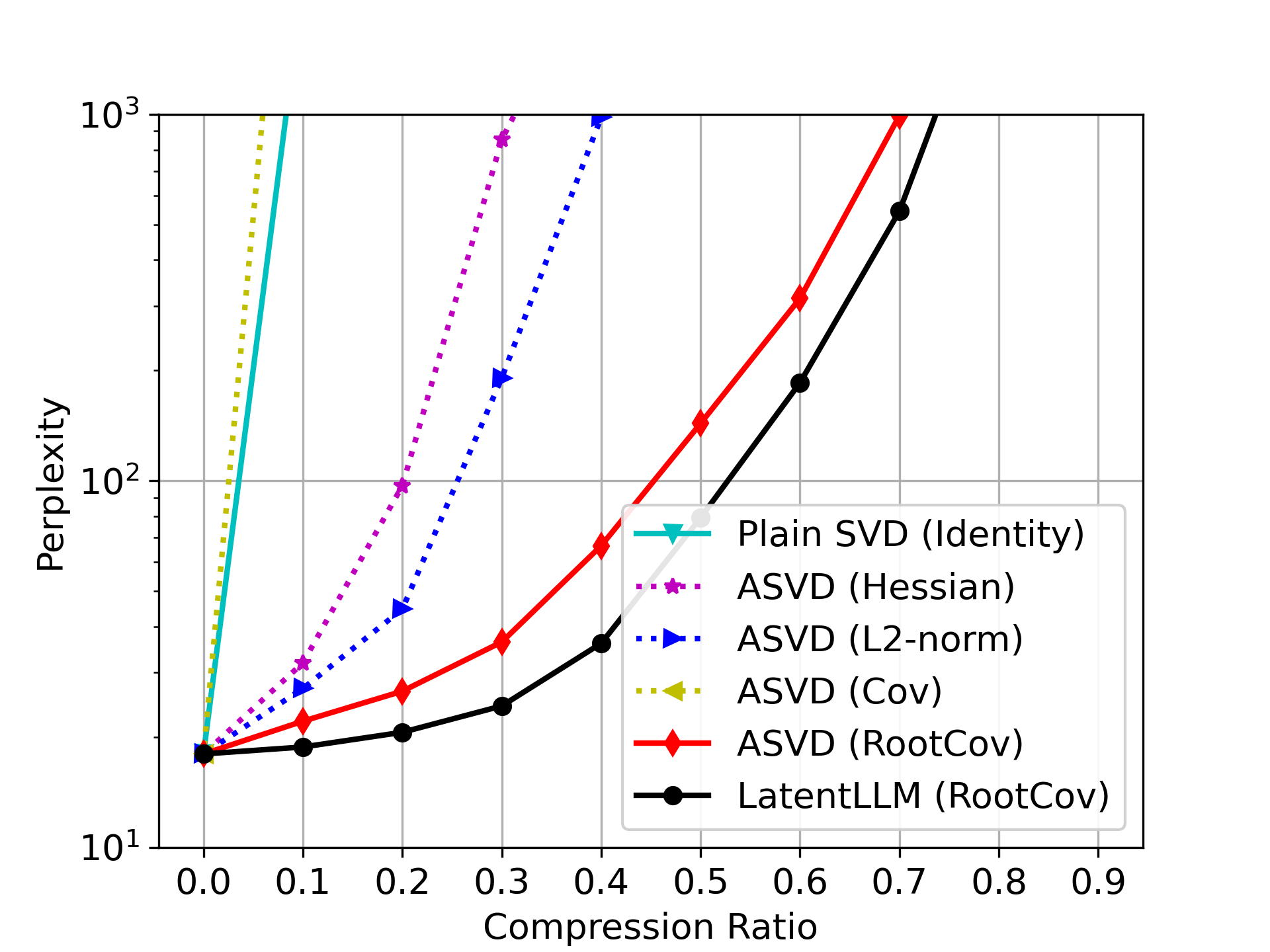}
\caption{PTB (OPT-2.7B)}
\end{subfigure}
\begin{subfigure}[b]{0.32\linewidth}
\includegraphics[width=\linewidth,,trim=10 0 40 30,clip]{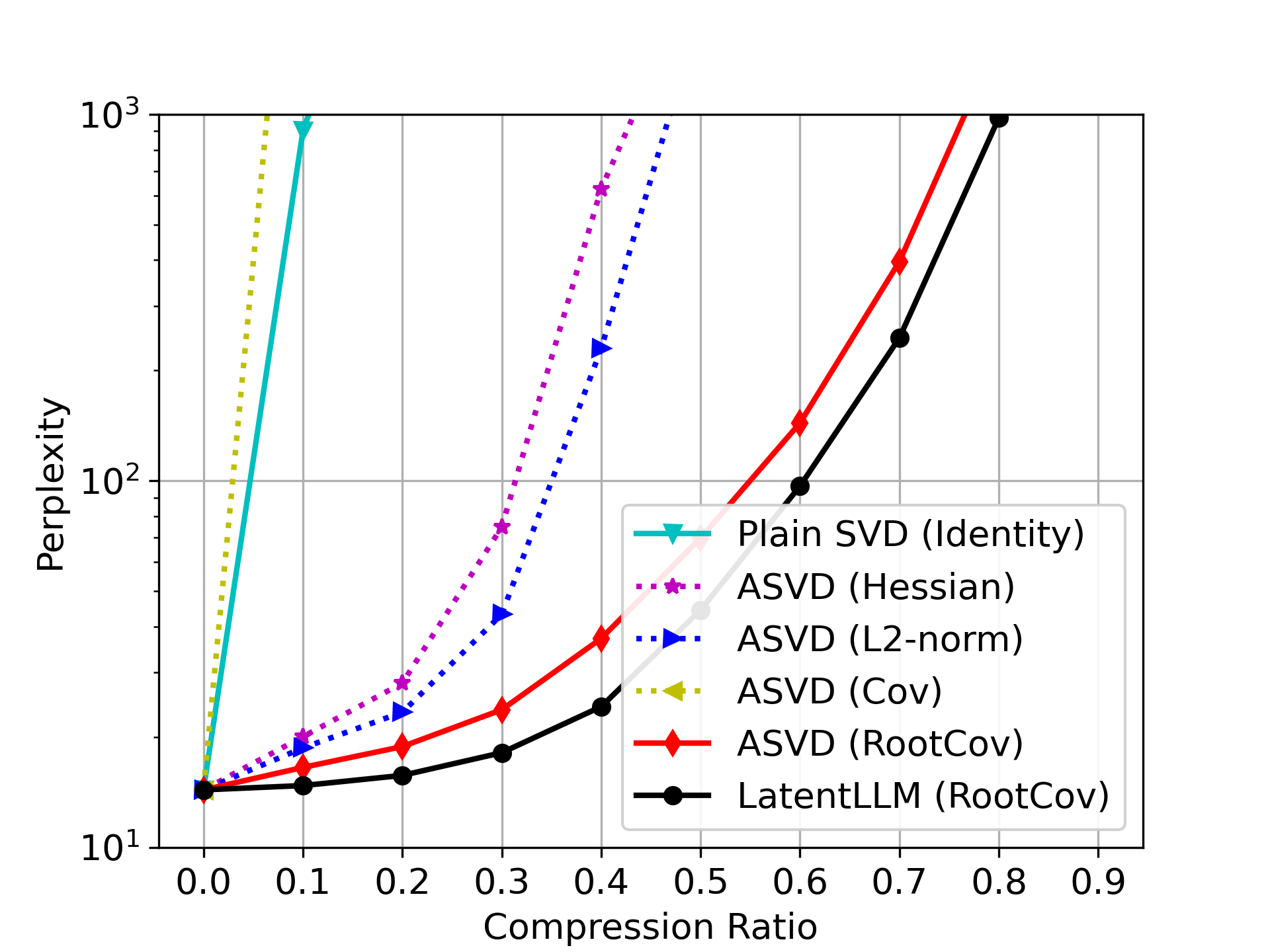}
\caption{C4 (OPT-2.7B)}
\end{subfigure}

\begin{subfigure}[b]{0.32\linewidth}
\includegraphics[width=\linewidth,,trim=10 0 40 30,clip]{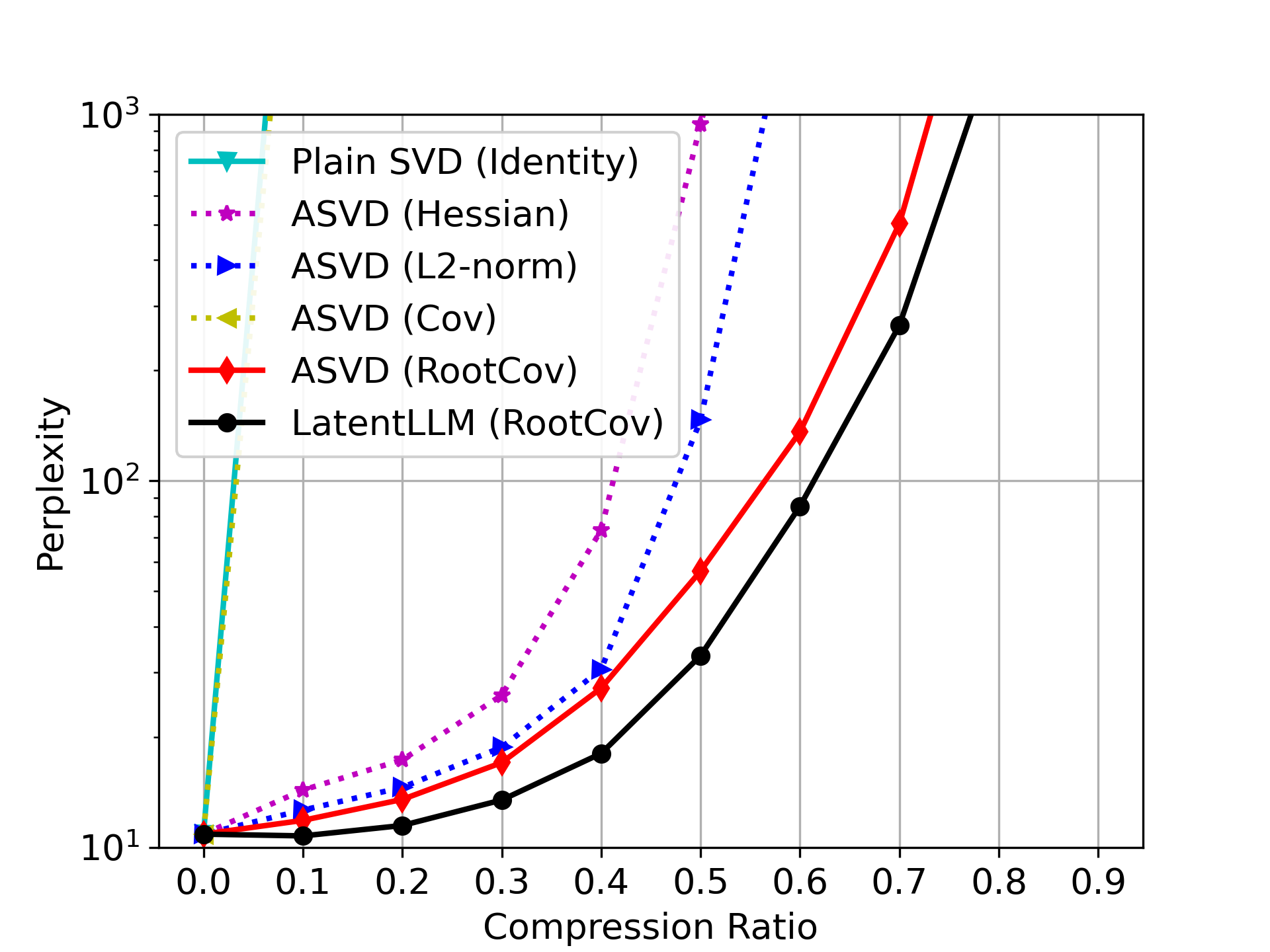}
\caption{WT2 (OPT-6.7B)}
\end{subfigure}
\begin{subfigure}[b]{0.32\linewidth}
\includegraphics[width=\linewidth,,trim=10 0 40 30,clip]{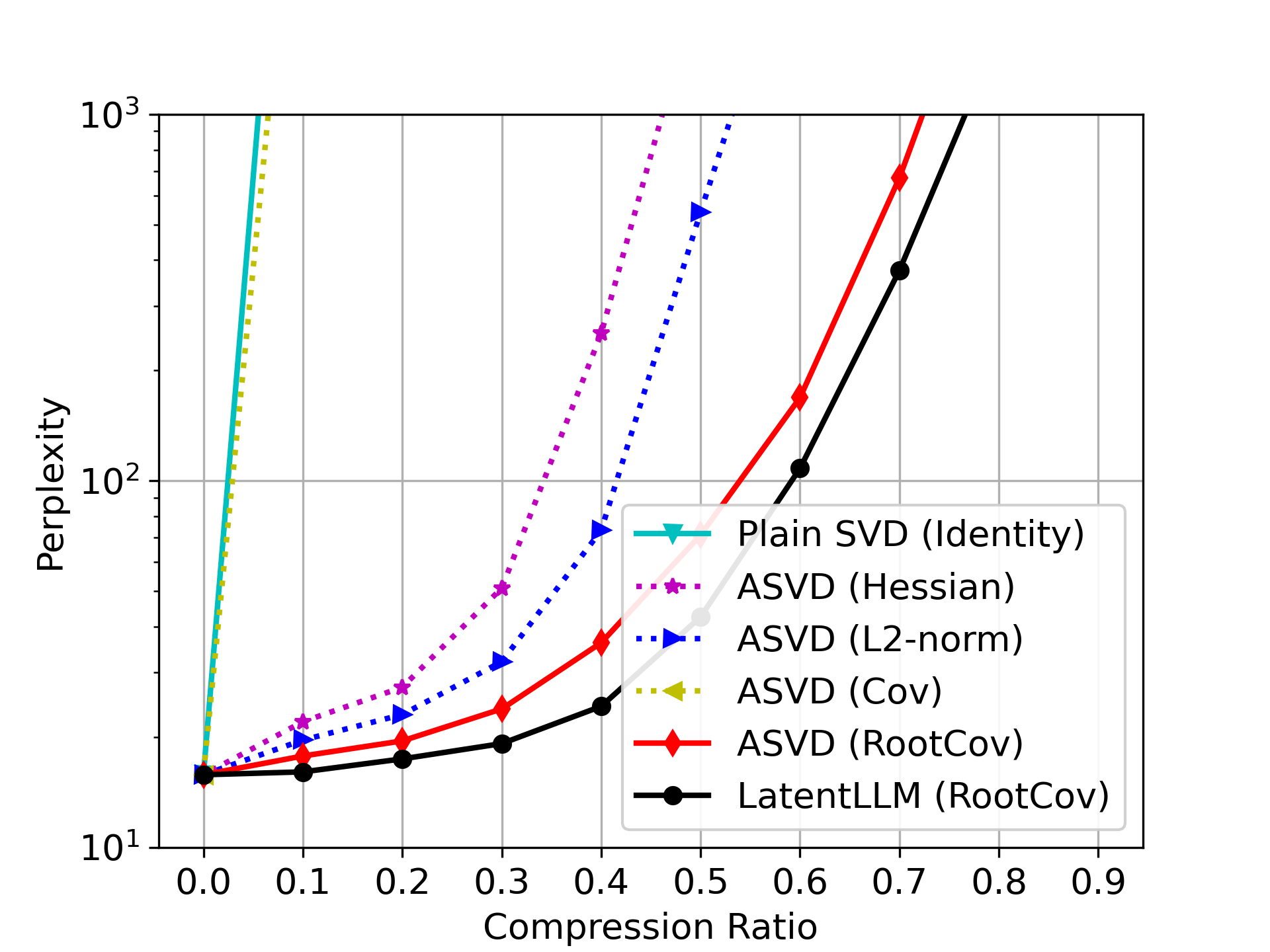}
\caption{PTB (OPT-6.7B)}
\end{subfigure}
\begin{subfigure}[b]{0.32\linewidth}
\includegraphics[width=\linewidth,,trim=10 0 40 30,clip]{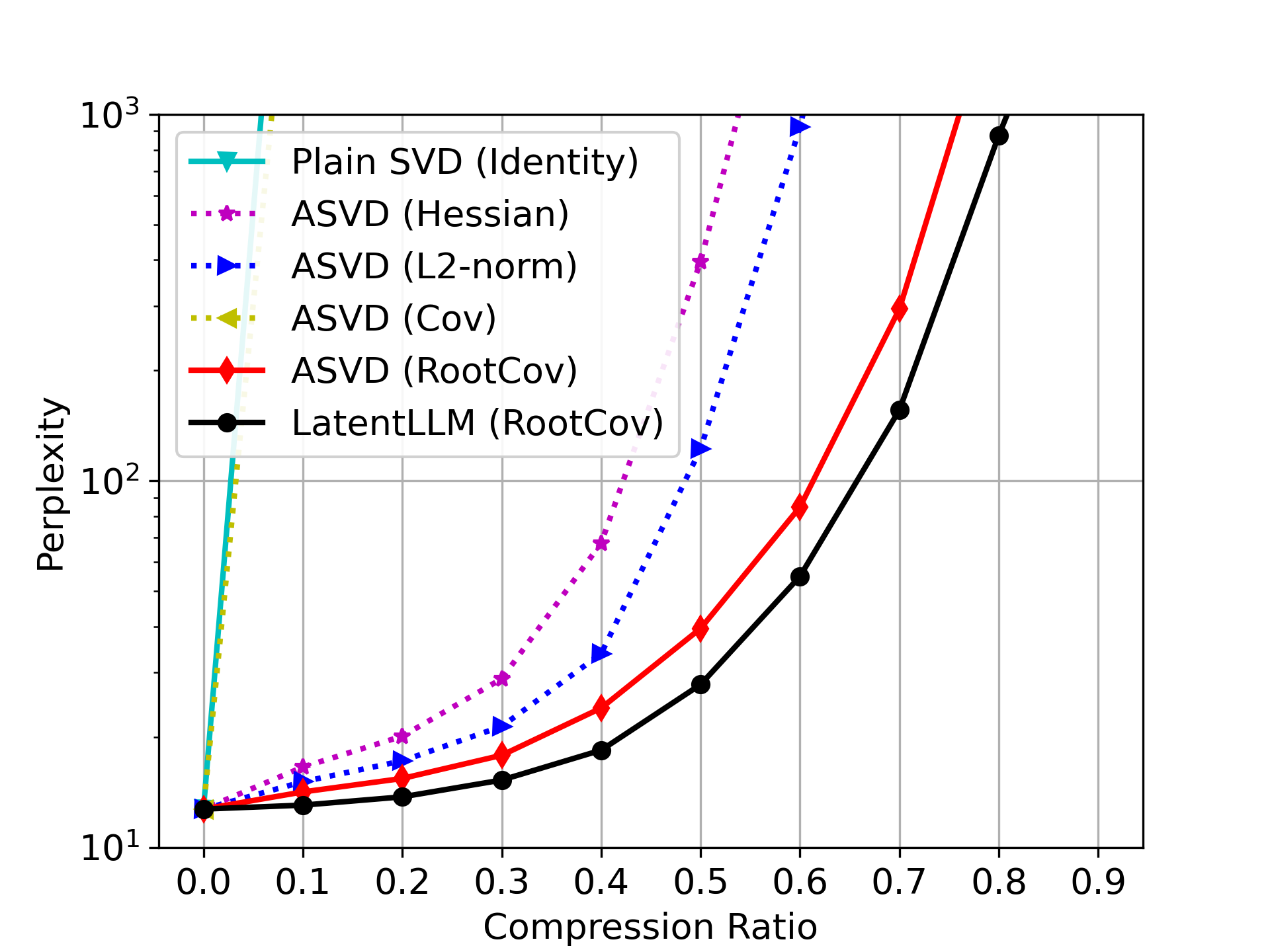}
\caption{C4 (OPT-6.7B)}
\end{subfigure}

\caption{Perplexity over compression ratio for OPT models. }
\label{fig:perp_opt}
\end{center}
\vskip -0.2in
\end{figure*}

\paragraph{Models and datasets} 

For LLM experiments, we consider the OPT model family~\cite{zhang2022opt} as it provides a wide range of model scales from 125M to 175B as noted in \cref{sec:model}. 
We show results on different sizes of models to provide a broader picture for the performance of
LatentLLM. 
In terms of metrics, we mainly focus on perplexity, which is known to be a challenging
and stable metric that is well-suited for evaluating the accuracy of compression methods~\cite{yao2022zeroquant, dettmers2023case}. 
We consider the test sets of raw-WikiText2 (WT2)~\cite{merity2016pointer} and the Penn Treebank (PTB)~\cite{marcus1994penn} as well as a subset of the C4 validation data, all
popular benchmarks in the LLM compression literature~\cite{frantar2023sparsegpt, frantar2022gptq, sun2023simple}. 

For LMM experiments, we use LLaVa~\cite{liu2023llava}, specifically \texttt{liuhaotian/llava-v1.6-vicuna-7b} model, which consists of language transformer based on Vicuna and vision transformer (ViT) based on the contrastive language-image pre-training (CLIP).
We evaluate the capability of the multi-modal answer reasoning based on the ScienceQA dataset,
which contains 21K vision-language multi-choice questions for three subjects: natural, social, and language science.
Some fractions of questions have image and/or text contexts, and the problem levels range from grade 1 to 12.

\paragraph{Baseline methods}
We compare our LatentLLM against several baselines: weight-based plain SVD compression applied locally, and several variants of local ASVD compression with different pre-conditioning matrix as listed in \cref{tab:cond}. 

\paragraph{Computational complexity}

When all linear modules are compressed with LatentLLM, the inference complexity is expected to be reduced with the compression ratio almost linearly.
Nevertheless LLMs/VLMs have extra complexity other than linear affine transforms, the inference complexity is not precisely proportional to the compression factor. 
We show the complexity analysis in \cref{tab:flops} for the compressed OPT-6.7B models, based on the \texttt{calflops}\footnote{\url{https://pypi.org/project/calflops/}} library.
We assume the token length of 128 for FLOPs analysis.
We found that the FLOPs, multiply-accumulation operations (MACs), and parameters are almost linearly reduced with the compression factor.

\begin{table}[t]
\centering
\caption{Computational complexity of OPT-6.7B models compressed by LatentLLM.}
\label{tab:flops}
\setlength{\tabcolsep}{1.5pt}
\small
\begin{tabular}{r rr rr rr}
\toprule
Compression & \multicolumn{2}{c}{FLOPs} & \multicolumn{2}{c}{MACs} & \multicolumn{2}{c}{Parameters} \\
\midrule
0\% & 
1.70T & \progressbar{100}[1.2][red!70] & 
851G & \progressbar{100}[1.2][red!70] & 
6.66B & \progressbar{100}[1.2][red!70] \\
10\% & 
1.53T & \progressbar{90}[1.2][red!70] & 
766G & \progressbar{90.01}[1.2][red!70] & 
6.20B & \progressbar{93.09}[1.2][red!70] \\
20\% & 
1.36T & \progressbar{80}[1.2][red!70] & 
681G & \progressbar{80.02}[1.2][red!70] & 
5.53B& \progressbar{83.03}[1.2][red!70] \\
30\% & 
1.19T & \progressbar{70}[1.2][red!70] & 
596G & \progressbar{70.04}[1.2][red!70] & 
4.87B & \progressbar{73.12}[1.2][red!70]\\
40\% & 
1.02T & \progressbar{60}[1.2][red!70] & 
511G & \progressbar{60.04}[1.2][red!70] & 
4.20B & \progressbar{63.06}[1.2][red!70] \\
50\% & 
851G & \progressbar{50.06}[1.2][red!70] & 
425G & \progressbar{49.94}[1.2][red!70] & 
3.54B & \progressbar{53.15}[1.2][red!70] \\
60\% &  
681G & \progressbar{40.06}[1.2][red!70] & 
340G & \progressbar{39.95}[1.2][red!70] & 
2.87B & \progressbar{43.09}[1.2][red!70] \\
70\% & 
511G & \progressbar{30.06}[1.2][red!70] & 
255G & \progressbar{29.96}[1.2][red!70] & 
2.21B & \progressbar{33.18}[1.2][red!70] \\
80\% & 
341G & \progressbar{20.06}[1.2][red!70] & 
170G & \progressbar{19.98}[1.2][red!70] & 
1.54B & \progressbar{23.12}[1.2][red!70] \\
90\% & 
171G & \progressbar{10.06}[1.2][red!70] & 
85.2G & \progressbar{9.69}[1.2][red!70] & 
880M & \progressbar{13.21}[1.2][red!70] \\
\bottomrule
\end{tabular}
\end{table}

\begin{figure*}[t]
\centering
\begin{subfigure}[b]{0.32\linewidth}
\includegraphics[width=\linewidth,,trim=10 0 40 30,clip]{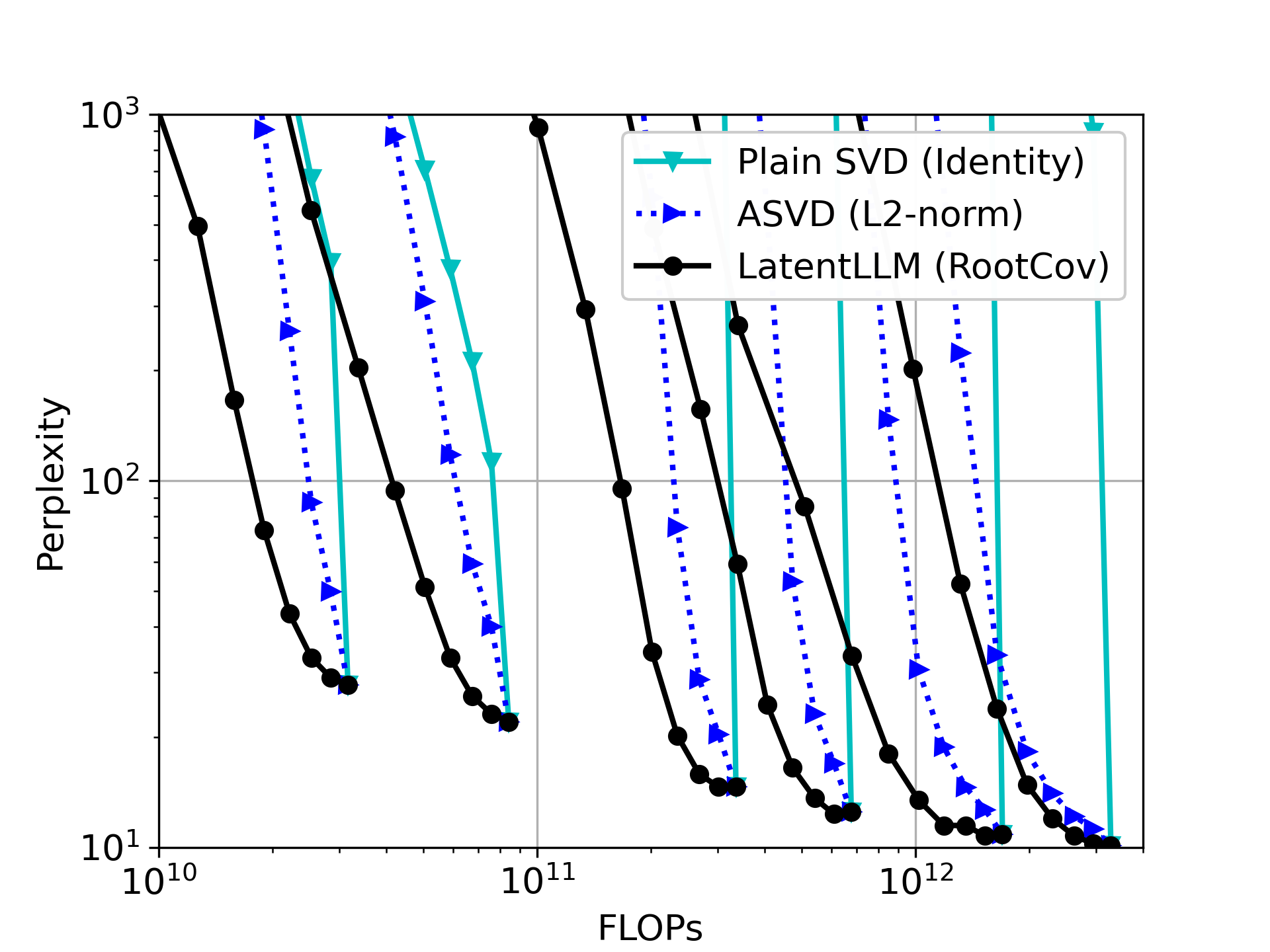}
\caption{WT2}
\end{subfigure}
\begin{subfigure}[b]{0.32\linewidth}
\includegraphics[width=\linewidth,,trim=10 0 40 30,clip]{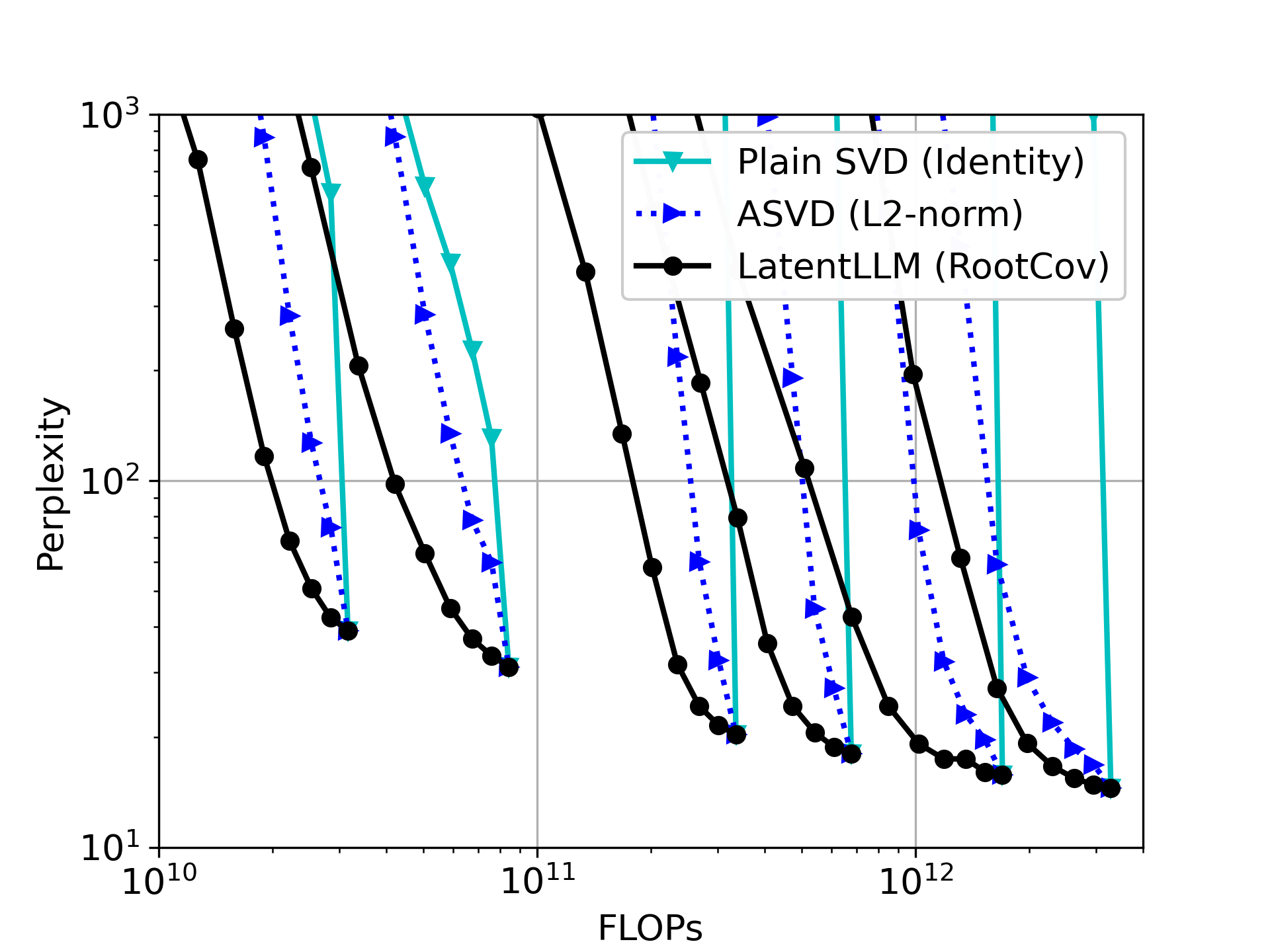}
\caption{PTB}
\end{subfigure}
\begin{subfigure}[b]{0.32\linewidth}
\includegraphics[width=\linewidth,,trim=10 0 40 30,clip]{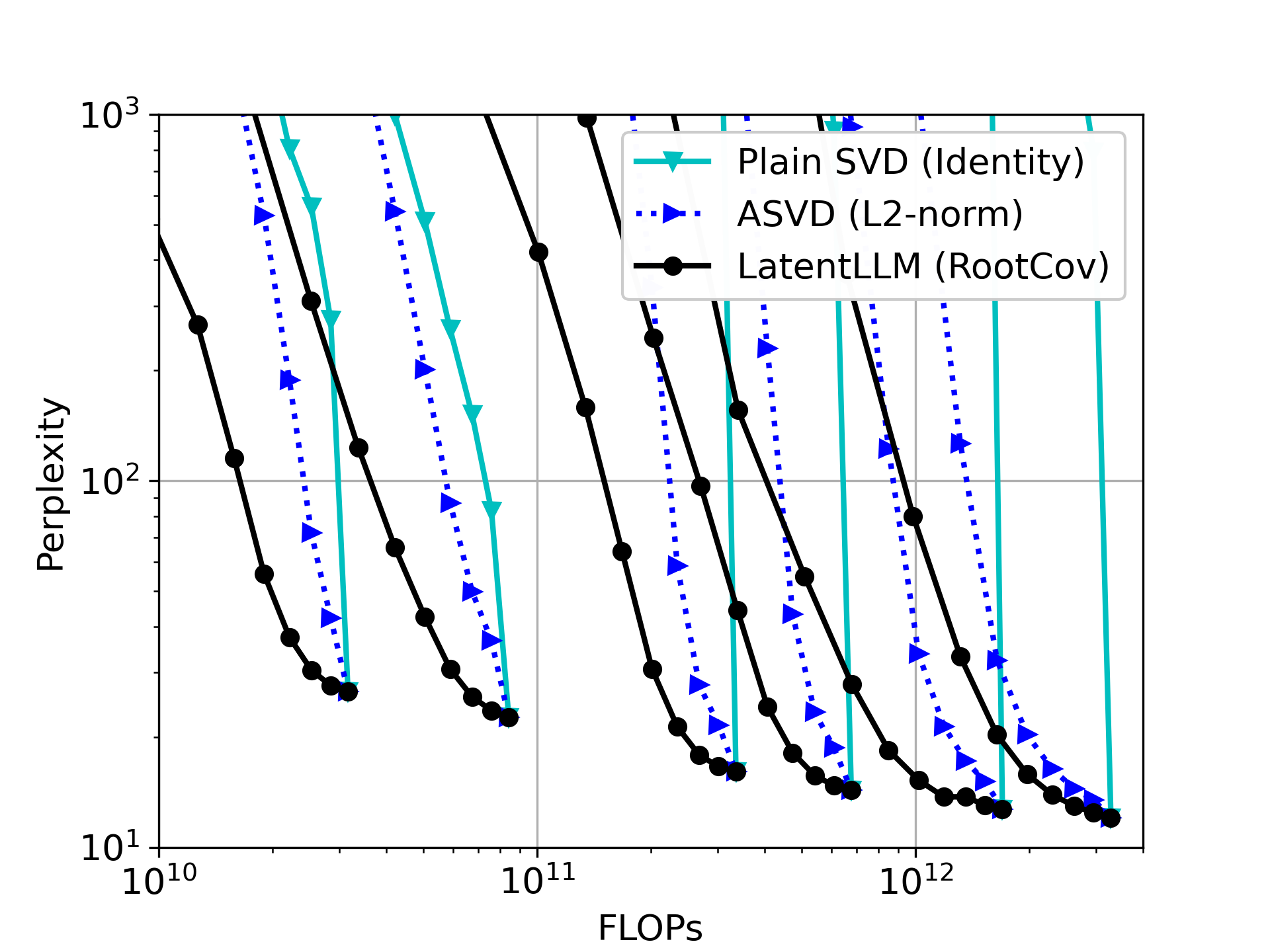}
\caption{C4}
\end{subfigure}

\caption{Perplexity vs.\ FLOPs of compressed OPT for 125M to 13B models.}
\label{fig:flops}
\end{figure*}

\paragraph{Compression over model size}

We first look into the compression capabilities of our LatentLLM across various model sizes in comparison to baseline methods. 
Detailed results are shown for a size reduction over $10$--$40\%$ in \cref{tab:perp_opt}, and  wider visualization in \cref{fig:perp_opt}.
The perplexity results of the original un-compressed LLM models are reported next to the names of the models in the table.

We can see that the conventional plain SVD has a poor performance, and that ASVD with a proper pre-conditioning can significantly improve the perplexity.
It was found that the diagonal Hessian is worse than the diagonal $\ell_2$-norm, whereas covariance pre-conditioning can be terrible in low compression regimes for larger LLMs.
In contrast, the superiority of root covariance is remarkable.
More importantly, our block identity transform is considerably beneficial to compress LLMs.
In addition, the joint SVD used for LatentLLM  offers an additional improvement for the most cases.
For LatentLLM, we used $4$ iterations for joint UD compression and $8$ iterations for joint QK compression.
Notice that our methods can also achieve slightly better performance than the original un-compressed LLMs for a few cases.

\begin{figure*}[t]
\centering
\begin{subfigure}[b]{0.32\linewidth}
\includegraphics[width=\linewidth,trim=0 0 60 20,clip]{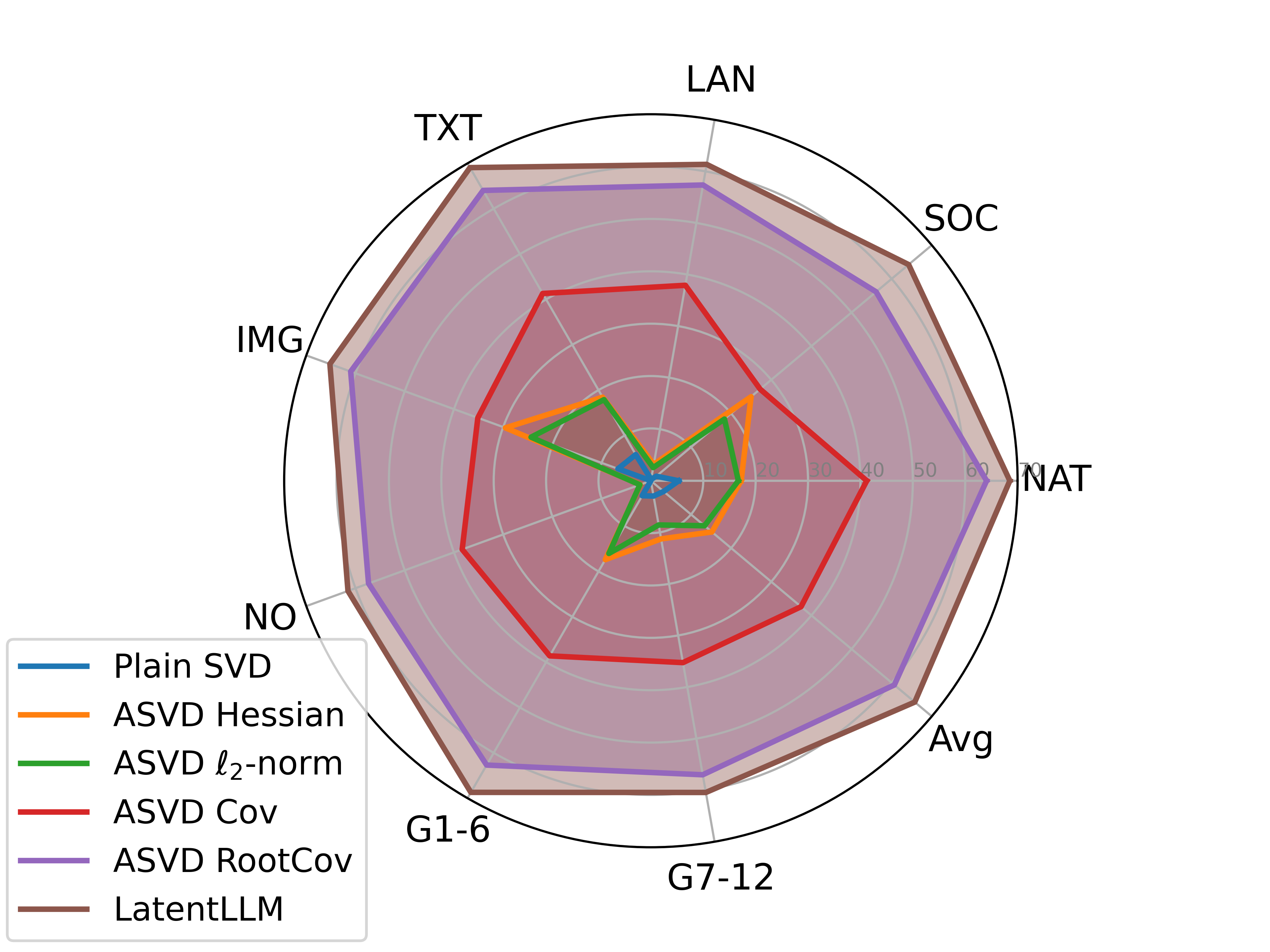}
\caption{10\% Compression}
\end{subfigure}
\begin{subfigure}[b]{0.32\linewidth}
\includegraphics[width=\linewidth,trim=0 0 60 20,clip]{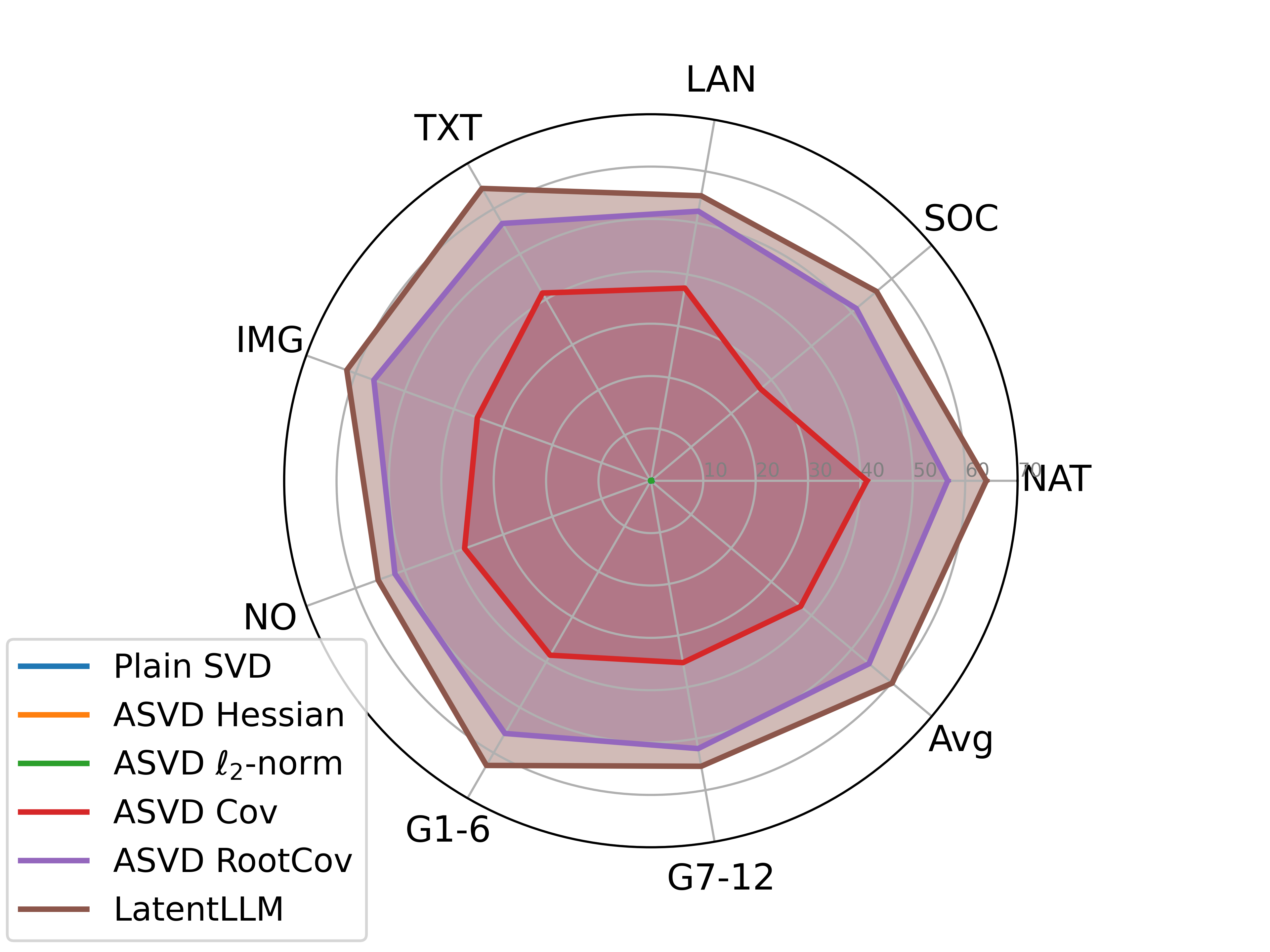}
\caption{30\% Compression}
\end{subfigure}
\begin{subfigure}[b]{0.32\linewidth}
\includegraphics[width=\linewidth,trim=0 0 60 20,clip]{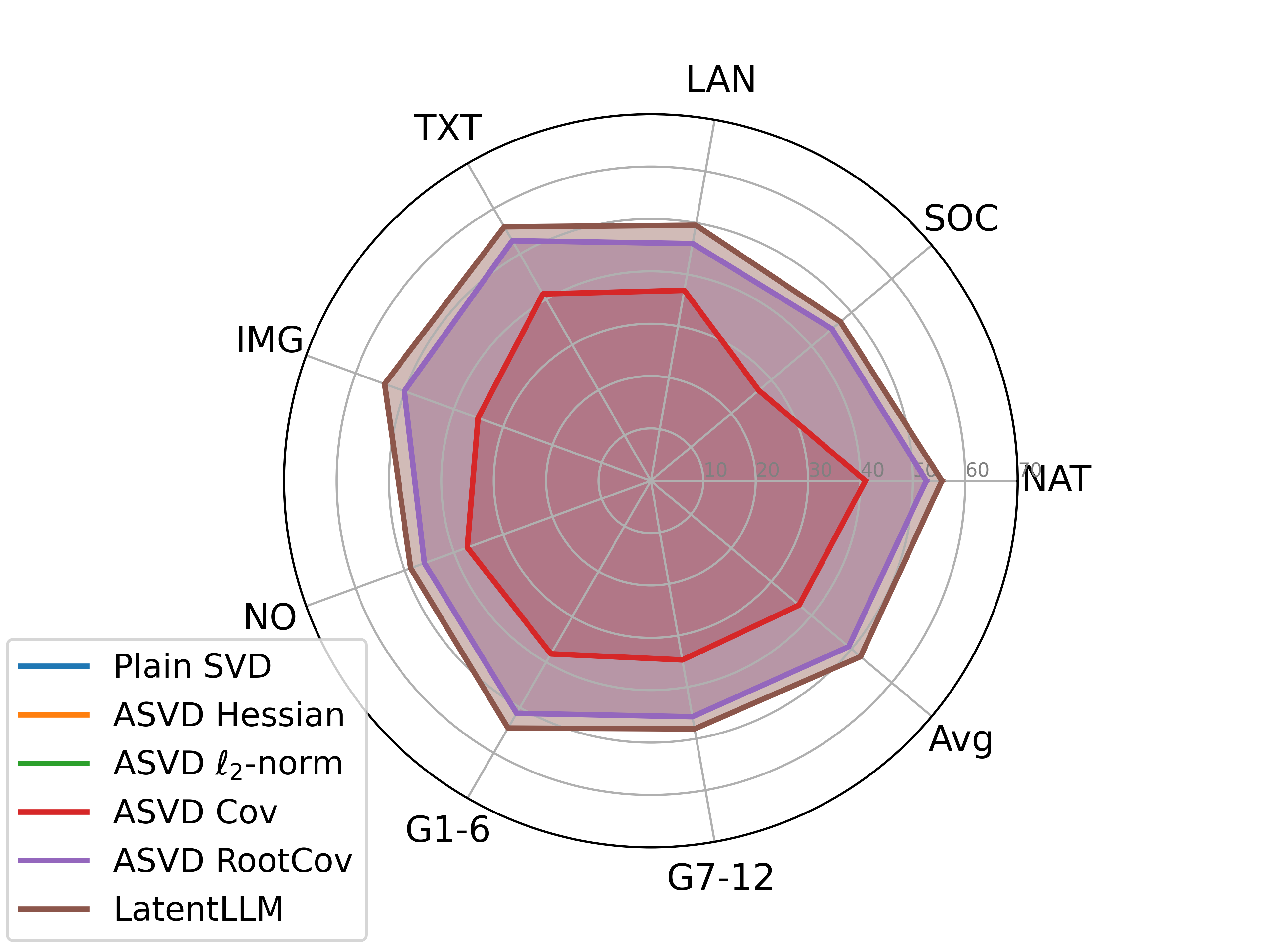}
\caption{50\% Compression}
\end{subfigure}
\caption{Radar plots of ScienceQA accuracy results across different subjects, context modalities, and grades at $10\%$--$50\%$ compressions.}
\label{fig:sqa}
\end{figure*}

\paragraph{Multi-modal reasoning capability}
We show the accuracy of latent LLaVa models for ScienceQA multi-modal reasoning benchmark in \cref{tab:sqa_llava}.
It is verified that our LatentLLM can significantly outperform other low-rank compression methods across diverse reasoning problems over different subjects/contexts/grades, approaching the performance of the original un-compressed LLaVa model.
\cref{fig:sqa} shows the corresponding radar plots at $10\%$--$50\%$ compressions.
It is seen that ASVD without using proper pre-donditioning matrix degrades the performance quickly with higher compression ratios, while our LatentLLM keeps relatively higher performance across all cases.
Also, we observe that context-less, language science, and higher grade questions tend to have lower performance.

\begin{table*}[t]
\centering
\caption{Accuracy in percent ($\uparrow$) on ScienceQA dataset of LLaVa model with different compression methods for $10\%$--$50\%$ size reduction.
Question
subjects: natural science (NAT); social science (SOC); language science (LAN).
Context modality: text (TXT);
image (IMG); or no context (NO).
Grades: 1--6 (G1-6); 7--12 (G7-12).
}
\label{tab:sqa_llava}
\small
\begin{tabular}{lc rrr rrr rr r}
  \toprule
   & & \multicolumn{3}{c}{Subject} & \multicolumn{3}{c}{Context Modality} & \multicolumn{2}{c}{Grades} \\
  \cmidrule(lr){3-5}
  \cmidrule(lr){6-8}
  \cmidrule(lr){9-10}
  Method & Compression &
  NAT & SOC & LAN &
  TXT & IMG & NO &
  G1-6 & G7-12 &
  Avg \\
  \midrule
  Original un-compressed & 0\% &
  72.47 & 69.18 & 65.73 & 
  73.51 & 68.82 & 65.99 &
  72.72 & 65.19 & 70.03
  \\
  \midrule 
  Plain SVD (Identity) & 10\% &
  5.33 & 1.35 & 0.27 & 
  5.77 & 6.69 & 0.00 &
  3.30 & 2.97 & 3.18 \\
  ASVD (Hessian) &  10\% &
  17.23 & 24.97 & 3.18 &
  18.43 & 29.55 & 2.16 &
  17.40 & 11.27 & 15.21 \\
  ASVD ($\ell_2$-norm) &  10\% &
  16.70 & 18.34 & 2.55 &
  17.89 & 24.34 & 2.23 & 
  16.04 & 8.57 & 13.37 \\
  ASVD (Cov) &  10\% &
  41.21 & 27.22& 37.91 & 
  41.30 & 35.15 & 38.33 &
  38.62 & 35.27 & 37.42 \\
  ASVD (RootCov) &  10\% &
  64.08 & 56.13 & 57.36 &
  64.03 & 60.98 & 57.35 & 
  62.70 & 57.02 & 60.67 \\
  \rowcolor{lgreen}
  LatentLLM (RootCov) &  10\% &
  \bf{68.52} & \bf{64.23} & \bf{61.36} &
  \bf{69.06} & \bf{65.20} & \bf{61.53} &
  \bf{68.72} & \bf{60.45} & \bf{65.76} \\
  \midrule
  Plain SVD (Identity) & 20\% &
  0.18 & 0.00 & 0.00 &
  0.20 & 0.20 & 0.00 & 
  0.04 & 0.20 & 0.09 \\
  ASVD (Hessian) &  20\% &
  3.82 & 2.81 & 0.00 &
  3.62 & 5.30 & 0.14 &
  3.01 & 1.91 & 2.62 \\
  ASVD ($\ell_2$-norm) &  20\% &
  0.44 & 0.79 & 0.00 &
  0.39 & 0.79 & 0.07 &
  0.51 & 0.20 & 0.40 \\
  ASVD (Cov) &  20\% &
  41.39 & 27.22 & 37.55 &
  41.45 & 35.35 & 38.12 & 
  38.69 & 35.14 & 37.42 \\
  ASVD (RootCov) &  20\% &
  61.19 & 53.43 & 53.36 &
  61.53 & 59.40 & 52.68 &
  58.96 & 54.98 & 57.53 \\
  \rowcolor{lgreen}
  LatentLLM (RootCov) &  20\% &
  \bf{66.39} & \bf{61.19} & \bf{60.82} &
  \bf{67.20} & \bf{63.41} & \bf{60.62} &
  \bf{66.41} & \bf{59.26} & \bf{63.85} \\
  \midrule
  Plain SVD (Identity) & 30\% &
  0.13 & 0.00 & 0.00 &
  0.10 & 0.00 & 0.07 & 
  0.11 & 0.00 & 0.07 \\
  ASVD (Hessian) &  30\% &
  0.00 & 0.00 & 0.00 &
  0.00 & 0.00 & 0.00 &
  0.00 & 0.00 & 0.00 \\
  ASVD ($\ell_2$-norm) &  30\% &
  0.09 & 0.00 & 0.00 &
  0.10 & 0.10 & 0.00 &
  0.04 & 0.07 & 0.05 \\
  ASVD (Cov) &  30\% &
  41.25 & 27.33 & 37.36 &
  41.40 & 35.25 & 37.84 & 
  38.47 & 35.27 & 37.33 \\
  ASVD (RootCov) & 30\% &
  56.66 & 51.18 & 52.27 & 
  56.74 & 56.27 & 51.99 & 
  55.73 & 51.94 & 54.37 \\
  \rowcolor{lgreen}
  LatentLLM (RootCov) &  30\% &
  \bf{64.03} & \bf{56.24} & \bf{55.27} &
  \bf{64.47} & \bf{61.77} & \bf{55.40} &
  \bf{62.78} & \bf{55.37} & \bf{60.13} \\
  \midrule
  Plain SVD (Identity) & 40\% &
  0.00 & 0.00 & 0.00 &
  0.00 & 0.00 & 0.00 & 
  0.00 & 0.00 & 0.00 \\
  ASVD (Hessian) &  40\% &
  0.04 & 0.00 & 0.54 &
  0.05 & 0.00 & 0.42 &
  0.07 & 0.33 & 0.17 \\
  ASVD ($\ell_2$-norm) &  40\% &
  0.00 & 0.00 & 0.27 &
  0.00 & 0.00 & 0.21 &
  0.07 & 0.07 & 0.07 \\
  ASVD (Cov) &  40\% &
  40.94 & 27.22 & 36.91 &
  41.06 & 35.10 & 37.49 & 
  38.25 & 34.81 & 37.02 \\
  ASVD (RootCov) & 40\% &
  55.37 & 49.04 & 48.36 & 
  55.62 & 54.54 & 48.50 & 
  54.19 & 48.71 & 52.23 \\
  \rowcolor{lgreen}
  LatentLLM (RootCov) &  40\% &
  \bf{56.62} & \bf{51.07} & \bf{53.27} &
  \bf{56.74} & \bf{56.12} & \bf{52.47} &
  \bf{55.51} & \bf{52.93} & \bf{54.59} \\
  \midrule
  Plain SVD (Identity) & 50\% &
  0.00 & 0.00 & 0.00 &
  0.00 & 0.00 & 0.00 & 
  0.00 & 0.00 & 0.00 \\
  ASVD (Hessian) &  50\% &
  0.00 & 0.00 & 0.00 &
  0.00 & 0.00 & 0.00 &
  0.00 & 0.00 & 0.00 \\
  ASVD ($\ell_2$-norm) &  50\% &
  0.00 & 0.00 & 0.00 &
  0.00 & 0.00 & 0.00 &
  0.00 & 0.00 & 0.00 \\
  ASVD (Cov) &  50\% &
  40.94 & 26.88 & 36.91 &
  41.20 & 35.10 & 37.28 & 
  38.18 & 34.74 & 36.95 \\
  ASVD (RootCov) & 50\% &
  52.58 & 45.11 & 46.00 & 
  52.93 & 50.07 & 45.99 & 
  51.28 & 45.75 & 49.30 \\
  \rowcolor{lgreen}
  LatentLLM (RootCov) &  50\% &
  \bf{55.55} & \bf{47.24} & \bf{49.55} &
  \bf{56.01} & \bf{54.09} & \bf{48.78} &
  \bf{54.55} & \bf{48.12} & \bf{52.25} \\
  \bottomrule
\end{tabular}
\end{table*}

\paragraph{Discussion}

Our framework with optimal pre-conditioning and joint Q/K, V/O distillations can be readily applied to pruning and quantization as well.
See some results and discussion in \cref{sec:spa} and \cref{sec:quant}.
Further fine-tuning is expected to be able to compensate for the performance degradation by the latent reduction.

\section{Summary}

We introduced LatentLLM which jointly compresses multiple tensors through the use of high-order tensor-rank decomposition.
We also provided some new perspectives for activation-aware compression when choosing the pre-conditioner and junction matrix.
With a proper selection, we demonstrated that the model compression performance can be significantly improved.
Our latent LLaVa showed a significant advantage in multi-modal reasoning tasks compared to other baseline methods.

{
    \small
    \bibliographystyle{ieeenat_fullname}
    \bibliography{ref}
}

\clearpage
\appendix
\onecolumn

\section{Weight-Aware Compression}
\label{sec:weight}

\subsection{Plain SVD}
Given a pretrained weight matrix $W\in\mathbb{R}^{d'\times d}$, we wish to approximate it with a low-rank structure:
\begin{align}
\hat{W} = B \times A,
\end{align}
where $\hat{W}\in\mathbb{R}^{d'\times d}$ is an approximated weight, $B\in\mathbb{R}^{d'\times r}$ and $A\in\mathbb{R}^{r\times d}$ are low-rank matrices with a rank $r\leq d,d'$.
We assume $d'\leq d$ for simplicity, as modifying for $d'\geq d$ is straightforward.

Consider the approximation loss:
\begin{align}
\mathcal{L} &= 
\| W - \hat{W} \|^2 \\
&=
\| W - BA \|^2. 
\label{eq:l2_w}
\end{align}

The best solution is given by SVD of $W$ as follows:
\begin{align}
A &= J^+ V,
\label{eq:a_w}\\
B &= U S J,
\label{eq:b_w}
\end{align}
where $U\in \mathbb{R}^{d'\times r}$ is $r$ most-principal left-singular vectors, $S=\mathsf{diag}[\sigma_1, \ldots, \sigma_{r}]\in\mathbb{R}^{r\times r}$ is diagonal singular-values, and $V\in\mathbb{R}^{r\times d}$ is the most-principal right-singular vectors for $W$:
\begin{align}
U S V &=\mathsf{svd}_r[W],
\label{eq:svd_w}
\end{align}
where we assume the singular values are sorted in the descending order: $\sigma_1 \geq \sigma_2 \geq \cdots \geq \sigma_{r}$.
The loss is the accumulation of all the squared singular-values outside the rank $r$: $\mathcal{L}_{\min} = \sum_{i>r} \sigma_i^2$.

\subsection{Junction Matrix}
\label{sec:junc}

There is no literature discussing the choice of a junction matrix $J\in\mathbb{R}^{r\times r}$, which has no impact on performance, provided that $S J J^+ = S$ is satisfied.
There are many suitable choices for this matrix $J$, such as:
\begin{itemize}
\item Left singular: $J=I$;
\item Right singular:
$J=S^+$;
\item Symmetry singular:
$J=[S^\frac{1}{2}]^+$.
\item Left block identity:
$J=[US]_{:r}^+$
\item Right block identity:
$J=[V]_{:r}$
\end{itemize}
Although there is no performance impact by the choice of $J$, it is notable that the block identity which is based on block LU factorization can significantly reduce the number of parameters and FLOPs by $r^2$.
This parameter reduction is particularly significant in high-dimensional (high-rank) latent cases.
For example, when the weight is a size of $d=d'=2048$, even with the half-rank latent $r=d/2=1024$, there is no parameter reduction as the dense compression and decompression matrices $B$ takes $2dr=d^2$ parameters.
Hence, the $50\%$ latent has no benefit in complexity but only for KV cache memory reduction as discussed in DeepSeek-V3~\cite{liu2024deepseek}.
It is even worse for $r>d/2$: e.g., if we use $75\%$ latent of $r=0.75d=1536$, the total parameter and FLOPs increases by 50\% of the original weight (i.e., $2rd-d^2=d^2/2$).
However, using the block identity form, we can save $r^2$, and the total FLOPs will be always less  than the original weight: $2rd-r^2 < d^2$ for any $r <d$.

\section{Activation-Aware Compression}
\label{sec:act}

Consider an input token $X\in\mathbb{R}^{d\times l}$ for a context length $l\gg d$, the linear projection output $Y\in\mathbb{R}^{d'\times l}$ is:
\begin{align}
Y = W X.
\end{align}
We assume that no bias is used for simplicity.

We wish to minimize the expected approximation error of output activation vectors between the true $Y$ and the approximation $\hat{Y}\in\mathbb{R}^{d'\times l}$: 
\begin{align}
 \hat{Y} = \hat{W} X,   
\end{align}
projected by a low-rank weight $\hat{W}=BA$.
Consider the loss function:
\begin{align}
    \mathcal{L} &=    
    \mathbb{E}_X 
    \| Y - \hat{Y} \|^2
    \\
    &=
    \mathbb{E}_X
    \|
    (W-\hat{W}) X
    \|^2
    \\
    &=
    \mathbb{E}_X
    \|
    (W-BA) X
    \|^2
    \\
    &=
    \mathrm{tr}
    [
    (W-BA)\mathbb{E}_X[XX^\top]
    (W-BA)^\top
    ]
    \\
    &=
    \mathrm{tr}
    [
    (W-BA)C
    (W-BA)^\top
    ]
    \\
    &=
    \|WC^\frac{1}{2} - BAC^\frac{1}{2}\|^2,
    \label{eq:l2_act}
\end{align}
where $C=\mathbb{E}_X[XX^\top]\in\mathbb{R}^{d\times d}$ is a (positive semidefinite) correlation matrix of the input vector.

The loss function of the activation-aware distillation in (\ref{eq:l2_act}) is identical to the weight-aware distillation in (\ref{eq:l2_w}) by transforming as:
\begin{align}
    W &\rightarrow W'= WC^\frac{1}{2},\\
    A &\rightarrow A'=AC^\frac{1}{2}.
\end{align}
Hence, the optimal solution is obtained similarly.

Specifically, from (\ref{eq:a_w}), (\ref{eq:b_w}), and (\ref{eq:svd_w}), we have
\begin{align}
A'=AC^\frac{1}{2} &= J^+ 
V',\label{eq:a_act}\\
B &= U' S' J,
\label{eq:b_act}
\end{align}
where $U'\in \mathbb{R}^{d'\times d'}$, $S'\in\mathbb{R}^{d'\times d}$, and $V'\in\mathbb{R}^{d\times d}$ are SVD of $W'=WC^\frac{1}{2}$:
\begin{align}
U' S' V' &=\mathsf{svd}[WC^\frac{1}{2}].
\label{eq:svd_act}
\end{align}
From (\ref{eq:a_act}), we finally obtain optimal $A$ as:
\begin{align}
    A &= J^+
    S' V' [C^{\frac{1}{2}}]^+.
\end{align}

\subsection{Pre-Conditioning Matrix}
\label{sec:cond}

ASVD~\cite{yuan2023asvd} proposed to use a projection matrix $P\in\mathbb{R}^{d\times d}$ on weights before SVD: $\mathsf{svd}[WP]$.
The optimal projection $P$ is apparently the square-root covariance $P=C^\frac{1}{2}$, while there are many other approximated projections that were considered in literature:
\begin{itemize}
    \item Root-covariance: $P=(XX^\top+\lambda I)^\frac{1}{2}$

    \item Covariance (e.g., CorDA~\cite{yang2024corda}):
    $P = XX^\top$

    \item Diagonal L2-norm (e.g., WandA~\cite{sun2023simple}):
    $P=\mathrm{diag}[XX^\top]^\frac{1}{2}$

    \item Diagonal L1-norm (e.g., AWQ~\cite{lin2024awq}, ASVD~\cite{yuan2023asvd}):
    $P=\mathrm{diag}[\|[X]_{1,:}\|_1, \ldots, \|[X]_{d,:}\|_1]$

    \item Diagonal Hessian (e.g., OBS~\cite{hassibi1993optimal}, GPTQ~\cite{frantar2022gptq}, SparseGPT~\cite{frantar2023sparsegpt}):
    $P=\mathrm{diag}[(XX^\top + \lambda I)^{-1}]^\frac{-1}{2}$

    \item Identity (Plain SVD, e.g., \cite{sainath2013low}): $P=I$
\end{itemize}

\begin{figure}[t]
\centering
\includegraphics[width=0.8\linewidth]{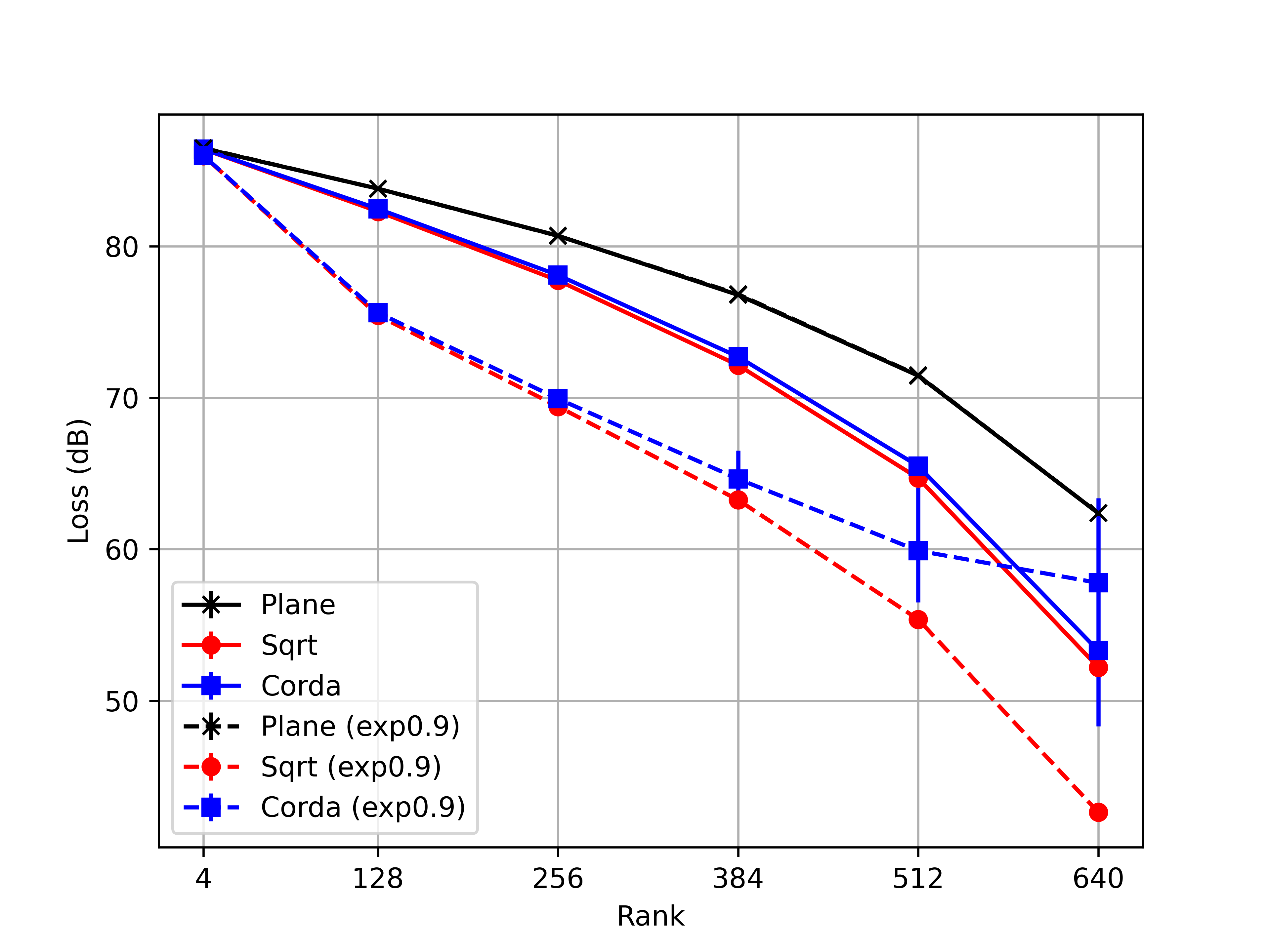}
\caption{Comparison of SVD, CorDA, and RootCorDA.}
\label{fig:corda}
\end{figure}

In the context of fine-tuning initialization, CorDA~\cite{yang2024corda} uses covariance matrix $C$ without square root, which should be worse than the square-root covariance.
Fig.~\ref{fig:corda} demonstrates the benefit for random weights approximation with covariance drawn from the Wishart distribution.
GPTQ~\cite{frantar2022gptq} and SparseGPT~\cite{frantar2023sparsegpt} use another preconditioning matrix based on optimal brain surgeon (OBS)~\cite{hassibi1993optimal} using Hessian, in the context of quantization and pruning.
Similarly, we can use it for low-rank compression as we have evaluated.
In the context of pruning, WandA~\cite{sun2023simple} proposed a simpler diagonal projection based on $\ell_2$-norm, while it achieves an excellent performance.
AWQ and ASVD used the diagonal $\ell_1$-norm, while the theoretical justification is missing. 
They introduced another exponent factor to adjust.

\subsection{Bias Update}
\label{sec:bias1}

When the bias is there, we have
\begin{align}
    \mathcal{L} &= \| (WX+b 1^\top) - (BAX+\hat{b}1^\top) \|^2.
\end{align}
The gradient with respect to $\hat{b}$ is
\begin{align}
    \nabla \hat{b} &=
    -2 ((WX+b 1^\top) - (BAX+\hat{b}1^\top)) 1.
\end{align}
Hence, the optimal bias modification is:
\begin{align}
    \hat{b} &=
    b + (W-BA)\mu,
\end{align}
where $\mu = X1 / 1^\top1 \in\mathbb{R}^{d\times 1}$ is a mean bias of input activation.
Plugging into the loss, we have
\begin{align}
    \mathcal{L} &=
    \| (W-BA)(X-\mu 1^\top) \|^2
    \\
    &=
    \mathrm{tr}[
    (W-BA)\underbrace{(X-\mu 1^\top)(X-\mu 1^\top)^\top}_{C_0\in\mathbb{R}^{d\times d}} (W-BA)^\top ]
    \\
    &=
    \| (W-BA) C_0^\frac{1}{2} \|^2
    .
\end{align}
Hence, the optimal solution is the SVD of weight multiplied with square root of covariance $C_0$ not auto-correlation ($XX^\top$):
\begin{align}
    C_0 &= (C - \mu \mu^\top)l.
\end{align}

\section{Activation-Aware Joint QKV Compression}
\label{sec:qkv}

Consider minimizing QKV activation:
\begin{align}
    \mathcal{L} &=
    \Bigg\| 
    \underbrace{
    \begin{bmatrix}
        W_\mathrm{q} \\
        W_\mathrm{k} \\
        W_\mathrm{v}
    \end{bmatrix}
    }_{W\in\mathbb{R}^{3d'\times d}}
    X
    -
    \underbrace{
    \begin{bmatrix}
        B_\mathrm{q} \\
        B_\mathrm{k} \\
        B_\mathrm{v}
    \end{bmatrix}
    }_{B\in\mathbb{R}^{3d'\times r}}
    A X
    \Bigg\|^2
    .
\end{align}
In this case, the optimal solution is an SVD of $WC^\frac{1}{2}$.

Note that this is different from QKV individual optimization:
\begin{align}
    \mathcal{L}'
    &=
    \sum_{i\in[\mathrm{q}, \mathrm{k}, \mathrm{v}]}
    \| W_i X - B_i A_i X \|^2
    \\
    &=
    \Bigg\| 
    \underbrace{
    \begin{bmatrix}
        W_\mathrm{q} \\
        W_\mathrm{k} \\
        W_\mathrm{v}
    \end{bmatrix}
    }_{A\in\mathbb{R}^{3d'\times d}}
    X
    -
    \underbrace{
    \begin{bmatrix}
        B_\mathrm{q} & O & O \\
        O & B_\mathrm{k} & O \\
        O & O & B_\mathrm{v}
    \end{bmatrix}
    }_{B\in\mathbb{R}^{3d'\times 3r}}
    \underbrace{
    \begin{bmatrix}
        A_\mathrm{q} \\
        A_\mathrm{k} \\
        A_\mathrm{v}
    \end{bmatrix}
    }_{W\in\mathbb{R}^{3r\times d}}
    X
    \Bigg\|^2
    .    
\end{align}
The solution is 3 SVDs: $W_iC^\frac{1}{2}$.

\begin{figure}
    \centering
    \includegraphics[width=0.8\linewidth]{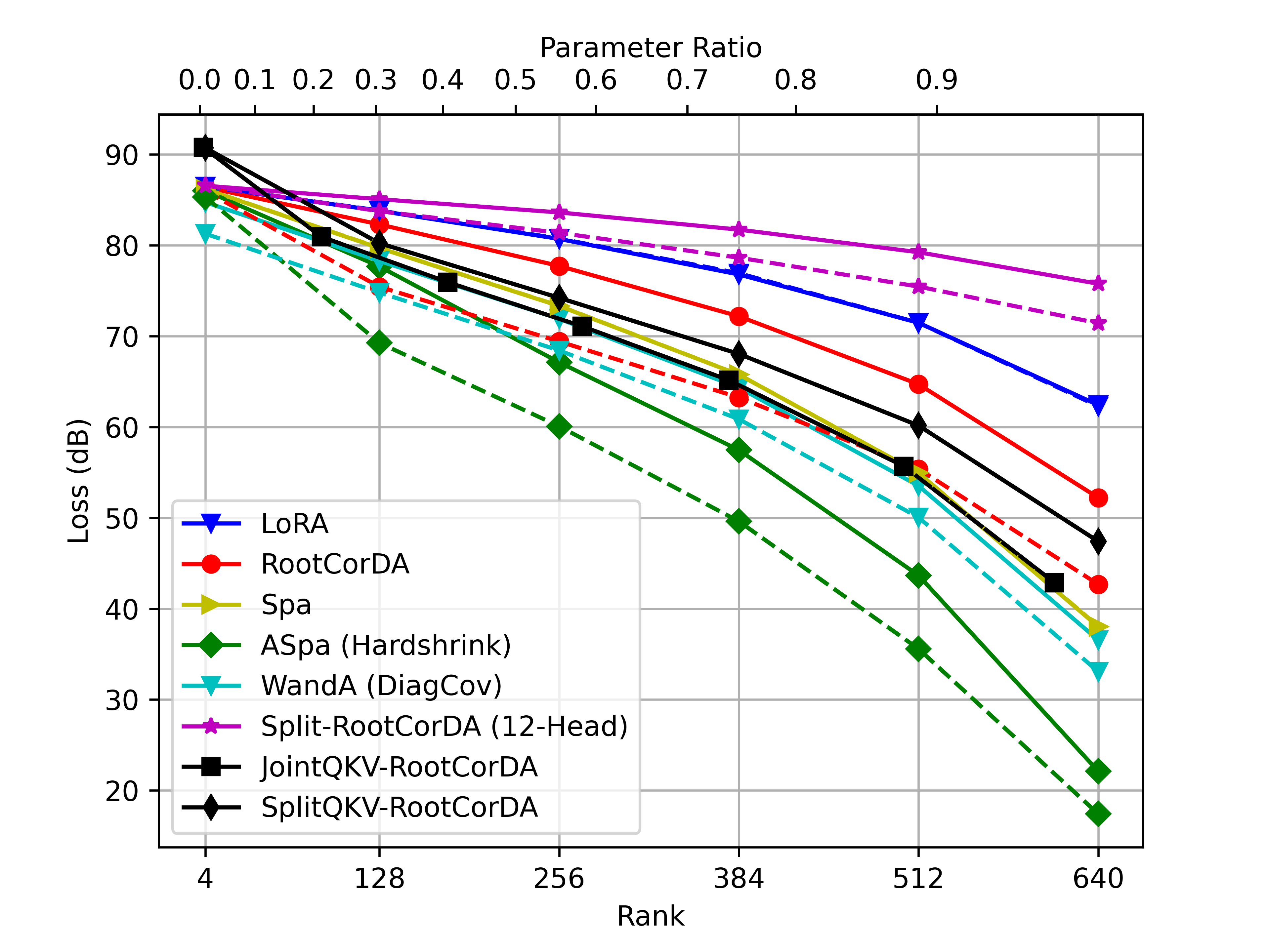}
    \caption{Joint-QKV vs split-QKV approximation.}
    \label{fig:jointQKV}
\end{figure}

The key difference: (i) the shared vs. non-shared compression matrix $A$; (ii) block-diagonal vs. dense decompression matrix $B$.
The number of parameters will be $r(3d'+d)$ from $3r(d'+d)$, allowing 50\% more rank for joint QKV when $d'=d$.
When we use LU factorizatin, the number of parameters will be $r(3d'+d-r)$ from $3r(d'+d-r)$.
We show the benefit of joint-QKV activation-aware distillation in Fig.~\ref{fig:jointQKV}.

Nevertheless, the relative magnitudes over Q/K/V are not well-treated for joint case, and it could be worse in the global performance in the end.

\section{Split-Head Compression}
\label{sec:split}

\begin{figure}
    \centering
    \includegraphics[width=0.8\linewidth]{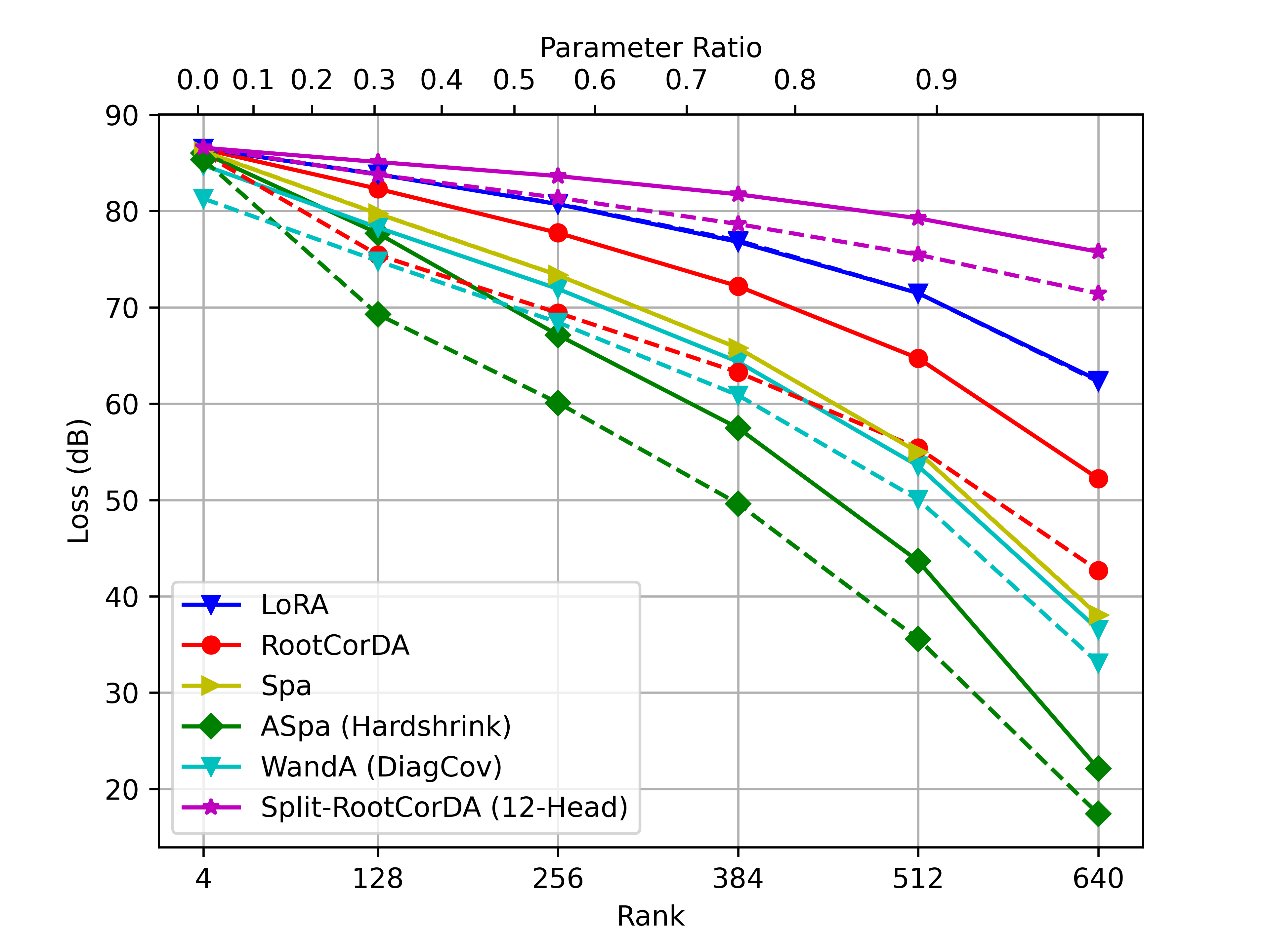}
    \caption{Split-head activation-aware approximation had terrible performance.}
    \label{fig:split}
\end{figure}

Typically, the weight matrix is split into multiple heads, what happens if we use split-head activation loss?
Consider
\begin{align}
    \mathcal{L} &=
    \sum_{i=1}^h \| W_i X - B_i A_i X \|^2
    \\
    &=
    \Bigg\| 
    \underbrace{
    \begin{bmatrix}
        W_1 \\
        \ddots \\
        W_h
    \end{bmatrix}
    }_{W\in\mathbb{R}^{d'\times d}}
    X
    -
    \underbrace{
    \mathrm{diag}[
    B_1, \ldots, B_h
    ]
    }_{B\in\mathbb{R}^{d'\times r}}
    \underbrace{
    \begin{bmatrix}
        A_1 \\
        \ddots \\
        A_h
    \end{bmatrix}
    }_{A\in\mathbb{R}^{r\times d}}
    X
    \Bigg\|^2
    .
\end{align}
where $W_i\in\mathbb{R}^{d'/h\times d}$, $B_i\in\mathbb{R}^{d'/h\times r/h}$, $A_i\in\mathbb{R}^{r/h\times d}$ are the $i$th head approximation with $h$ being the number of heads.
The solution is individual SVD of $W_i C^\frac{1}{2}$.
However, as decompression matrix $B$ is sparse block diagonal, it is not efficient than joint head approximation.
It is shown in Fig.~\ref{fig:split}.

\section{Multi-Head Attention (MHA)}
\label{sec:att}

Typically, the attention projection uses a square weight $d'=d$, but it is divided into multiple heads such that:
\begin{align}
    W_\mathrm{q} &= 
    \begin{bmatrix}
    W_{\mathrm{q},1} \\
    W_{\mathrm{q},2} \\
    \vdots \\
    W_{\mathrm{q},h}
    \end{bmatrix}
    \in\mathbb{R}^{d\times d},
    \qquad
    W_\mathrm{k} =
    \begin{bmatrix}
    W_{\mathrm{k},1} \\
    W_{\mathrm{k},2} \\
    \vdots \\
    W_{\mathrm{k},h}
    \end{bmatrix}
    \in\mathbb{R}^{d\times d},    
\end{align}
where $W_{\mathrm{q},i}\in\mathbb{R}^{d/h\times d}$ and $W_{\mathrm{k},i}\in\mathbb{R}^{d/h\times d}$ are the $i$th head projection weights, and $h$ is number of heads. 
The analysis so far is still valid for per-head low-rank approximation to regard $d'=d/h$.

However, joint-head low-rank approximation may have a benefit over head-wise low-rank approximation.
The $i$th head attention map is given as:
\begin{align}
    M_i &=
    X^\top W_{\mathrm{q},i}^\top W_{\mathrm{k}, i} X.
\end{align}
When we jointly decompose the low-rank with independent rank $r_\mathrm{q}$ and $r_\mathrm{k}$: $\hat{W}_\mathrm{q}=B_\mathrm{q} A_\mathrm{q}$ and $\hat{W}_\mathrm{k}=B_\mathrm{k} A_\mathrm{k}$ 
for $B_\mathrm{q}, A_\mathrm{q}^\top \in\mathbb{R}^{d\times r_\mathrm{q}}$, 
$B_\mathrm{k}, A_\mathrm{k}^\top \in\mathbb{R}^{d\times r_\mathrm{k}}$, we can write
$\hat{W}_{\mathrm{q},i} = B_{\mathrm{q},i} A_\mathrm{q}$ and $\hat{W}_{\mathrm{k},i}=B_{\mathrm{k},i} A_\mathrm{k}$ for $B_{\mathrm{q},i}\in\mathbb{R}^{d/h\times r_\mathrm{q}}$ and $B_{\mathrm{k},i}\in\mathbb{R}^{d/h\times r_\mathrm{k}}$, i.e., the compression matrix $A$ is shared, and decompression matrix $B$ is individual over multiple heads.
It suggests that the rank $r_\mathrm{q}$ and $r_\mathrm{k}$ should not be lower than $d/h$, otherwise some heads can be redundant.

For arbitrary $X$ (worst-case), we may consider minimizing
\begin{align}
\mathcal{L} &=
    \sum_{i=1}^{h} \|
    \underbrace{W_{\mathrm{q},i}^\top W_{\mathrm{k}, i}}_{G_i\in\mathbb{R}^{d\times d}} - A_\mathrm{q}^\top \underbrace{B_{\mathrm{q},i}^\top B_{\mathrm{k},i}}_{H_i\in\mathbb{R}^{r_\mathrm{q}\times r_\mathrm{k}}} A_\mathrm{k} \|^2 
    \\
    &=
    \sum_{i=1}^h
    \| 
    G_i - A_\mathrm{q}^\top H_i A_\mathrm{k}
    \|^2.
    \label{eq:head_loss}
\end{align}
Note that $G_i=W_{\mathrm{q},i}^\top W_{\mathrm{k},i}\in\mathbb{R}^{d\times d}$ is at most of rank $d/h$.
And the rank of $H_i=B_{\mathrm{q},i}^\top B_{\mathrm{k},i}\in\mathbb{R}^{r_\mathrm{q}\times r_\mathrm{k}}$ is not greater than $r=\min(r_\mathrm{q}, r_\mathrm{k}, d/h)$.

Note that there is no loss in generality to restrict that $A_\mathrm{q}$ and $A_\mathrm{k}$ are ortho-normal, i.e., $A_\mathrm{q} A_\mathrm{q}^\top=I_{r_\mathrm{q}}$ and $ A_\mathrm{k}A_\mathrm{k}^\top=I_{r_\mathrm{k}}$, as a full matrix $H_i$ can absorb any non-ortho-normal impact.
Then, we have
\begin{align}
    \mathcal{L} 
    &=
    \sum_{i=1}^h
    \mathrm{tr}[
    G_i G_i^\top
    +
    A_\mathrm{q}^\top H_i \underbrace{A_\mathrm{k} A_\mathrm{k}^\top}_{I_{r_\mathrm{k}}} H_i^\top A_\mathrm{q}
    -
    A_\mathrm{q}^\top H_i A_\mathrm{k} G_i^\top 
    -
    G_i A_\mathrm{k}^\top H_i^\top A_\mathrm{q} 
    ]
    \\
    &=
    \sum_{i=1}^h
    \|
    G_i \|^2
    +
    \mathrm{tr}[
    H_i
    H_i^\top 
    \underbrace{
    A_\mathrm{q} A_\mathrm{q}^\top}_{I_{r_\mathrm{q}}} 
    ]
    -
    \mathrm{tr}[
    H_i A_\mathrm{k} G_i^\top A_\mathrm{q}^\top 
    +
    A_\mathrm{q} G_i A_\mathrm{k}^\top H_i^\top  
    ]
    \\
    &=
    \sum_{i=1}^h \| G_i \|^2 + \| H_i\|^2
    -2
    \mathrm{tr}[    H_i A_\mathrm{k} G_i^\top A_\mathrm{q}^\top
    ].
\end{align}
The gradients:
\begin{align}
    \nabla_{H_i} \mathcal{L}
    &=
    2H_i - 2A_\mathrm{q} G_i A_\mathrm{k}^\top.
\end{align}
The KKT condition for $H_i$ given $A_\mathrm{q}$ and $A_\mathrm{k}$ is thus:
\begin{align}
    H_i &= A_\mathrm{q} G_i A_\mathrm{k}^\top\\
    &=
    A_\mathrm{q} W_{\mathrm{q},i}^\top W_{\mathrm{k},i} A_\mathrm{k}^\top.
    \label{eq:head_hi}
\end{align}
Putting it back to the loss, we obtain:
\begin{align}
    \mathcal{L} 
    &=
    \sum_{i=1}^h \| G_i \|^2 + \| H_i\|^2
    -2
    \mathrm{tr}[    H_i A_\mathrm{k} G_i^\top A_\mathrm{q}^\top
    ]
    \\
    &=
    \sum_i \| G_i \|^2 + \| A_\mathrm{q} G_i A_\mathrm{k}^\top\|^2
    -2
    \mathrm{tr}[    A_\mathrm{q} G_i A_\mathrm{k}^\top A_\mathrm{k} G_i^\top A_\mathrm{q}^\top
    ]
    \\
    &=
    \sum_{i=1}^h \| G_i \|^2 
    + 
    \| A_\mathrm{q} G_i A_\mathrm{k}^\top\|^2
    -2
    \| A_\mathrm{q} G_i A_\mathrm{k}^\top\|^2
    \\
    &=
    \sum_{i=1}^h \| G_i \|^2 
    -
    \|
    A_\mathrm{q} G_i A_\mathrm{k}^\top 
    \|^2.
\end{align}

Let's rewrite as
\begin{align}
    \mathcal{L} &=
    \sum_i \|G_i\|^2
    - \| (A_\mathrm{k} \otimes A_\mathrm{q}) \mathrm{vec}[G_i] \|^2
    \\
    &=
    \sum_i \|G_i\|^2
    - 
    \sum_i \| (A_\mathrm{k} \otimes A_\mathrm{q}) \mathrm{vec}[G_i] \|^2
    \\
    &=
    \sum_i \|G_i\|^2
    - 
    \| (A_\mathrm{k} \otimes A_\mathrm{q}) 
    \underbrace{
    [
    \mathrm{vec}[G_1],
    \mathrm{vec}[G_2],
    \ldots,
    \mathrm{vec}[G_h] 
    ]
    }_{G\in\mathbb{R}^{dd\times h}}
    \|^2 
    \\
    &=
    \|G \|^2
    - 
    \| (A_\mathrm{k} \otimes A_\mathrm{q}) 
    G
    \|^2         
        \\
    &=
    \|G \|^2
    - 
    \| (I_h \otimes A_\mathrm{k} \otimes A_\mathrm{q}) 
    \mathrm{vec}[G]
    \|^2.
\end{align}
This is the 3-mode tensor-rank decomposition problem involving the high-order SVD (HOSVD) for folding $G$ into the size of $d\times d\times h$, but with a restriction that the first mode plane is identity (it may suggest that we may be able to improve by relaxing this constraint).

HOSVD has no simple solution, while alternating method works well in practice.
Specifically, each tensor plane is alternatingly obtained by left singular of the unfolded tensor in different axis.
Given $A_\mathrm{k}$, the best $A_\mathrm{q}$ is the $r_\mathrm{q}$ left singular:
\begin{align}
    A_\mathrm{q}^\top &\leftarrow
    \mathrm{LeftSingular}_{r_\mathrm{q}}(
    \underbrace{
    [G_1 A_\mathrm{k}^\top, G_2 A_\mathrm{k}^\top, \ldots,
    G_h A_\mathrm{k}^\top]
    }_{\mathbb{R}^{d\times r_\mathrm{k}h}}
    )
    \\
    &=
    \mathrm{LeftSingular}_{r_\mathrm{q}}(
    \sum_i G_i A_\mathrm{k}^\top A_\mathrm{k} G_i^\top
    ).
\end{align}
The loss will be the residual accumulation of the eigenvalues beyond the rank $r_\mathrm{q}$.
Then given $A_\mathrm{q}$, the best $A_\mathrm{k}$ is the $r_\mathrm{k}$ left singular:
\begin{align}
    A_\mathrm{k}^\top &\leftarrow
    \mathrm{LeftSingular}_{r_\mathrm{k}}(
    \underbrace{
    [G_1^\top A_\mathrm{q}^\top, G_2^\top A_\mathrm{q}^\top, \ldots,
    G_h^\top A_\mathrm{q}^\top]
    }_{\mathbb{R}^{d\times r_\mathrm{q}h}}
    )
    \\
    &= 
    \mathrm{LeftSingular}_{r_\mathrm{k}}(
    \sum_i G_i^\top A_\mathrm{q}^\top A_\mathrm{q} G_i
    ).
\end{align}
The loss will be the residual accumulation of the eigenvalues beyond the rank $r_\mathrm{k}$.
Iterating the above often achieves a good solution.
NOTE: singular vectors of $\sum_i G_i G_i^\top$ and $\sum_i G_i^\top G_i$ can be a good initialization of $A_\mathrm{q}$ and $A_\mathrm{k}$.

NOTE: the non-uniform choice of ranks $r_\mathrm{q}$ and $r_\mathrm{k}$ can be optimized to minimize the loss, rather than using the same rank.
It can be adaptively adjusted from the eigenvalue distributions.

Once we obtained the HOSVD solution for tensor planes $A_\mathrm{q}$ and $A_\mathrm{k}$, the tensor core $H_i\in\mathbb{R}^{r_\mathrm{q}\times r_\mathrm{k}}$ is generated by (\ref{eq:head_hi}) as $H_i=A_\mathrm{q}G_i A_\mathrm{k}^\top = A_\mathrm{q}W_{\mathrm{q},i}^\top W_{\mathrm{k},i} A_\mathrm{k}^\top$.
Given optimized $H_i$, any arbitrary $B_{\mathrm{q},i}$ and $B_{\mathrm{k},i}$ provides the same error as long as it holds:
\begin{align}
    H_i &= B_{\mathrm{q},i}^\top B_{\mathrm{k},i}.
\end{align}
The solution is 
\begin{align}
    B_{\mathrm{q},i} &= 
    J_i^\top
    W_{\mathrm{q},i} A_\mathrm{q}^\top, \\
    B_{\mathrm{k},i} &=
    J_i^+
    W_{\mathrm{k},i} A_\mathrm{k}^\top,
\end{align}
where $J_i\in\mathbb{R}^{d/h\times d/h}$ is any arbitrary full-rank matrix of our choice.
The most natural choice is identity: $J_i=I_{d/h}$.

Nevertheless, another simple solution will be
\begin{align}
    B_{\mathrm{q},i} &= 
    I_{d/h\times r_\mathrm{q}},\\
    B_{\mathrm{k},i} &=
    \begin{bmatrix}
        H_i \\
        O_{(d/h-r_\mathrm{q})\times r_\mathrm{k}}
    \end{bmatrix},
\end{align}
when $r_\mathrm{q} \leq d/h$.
This has a benefit that query decompression does not require any memory and key decompression is a block sparse.

Another solution will be
\begin{align}
    B_{\mathrm{q},i} &= 
    \begin{bmatrix}
        H_i^\top \\
        O_{(d/h-r_\mathrm{k})\times r_\mathrm{q}}
    \end{bmatrix},
    \\
    B_{\mathrm{k},i} &=
    I_{d/h\times r_\mathrm{k}},
\end{align}
when $r_\mathrm{k}\leq d/h$. 
Similar benefit, but probably removing the requirement of query decompression is more beneficial than key decompression in practice.

When $r_\mathrm{k}, r_\mathrm{q} \geq d/h$ (in most case?), we can select $J_i$ such that $B_{\mathrm{q},i}$ or $B_{\mathrm{k},i}$ is block matrix to save $(d/h)^2$ parameters from $(r_\mathrm{q}+r_\mathrm{k})d/h$.
For this case, fine-tuning two decompression $B_{\mathrm{q},i}$ and $B_{\mathrm{k},i}$ rather than a product $H_i$ will be more parameter-efficient.

\subsection{Activation-Aware Multi-Head Latent Attention}
\label{sec:mha_cov}

Consider the loss:
\begin{align}
    \mathcal{L} &= \sum_i \| M_i - \hat{M}_i \|^2
    \\
    &=
    \sum_i
    \| X^\top \underbrace{(G_i - A_\mathrm{q}^\top H_i A_\mathrm{k})}_{\varDelta_i \in\mathbb{R}^{d\times d}} X \|^2
    \\
    &=
    \sum_i
    \mathrm{tr}[X^\top \varDelta_i X X^\top \varDelta_i^\top X] 
    \\
    &=
    \sum_i
    \mathrm{tr}[ \varDelta_i \underbrace{X X^\top}_{C\in\mathbb{R}^{d\times d}} \varDelta_i^\top XX^\top] 
    \\
    &=
    \sum_i
    \mathrm{tr}[ \varDelta_i C \varDelta_i^\top C]
    \\
    &=
    \sum_i
    \mathrm{tr}[ \varDelta_i C^\frac{1}{2}C^\frac{1}{2} \varDelta_i^\top C^\frac{1}{2} C^\frac{1}{2}]
    \\
    &=
    \sum_i
    \| C^\frac{1}{2}\varDelta_i C^\frac{1}{2} \|^2
    \\
    &=
    \sum_i
    \| \underbrace{C^\frac{1}{2}G_i C^\frac{1}{2}}_{G_i'}
    -
    \underbrace{C^\frac{1}{2}A_\mathrm{q}^\top}_{A_\mathrm{q}'^\top} 
    H_i
    \underbrace{
    A_\mathrm{k} C^\frac{1}{2}
    }_{A_\mathrm{k}'}
    \|^2
    ,
    \label{eq:head_loss_cov}
\end{align}
where $C$ is a positive semi-definite of rank no greater than $\min(d, l)$.

In fact, the solution is same as the case without $X$ comparing (\ref{eq:head_loss}) and (\ref{eq:head_loss_cov}), where we can regard as
\begin{align}
    G_i &\rightarrow G_i' = C^\frac{1}{2} G_i C^\frac{1}{2},
    \\
    A_\mathrm{q} &\rightarrow A_\mathrm{q}' = A_\mathrm{q} C^\frac{1}{2},
    \\
    A_\mathrm{k} &\rightarrow A_\mathrm{k}' = A_\mathrm{k} C^\frac{1}{2}.
\end{align}
Here, we can consider $A_\mathrm{q}C^\frac{1}{2}$ and $A_\mathrm{k}C^\frac{1}{2}$ are instead orthogonal, and thus the solution can be given by HOSVD likewise.

\begin{figure}[t]
\centering
\includegraphics[width=0.8\linewidth]{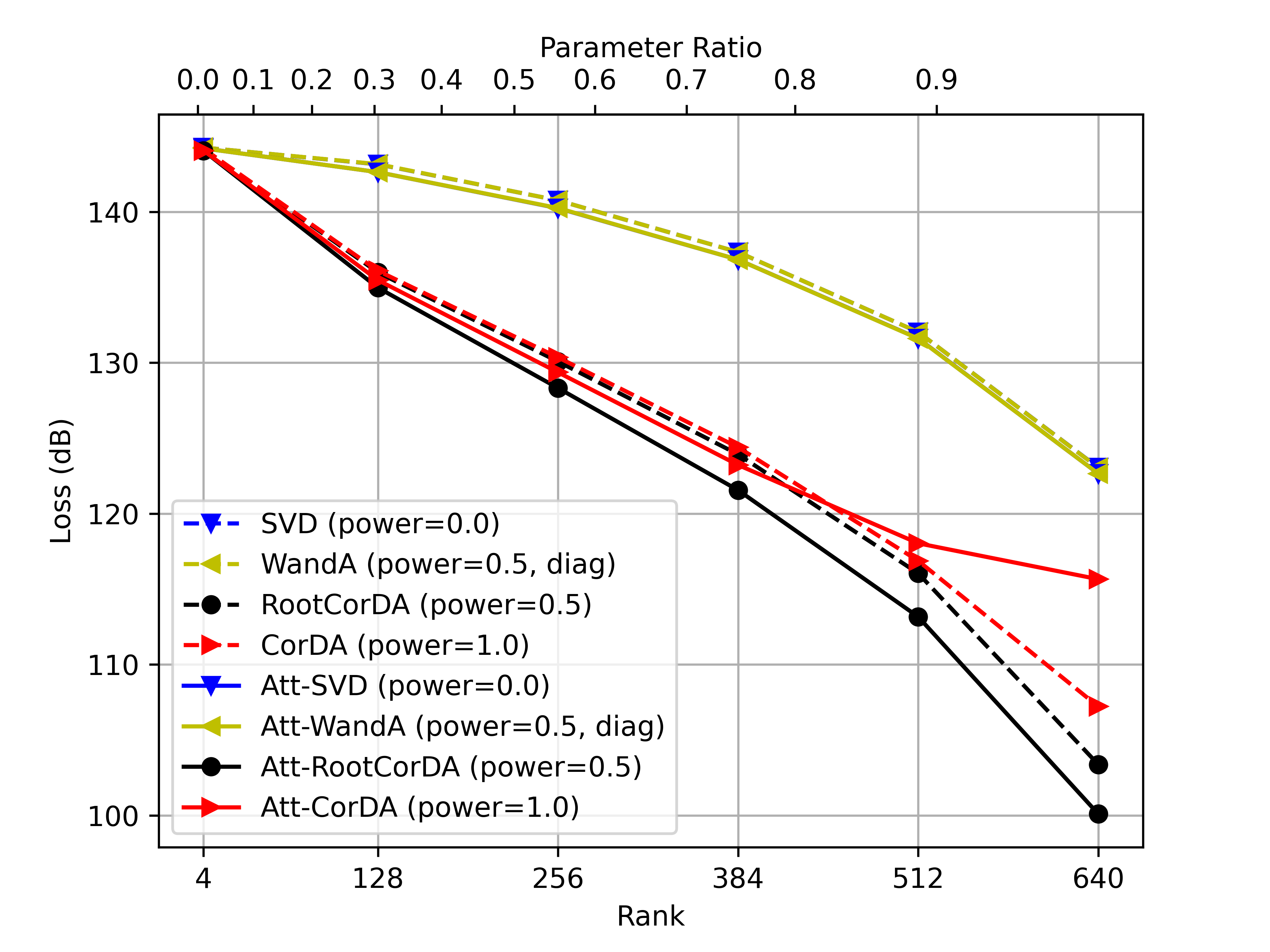}
\caption{Attention-Aware vs. Activation-Aware Approximation. 
Loss is attention map error.
Random query/key projections with Wishart sample correlation ($0.9$ decaying).
WandA uses diagonal correlation.}
\label{fig:att}
\end{figure}

Fig.~\ref{fig:att} shows the comparison between adaptive and non-adaptive distillation with activation/attention-aware methods.

\begin{figure}[t]
\centering
\includegraphics[width=0.8\linewidth]{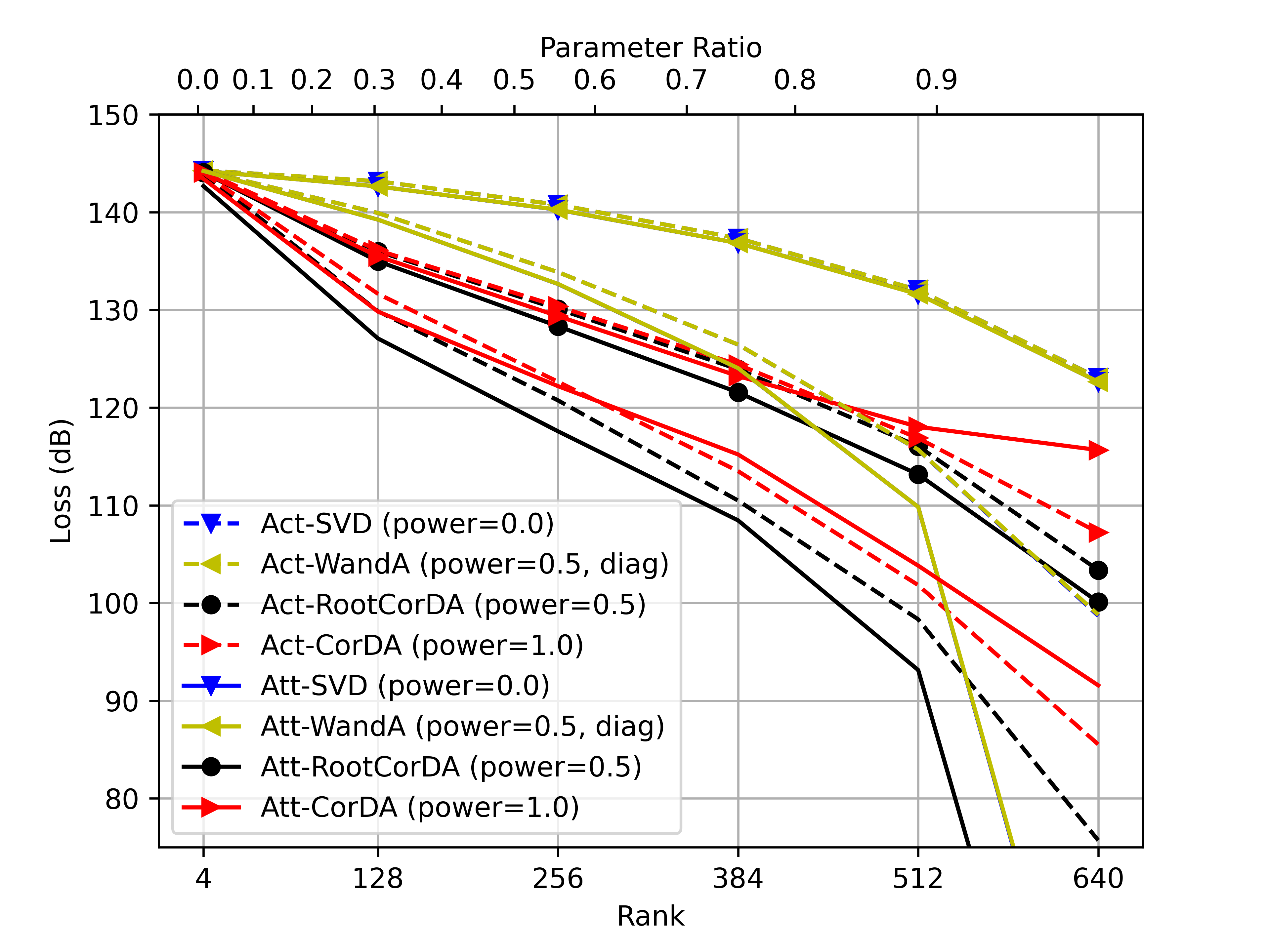}
\caption{Sparse approximation for Attention-Aware vs. Activation-Aware distillation. 
No markers are sparse approximation.
Sparse is better than low-rank.}
\label{fig:att_spa}
\end{figure}

\subsection{Bias Update}
\label{sec:bias2}

Some LLMs use bias for QKV. 
For this case, we need to modify the bias term as well.
We have
\begin{align}
    \mathcal{L} &=
    \sum_i
    \| 
    (W_{\mathrm{q},i} X + b_{\mathrm{q},i} 1^\top)^\top
    (W_{\mathrm{k},i} X + b_{\mathrm{k},i} 1^\top)
    -
    (\hat{W}_{\mathrm{q},i} X + \hat{b}_{\mathrm{q},i} 1^\top)^\top
    (\hat{W}_{\mathrm{k},i} X + \hat{b}_{\mathrm{k},i} 1^\top)
    \|^2
    \\
    &=
    \sum_i
    \Big\| 
    \big(
    \underbrace{
    \begin{bmatrix}
        W_{\mathrm{q},i} &
        b_{\mathrm{q},i}
    \end{bmatrix}
    }_{W_{\mathrm{q},i}'\in\mathbb{R}^{d/h\times (d+1)}}
    \underbrace{
    \begin{bmatrix}
        X \\
        1^\top
    \end{bmatrix}
    }_{X'\in\mathbb{R}^{(d+1)\times l}}
    \big)^\top
    \big(
    \underbrace{
    \begin{bmatrix}
        W_{\mathrm{k},i} &
        b_{\mathrm{k},i}
    \end{bmatrix}
    }_{W'_{\mathrm{k},i}\in\mathbb{R}^{d/h\times (d+1}}
    \begin{bmatrix}
        X \\
        1^\top
    \end{bmatrix}
    \big)
    -        
    \big(
    \underbrace{
    \begin{bmatrix}
        \hat{W}_{\mathrm{q},i} &
        \hat{b}_{\mathrm{q},i}
    \end{bmatrix}
    }_{\hat{W}_{\mathrm{q},i}'\in\mathbb{R}^{d/h\times (d+1)}}
    \begin{bmatrix}
        X \\
        1^\top
    \end{bmatrix}
    \big)^\top
    \big(
    \underbrace{
    \begin{bmatrix}
        \hat{W}_{\mathrm{k},i} &
        \hat{b}_{\mathrm{k},i}
    \end{bmatrix}
    }_{\hat{W}_{\mathrm{k},i}'\in\mathbb{R}^{d/h\times (d+1)}}
    \begin{bmatrix}
        X \\
        1^\top
    \end{bmatrix}
    \big)
    \Big\|^2
    \\
    &=
    \sum_i
    \| X'^\top (W_{\mathrm{q},i}'^\top 
    W_{\mathrm{k},i}'
    -
    \hat{W}_{\mathrm{q},i}'^\top 
    \hat{W}_{\mathrm{k},i}'
    )
    X'
    \|^2
    \\
    &=
    \sum_i 
    \| \tilde{C}^\frac{1}{2} (W_{\mathrm{q},i}'^\top 
    W_{\mathrm{k},i}'
    -
    \hat{W}_{\mathrm{q},i}'^\top 
    \hat{W}_{\mathrm{k},i}'
    )
    \tilde{C}^\frac{1}{2}
    \|^2
\end{align}
where we have a modified covariance:
\begin{align}
    \tilde{C} &=X' X'^\top\in\mathbb{R}^{(d+1)\times (d+1)}
    \\
    &=
    \begin{bmatrix}
        X \\
        1^\top
    \end{bmatrix}
    \begin{bmatrix}
        X^\top &
        1
    \end{bmatrix}
    \\
    &=
    \begin{bmatrix}
        l C & l \mu \\
        l \mu^\top & l
    \end{bmatrix}
    =
    l 
    \begin{bmatrix}
    C & \mu \\
    \mu^\top & 1
    \end{bmatrix}
    \\
    &=
    l 
    \begin{bmatrix}
    C^\frac{1}{2} & O \\
    \mu^\top C^\frac{-1}{2} &
    (1-\mu^\top C^+ \mu)^\frac{1}{2}
    \end{bmatrix}
    \begin{bmatrix}
    C^\frac{1}{2} & C^\frac{-1}{2}\mu \\
    O &
    (1-\mu^\top C^+ \mu)^\frac{1}{2}
    \end{bmatrix}
\end{align}
where we assume $C$ is normalized as $C=XX^\top / l$, and $\mu\in\mathbb{R}^{d\times 1}$ is a mean of input tokens: $\mu=X 1/l$.
Then, we can omit $l$.
Similar format but it cannot be solved by the same way as we have a structured low-rank expression:
\begin{align}
    \hat{W}_{\mathrm{q},i}'^\top
    \hat{W}_{\mathrm{k},i}'
    &=
    \begin{bmatrix}
        A_\mathrm{q}^\top B_{\mathrm{q},i}^\top
        \\
        \hat{b}_{\mathrm{q},i}^\top
    \end{bmatrix}
    \begin{bmatrix}
         B_{\mathrm{k},i}
         A_\mathrm{k}
        &
        \hat{b}_{\mathrm{k},i}
    \end{bmatrix}
    \\
    &=
    \underbrace{
    \begin{bmatrix}
        A_\mathrm{q}^\top & O_{d\times 1} \\
        O_{1\times r_\mathrm{q}} & 1
    \end{bmatrix}
    }_{A_\mathrm{q}'^\top\in\mathbb{R}^{(d+1)\times (r_\mathrm{q}+1)}}
    \overbrace{
    \underbrace{
    \begin{bmatrix}          B_{\mathrm{q},i}^\top
        \\
        \hat{b}_{\mathrm{q},i}^\top
    \end{bmatrix}
    }_{B_{\mathrm{q,i}}'^\top\in\mathbb{R}^{(r_\mathrm{q}+1)\times d}}
    \underbrace{
    \begin{bmatrix}
         B_{\mathrm{k},i}
        &
        \hat{b}_{\mathrm{k},i}
    \end{bmatrix}    
    }_{B_{\mathrm{k},i}'\in\mathbb{R}^{d\times (r_\mathrm{k}+1)}}
    }^{H_i'\in\mathbb{R}^{(r_\mathrm{q}+1)\times(r_\mathrm{k}+1)}}
    \underbrace{
    \begin{bmatrix}
         A_\mathrm{k} & O_{r_\mathrm{k}\times 1} \\
        O_{1\times d} & 1
    \end{bmatrix}
    }_{A_\mathrm{k}'\in\mathbb{R}^{(r_\mathrm{k}+1)\times (d+1)}}.
\end{align}
We may use the HOSVD to decompose with one more rank for bias, while the compression matrix $A'_\mathrm{q}$ and $A'_\mathrm{k}$ needs to be a particular format.
Nonetheless, we can modify the bias by the KKT condition:
\begin{align}
    A_\mathrm{q}' 
    \tilde{C}
    G_i 
    \tilde{C}
    A_\mathrm{k}'^\top
    &=
    A_\mathrm{q}' 
    \tilde{C}
    A_\mathrm{q}'^\top 
    H_i' 
    A_\mathrm{k}' 
    \tilde{C}
    A_\mathrm{k}'^\top.
\end{align}
Hence we have
\begin{align}
    H_i' &=
    (A_\mathrm{q}' 
    \tilde{C}
    A_\mathrm{q}'^\top)^+
    A_\mathrm{q}'
    \tilde{C}
    G_i 
    \tilde{C}
    A_\mathrm{k}'^\top
    (A_\mathrm{k}' 
    \tilde{C}
    A_\mathrm{k}'^\top)^+
    \\
    &=
    \underbrace{
    (A_\mathrm{q}' 
    \tilde{C}
    A_\mathrm{q}'^\top)^+
    A_\mathrm{q}'
    \tilde{C}
    W_{\mathrm{q},i}'^\top
    }_{B_{\mathrm{q},i}^\top}
    \underbrace{
    W_{\mathrm{k},i}' 
    \tilde{C}
    A_\mathrm{k}'^\top
    (A_\mathrm{k}' 
    \tilde{C}
    A_\mathrm{k}'^\top)^+
    }_{B_{\mathrm{k},i}}.
\end{align}
Thus, given optimized $A_{\mathrm{q}}$ and $A_\mathrm{k}$, we have optimized decompression matrix with updated bias:
\begin{align}
    B_{\mathrm{q},i}' &=    
    \begin{bmatrix}
        B_{\mathrm{q},i} & \hat{b}_{\mathrm{q},i}
    \end{bmatrix}
    \\
    &=
    J_i^\top W_{\mathrm{q},i}'
    \tilde{C} A_\mathrm{q}'^\top
    (A_\mathrm{q}' \tilde{C} A_\mathrm{q}'^\top)^+
    \\
    &=
    J_i^\top
    \begin{bmatrix}
        W_{\mathrm{q},i} &
        b_{\mathrm{q},i}
    \end{bmatrix}
    \tilde{C} A_\mathrm{q}'^\top
    (A_\mathrm{q}' \tilde{C} A_\mathrm{q}'^\top)^+   
    \\
    &=
    J_i^\top
    \begin{bmatrix}
        W_{\mathrm{q},i} &
        b_{\mathrm{q},i}
    \end{bmatrix}
    \begin{bmatrix}
        C A_\mathrm{q}^\top & \mu \\
        \mu^\top A_\mathrm{q}^\top & 1
    \end{bmatrix}
    \begin{bmatrix}
        A_\mathrm{q} C A_\mathrm{q}^\top & A_\mathrm{q} \mu \\
        \mu^\top A_\mathrm{q}^\top &
        1
    \end{bmatrix}^{+}    
    \\
    &=
    J_i^\top
    \begin{bmatrix}
        W_{\mathrm{q},i} &
        b_{\mathrm{q},i}
    \end{bmatrix}
    \begin{bmatrix}
        C A_\mathrm{q}^\top & \mu \\
        \mu^\top A_\mathrm{q}^\top & 1
    \end{bmatrix}
    \begin{bmatrix}
        I & O \\
        - \mu^\top A_\mathrm{q}^\top &
        1
    \end{bmatrix}        
    \begin{bmatrix}
        ( A_\mathrm{q} C A_\mathrm{q}^\top -
        A_\mathrm{q} \mu \mu^\top A_\mathrm{q}^\top)^+
         & O \\
        O &
        1
    \end{bmatrix}        
    \begin{bmatrix}
        I & 
        -  A_\mathrm{q} \mu \\
        O & 1
    \end{bmatrix}    
    \\
    &=
    J_i^\top
    \begin{bmatrix}
        W_{\mathrm{q},i} &
        b_{\mathrm{q},i}
    \end{bmatrix}
    \begin{bmatrix}
        (C-\mu\mu^\top)A_\mathrm{q}^\top ( A_\mathrm{q} C A_\mathrm{q}^\top -
        A_\mathrm{q} \mu \mu^\top A_\mathrm{q}^\top)^+ & 
        -(C-\mu\mu^\top) A_\mathrm{q}^\top ( A_\mathrm{q} C A_\mathrm{q}^\top -
        A_\mathrm{q} \mu \mu^\top A_\mathrm{q}^\top)^+ A_\mathrm{q} \mu  + \mu\\
        O & 1
    \end{bmatrix}
    \\
    &=
    J_i^\top
    \begin{bmatrix}
        W_{\mathrm{q},i}
        (C-\mu\mu^\top)A_\mathrm{q}^\top ( A_\mathrm{q} C A_\mathrm{q}^\top -
        A_\mathrm{q} \mu \mu^\top A_\mathrm{q}^\top)^+ & 
        -W_{\mathrm{q},i}
        (C-\mu\mu^\top) A_\mathrm{q}^\top ( A_\mathrm{q} C A_\mathrm{q}^\top -
        A_\mathrm{q} \mu \mu^\top A_\mathrm{q}^\top)^+ A_\mathrm{q} \mu  + W_{\mathrm{q},i}\mu
        +b_{\mathrm{q},i}
    \end{bmatrix}
    .
\end{align}
It gives the bias modifications:
\begin{align}
    \hat{b}_\mathrm{q}
    &=
    \mathrm{diag}[J_i]
    \big(
    b_\mathrm{q} + W_\mathrm{q}\mu 
    - W_\mathrm{q} (C-\mu\mu^\top) A_\mathrm{q}^\top ( A_\mathrm{q} C A_\mathrm{q}^\top
    -
    A_\mathrm{q} \mu \mu^\top A_\mathrm{q}^\top)^+ A_\mathrm{q} \mu
    \big),
    \\
    \hat{b}_\mathrm{k}
    &=
    \mathrm{diag}[J_i^+]
    \big(
    b_\mathrm{k} 
    + W_\mathrm{k}\mu
    - W_\mathrm{k} (C-\mu\mu^\top) A_\mathrm{k}^\top ( A_\mathrm{k} C A_\mathrm{k}^\top
    -
    A_\mathrm{k} \mu \mu^\top A_\mathrm{k}^\top)^+ A_\mathrm{k} \mu
    \big).
\end{align}
We define the centered auto-correlation:
\begin{align}
    C_0 &= C - \mu \mu^\top.
\end{align}
Then, we assume that the optimal compression matrices $A_\mathrm{q}$ and $A_\mathrm{k}$ are orthogonal on $C_0^\frac{1}{2}$:
\begin{align}
    A_\mathrm{q} C_0 A_\mathrm{q}^\top = I_{r_\mathrm{q}},\\
    A_\mathrm{k} C_0 A_\mathrm{k}^\top = I_{r_\mathrm{k}}.
\end{align}
In this case, the bias modification can reduce to
\begin{align}
        \hat{b}_\mathrm{q}
    &=
    \mathrm{diag}[J_i]
    \big(
    b_\mathrm{q} + W_\mathrm{q}\mu 
    - W_\mathrm{q} C_0 A_\mathrm{q}^\top A_\mathrm{q} \mu
    \big),
    \\
    \hat{b}_\mathrm{k}
    &=
    \mathrm{diag}[J_i^+]
    \big(
    b_\mathrm{k} 
    + W_\mathrm{k}\mu
    - W_\mathrm{k} C_0 A_\mathrm{k}^\top A_\mathrm{k} \mu
    \big).
\end{align}

For this case, we have
\begin{align}
    A_\mathrm{q}'\tilde{C}A_\mathrm{q}^\top
    &=
    \begin{bmatrix}
        A_\mathrm{q} C A_\mathrm{q}^\top & A_\mathrm{q} \mu \\
        \mu^\top A_\mathrm{q}^\top &
        1
    \end{bmatrix}
    ,
    \\
    (A_\mathrm{q}'\tilde{C}A_\mathrm{q}^\top)^+
    &=
    \begin{bmatrix}
        I & -A_\mathrm{q}\mu \\
        -\mu^\top A_\mathrm{q}^\top &
        1+\mu^\top A_\mathrm{q}^\top A_\mathrm{q} \mu
    \end{bmatrix}
    ,
    \\
    A_\mathrm{q}' (A_\mathrm{q}'\tilde{C}A_\mathrm{q}'^\top)^+
    A_\mathrm{q}'^\top
    &=
    \begin{bmatrix}
        A_\mathrm{q}^\top A_\mathrm{q} & -A_\mathrm{q}^\top A_\mathrm{q} \mu \\
        -\mu^\top A_\mathrm{q}^\top A_\mathrm{q} &
        1 + \mu^\top A_\mathrm{q}^\top A_\mathrm{q} \mu
    \end{bmatrix}
    ,
    \\
    A_\mathrm{q}' (A_\mathrm{q}'\tilde{C}A_\mathrm{q}'^\top)^+
    A_\mathrm{q}'^\top
    \tilde{C}
    &=
    \begin{bmatrix}
        A_\mathrm{q}^\top A_\mathrm{q} C_0  & O \\
        \mu^\top -\mu^\top A_\mathrm{q}^\top A_\mathrm{q} C_0 & 1
    \end{bmatrix}
    ,
    \\
    \tilde{C}
    A_\mathrm{q}' (A_\mathrm{q}'\tilde{C}A_\mathrm{q}'^\top)^+
    A_\mathrm{q}'^\top
    &=
    \begin{bmatrix}
        C_0 A_\mathrm{q}^\top A_\mathrm{q}   & \mu - C_0 A_\mathrm{q}^\top A_\mathrm{q} \mu  \\
        O & 1
    \end{bmatrix}
    ,
    \\
    \tilde{C}
    A_\mathrm{q}' (A_\mathrm{q}'\tilde{C}A_\mathrm{q}'^\top)^+
    A_\mathrm{q}'^\top
    \tilde{C}
    &=
    \begin{bmatrix}
        C_0 A_\mathrm{q}^\top A_\mathrm{q} C_0 + \mu \mu^\top & \mu \\
        \mu^\top & 1
    \end{bmatrix}
    .
\end{align}

Plugging the optimized $H_i$, the loss is expressed as
\begin{align}
    \mathcal{L}
    &=
    \sum_i \big\|
    \tilde{C}^\frac{1}{2}
    W_{\mathrm{q},i}'^\top W_{\mathrm{k},i}' 
    \tilde{C}^\frac{1}{2}
    -
    \tilde{C}^\frac{1}{2}
    A_\mathrm{q}'^\top
    (A_\mathrm{q}' \tilde{C} A_\mathrm{q}'^\top)^+
    A_\mathrm{q}'
    \tilde{C}
    W_{\mathrm{q},i}'^\top
    W_{\mathrm{k},i}' 
    \tilde{C}
    A_\mathrm{k}'^\top
    (A_\mathrm{k}' \tilde{C} A_\mathrm{k}'^\top)^+
    A_\mathrm{k}'
    \tilde{C}^\frac{1}{2}
    \big\|^2
    \\
    &=
    \sum_i \|
    \tilde{C}^\frac{1}{2} 
    W_{\mathrm{q},i}'^\top
    W_{\mathrm{k},i}'
    \tilde{C}^\frac{1}{2} 
    |^2
    -
    \|\tilde{C}^\frac{1}{2} A_\mathrm{q}^\top H_i
    A_\mathrm{k} \tilde{C}^\frac{1}{2}
    \|^2
    \\
    &=
    \sum_i \|
    \tilde{C}^\frac{1}{2} 
    W_{\mathrm{q},i}'^\top
    W_{\mathrm{k},i}'
    \tilde{C}^\frac{1}{2} 
    |^2
    -
    \|
    \tilde{C}^\frac{1}{2}
    A_\mathrm{q}'^\top
    (A_\mathrm{q}' \tilde{C} A_\mathrm{q}'^\top)^+
    A_\mathrm{q}'
    \tilde{C}
    W_{\mathrm{q},i}'^\top
    W_{\mathrm{k},i}' 
    \tilde{C}
    A_\mathrm{k}'^\top
    (A_\mathrm{k}' \tilde{C} A_\mathrm{k}'^\top)^+
    A_\mathrm{k}'
    \tilde{C}^\frac{1}{2}
    \|^2
    \\    
    \\
    &=
    \sum_i \|
    \tilde{C}^\frac{1}{2} 
    W_{\mathrm{q},i}'^\top
    W_{\mathrm{k},i}'
    \tilde{C}^\frac{1}{2} 
    |^2
    -
    \mathrm{tr}
    [
    \tilde{C}
    A_\mathrm{q}'^\top
    (A_\mathrm{q}' \tilde{C} A_\mathrm{q}'^\top)^+
    A_\mathrm{q}'
    \tilde{C}
    \underbrace{    
    W_{\mathrm{q},i}'^\top
    W_{\mathrm{k},i}' 
    \tilde{C}
    A_\mathrm{k}'^\top
    (A_\mathrm{k}' \tilde{C} A_\mathrm{k}'^\top)^+
    A_\mathrm{k}'
    \tilde{C} W_{\mathrm{k},i}'^\top
    W_{\mathrm{q},i}'
    }_{G_{\mathrm{q},i}\in\mathbb{R}^{(d+1)\times (d+1)}}
    ]
    .
\end{align}
Focusing on optimizing $A_\mathrm{q}$, the second term will be
\begin{align}
    \sum_i
    \mathrm{tr}
    \Big[
    \Big(
    \begin{bmatrix}
        C_0 A_\mathrm{q}^\top A_\mathrm{q} C_0 & O \\
        O & 0 \\
    \end{bmatrix}
    +
    \begin{bmatrix}
        \mu \\
        1
    \end{bmatrix}
    \begin{bmatrix}
        \mu \\
        1
    \end{bmatrix}^\top
    \Big)
    G_{\mathrm{q},i}
    \Big]
    &=
    \sum_i 
    \mathrm{tr}\Big[
    \begin{bmatrix}
        C_0 A_\mathrm{q}^\top A_\mathrm{q} C_0 & O \\
        O & 0 \\
    \end{bmatrix}
    G_{\mathrm{q},i}    
    \Big]
    + \mathrm{c.c.}
    \\
    &=
    \mathrm{tr}
    \Big[
    C_0 A_\mathrm{q}^\top A_\mathrm{q} C_0
    I_{d\times (d+1)}
    \sum_i
    G_{\mathrm{q},i}
    I_{(d+1)\times d}
    \Big]
    +
    \mathrm{c.c.}
    \\
    &=
    \| 
    A_\mathrm{q} C_0
    (
    I_{d\times (d+1)}
    \sum 
    G_{\mathrm{q},i}
    I_{(d+1)\times d}
    )^\frac{1}{2}
    \|^2
    +\mathrm{c.c.}
    .    
\end{align}
Hence the optimal $A_\mathrm{q}$ is the right-singular vectors:
\begin{align}
    A_\mathrm{q} C_0^\frac{1}{2} &=
    \mathsf{RightSingular}_{r_\mathrm{q}}\big[
    C_0^\frac{1}{2}
    I_{d\times (d+1)}
    (
    \sum_i 
    G_{\mathrm{q},i}
    )
    I_{(d+1)\times d}
    C_0^\frac{1}{2}
    \big]
    .
\end{align}
In fact, we can re-write $G_{\mathrm{q},i}$ as
\begin{align}
    G_{\mathrm{q},i} 
    &=
    \begin{bmatrix}
        W_{\mathrm{q},i}^\top \\
        b_{\mathrm{q},i}^\top
    \end{bmatrix}
    \begin{bmatrix}
        W_{\mathrm{k},i} & 
        b_{\mathrm{k},i}
    \end{bmatrix}
    \Big(
    \begin{bmatrix}
        C_0 A_\mathrm{k}^\top A_\mathrm{k} C_0 & O \\
        O & 0
    \end{bmatrix}
    +
    \begin{bmatrix}
        \mu \\
        1
    \end{bmatrix}
    \begin{bmatrix}
        \mu \\
        1
    \end{bmatrix}^\top
    \Big)
    \begin{bmatrix}
        W_{\mathrm{k},i}^\top \\
        b_{\mathrm{k},i}^\top
    \end{bmatrix}
    \begin{bmatrix}
        W_{\mathrm{q},i} &
        b_{\mathrm{q},i}
    \end{bmatrix}
    \\
    &=
    \begin{bmatrix}
        W_{\mathrm{q},i}^\top \\
        b_{\mathrm{q},i}^\top
    \end{bmatrix}
    \Big(
    W_{\mathrm{k},i}C_0
    A_\mathrm{k}^\top
    A_\mathrm{k}
    C_0
    W_{\mathrm{k},i}^\top 
    +
    (W_{\mathrm{k},i} \mu + b_{\mathrm{k},i})
    (W_{\mathrm{k},i} \mu + b_{\mathrm{k},i})^\top
    \Big)
    \begin{bmatrix}
        W_{\mathrm{q},i} &
        b_{\mathrm{q},i}
    \end{bmatrix}
    .
\end{align}
Hence, we have
\begin{align}
    A_\mathrm{q} C_0^\frac{1}{2} &=
    \mathsf{RightSingular}_{r_\mathrm{k}}\big[
    \sum_i
    C_0^\frac{1}{2}
    W_{\mathrm{q},i}^\top
    W_{\mathrm{k},i}C_0
    A_\mathrm{k}^\top
    A_\mathrm{k}
    C_0
    W_{\mathrm{k},i}^\top 
    W_{\mathrm{q},i}
    C_0^\frac{1}{2}
    \notag\\
    &\qquad
    +
    \sum_i
    C_0^\frac{1}{2}
    W_{\mathrm{q},i}^\top
    (W_{\mathrm{k},i} \mu + b_{\mathrm{k},i})
    (W_{\mathrm{k},i} \mu + b_{\mathrm{k},i})^\top
    W_{\mathrm{q},i}
    C_0^\frac{1}{2}
    \big].    
\end{align}
The first term is the solution if no bias and mean are present.

Similarly the solution for $A_\mathrm{k}$ is given by
\begin{align}
    A_\mathrm{k}C_0^\frac{1}{2}
    &=
    \mathsf{RightSingular}_{r_\mathrm{k}}\big[
    C_0^\frac{1}{2}
    I_{d\times (d+1)}
    (
    \sum_i 
    G_{\mathrm{k},i}
    )
    I_{(d+1)\times d}
    C_0^\frac{1}{2}
    \big]
    \\
    &=
    \mathsf{RightSingular}_{r_\mathrm{q}}\big[
    \sum_i
    C_0^\frac{1}{2}
    W_{\mathrm{k},i}^\top
    W_{\mathrm{q},i}C_0
    A_\mathrm{q}^\top
    A_\mathrm{q}
    C_0
    W_{\mathrm{q},i}^\top 
    W_{\mathrm{k},i}
    C_0^\frac{1}{2}
    \notag\\
    &\qquad
    +
    \sum_i
    C_0^\frac{1}{2}
    W_{\mathrm{k},i}^\top
    (W_{\mathrm{q},i} \mu + b_{\mathrm{q},i})
    (W_{\mathrm{q},i} \mu + b_{\mathrm{q},i})^\top
    W_{\mathrm{k},i}
    C_0^\frac{1}{2}
    \big]
    .
\end{align}

where
\begin{align}
    G_{\mathrm{k},i} &=
    W_{\mathrm{k},i}'^\top
    W_{\mathrm{q},i}' 
    \tilde{C}
    A_\mathrm{q}'^\top
    (A_\mathrm{q}' \tilde{C} A_\mathrm{q}'^\top)^+
    A_\mathrm{q}'
    \tilde{C} W_{\mathrm{q},i}'^\top
    W_{\mathrm{k},i}'
    \\
    &=
    \begin{bmatrix}
        W_{\mathrm{k},i}^\top \\
        b_{\mathrm{k},i}^\top
    \end{bmatrix}
    W_{\mathrm{q},i}C_0
    A_\mathrm{q}^\top
    A_\mathrm{q}
    C_0
    W_{\mathrm{q},i}^\top 
    \begin{bmatrix}
        W_{\mathrm{k},i} &
        b_{\mathrm{k},i}
    \end{bmatrix}
    +
    \begin{bmatrix}
        W_{\mathrm{k},i}^\top \\
        b_{\mathrm{k},i}^\top
    \end{bmatrix}
    (W_{\mathrm{q},i} \mu + b_{\mathrm{q},i})
    (W_{\mathrm{q},i} \mu + b_{\mathrm{q},i})^\top
    \begin{bmatrix}
        W_{\mathrm{k},i} &
        b_{\mathrm{k},i}
    \end{bmatrix}
    .
\end{align}

\subsection{Grouped Query Attention (GQA)}
\label{sec:gqa}

MHA (e.g., for Llama-2) uses $h$-heads for query, key, and value.
However, Llama-3 uses grouped query attention (GQA), where the number of heads for key and value are smaller than the number of heads for query.
Let $n_\mathrm{q}$ be the query group size.
Then, the number of query heads is $n_\mathrm{q} h$, whereas $h$ is the number of KV heads.
Suppose $n_\mathrm{q}$ is the integer so that simple repetition can be used.
Q and K projections:
\begin{align}
    W_\mathrm{q} &=
    \begin{bmatrix}
    W_{\mathrm{q},1} \\
    W_{\mathrm{q},2} \\
    \vdots \\
    W_{\mathrm{q},n_\mathrm{q}h}     
    \end{bmatrix}
    \in\mathbb{R}^{n_\mathrm{q}hd'\times d}
    ,
    \qquad
    W_\mathrm{k} =
    \begin{bmatrix}
    W_{\mathrm{k},1} \\
    W_{\mathrm{k},2} \\
    \vdots \\
    W_{\mathrm{k},h}     
    \end{bmatrix}
    \in
    \mathbb{R}^{hd'\times d},
\end{align}
for $W_{\mathrm{q},i}\in\mathbb{R}^{d'\times d}$, $W_{\mathrm{k},i}\in\mathbb{R}^{d'\times d}$ with the head dimension $d'$.
Llama-3 uses  repeat-interleave to match the number of heads by repeating the KV projections $n_\mathrm{q}$-times:
\begin{align}
    W_\mathrm{k}' &=
    \begin{bmatrix}
    W_{\mathrm{k},1}  \\
    W_{\mathrm{k},1}  \\
    \vdots \\
    W_{\mathrm{k},1}  \\
    \vdots \\
    W_{\mathrm{k},h}         
    \end{bmatrix}
    \in 
    \mathbb{R}^{n_\mathrm{q}hd'\times d}.
\end{align}

For such GQA, we have attention map for the $j$th head in the $i$the group ($j\in\mathbb{Z}^+_{h}$, $i\in\mathbb{Z}^+_{n_\mathrm{q}}$):
\begin{align}
    M_{i,j} &=
    X^\top W_{\mathrm{q},i,j}^\top
    W_{\mathrm{k},i}X,
\end{align}
where we use an index convention: $W_{\mathrm{q},i,j}=W_{\mathrm{q},in_\mathrm{q}+j}$.

Consider the loss:
\begin{align}
    \mathcal{L} &=
    \sum_{i,j}
    \| M_{i,j} - \hat{M}_{i,j} \|^2
    \\
    &=
    \sum_{i,j} 
    \| X^\top 
    \underbrace{(
    \overbrace{
    W_{\mathrm{q},i,j}^\top
    W_{\mathrm{k},i}
    }^{G_{i,j}\in\mathbb{R}^{d\times d}}
    -
    A_\mathrm{q}^\top
    \overbrace{
    B_{\mathrm{q},i,j}^\top
    B_{\mathrm{k},i}
    }^{H_{i,j}\in\mathbb{R}^{r_\mathrm{q}\times r_\mathrm{k}}}
    A_{\mathrm{k}}    
    )
    }_{\varDelta_{i,j}\in\mathbb{R}^{d\times d}}
    X
    \|^2    
    \\
    &=
    \sum_{i,j} 
    \| C^\frac{1}{2} 
    G_{i,j}
    C^\frac{1}{2}
    -
    C^\frac{1}{2}
    A_\mathrm{q}^\top
    H_{i,j}
    A_{\mathrm{k}}    
    C^\frac{1}{2}
    \|^2    
    .
\end{align}
Hence the solution can be obtained with HOSVD likewise MHA in Sec.~\ref{sec:mha_cov}.

\section{Positional Encoding}
\label{sec:pe}

\subsection{Additive PE}

Consider additive PE for a token $X\in\mathbb{R}^{d\times l}$:
\begin{align}
    X' &= X + E,
\end{align}
where $E\in\mathbb{R}^{d\times l}$ is a positional embedding matrix.
Often it is sinusoidal like
\begin{align}
    E_{i,j} &= \exp(\jmath 2\pi f_i j / l),
\end{align}
with a predefined frequency $f_i$ for $i\in\mathbb{Z}_{d}$.
Note that complex rotation is not used in typical case, and instead split into $\cos$ and $\sin$.
Many work also considered trainable PE~\cite{radford2018improving, kenton2019bert}.

In this additive PE case, the solution is same by replacing the correlation matrix $C$ with
\begin{align}
    C' &= \mathbb{E}_{X}[X'X'^\top]
    \\
    &=
    \mathbb{E}_X[(X+E)(X+E)^\top]
    \\
    &=
    C + EE^\top 
    + \mathbb{E}_X[XE^\top + EX^\top].
\end{align}
For zero-mean token case, is reduces to $C+EE^\top$.
For static token case, we may use $(X+E)(X+E)^\top$ directly.

Nevertheless, some PE methods~\cite{dai1901transformer} use different additive PE for query and key individually:
\begin{align}
    X_\mathrm{q} &= X + E_\mathrm{q},
    \\
    X_\mathrm{k} &= X + E_\mathrm{k}.
\end{align}
In this case, the attention map will have bias terms:
\begin{align}
    M_i &= X_\mathrm{q}^\top 
    W_{\mathrm{q},i}^\top W_{\mathrm{k},i} 
    X_\mathrm{k}
    \\
    &=
    X^\top G_i X
    + 
    X^\top G_i E_\mathrm{k}
    +
    E_\mathrm{q}^\top G_i X
    +
    E_\mathrm{q}^\top G_i E_\mathrm{k}.
\end{align}
There are many variants to relax them or generalize them.

Consider loss:
\begin{align}
\mathcal{L} &=
\sum_i 
\|
X_\mathrm{q}^\top \varDelta_i X_\mathrm{k}
\|^2
\\
&=
\sum_i 
\mathrm{tr}
[
\varDelta_i 
\underbrace{
X_\mathrm{k}
X_\mathrm{k}^\top
}_{C_\mathrm{k}\in\mathbb{R}^{d\times d}}
\varDelta_i^\top
\underbrace{
X_\mathrm{q}
X_\mathrm{q}^\top 
}_{C_\mathrm{q}\in\mathbb{R}^{d\times d}}
]
\\
&=
\sum_i 
\mathrm{tr}
[
\varDelta_i 
C_\mathrm{k}
\varDelta_i^\top
C_\mathrm{q}
]
\\
&=
\sum_i 
\mathrm{tr}
[
\varDelta_i 
C_\mathrm{k}^\frac{1}{2}
C_\mathrm{k}^\frac{1}{2}
\varDelta_i^\top
C_\mathrm{q}^\frac{1}{2}
C_\mathrm{q}^\frac{1}{2}
]
\\
&=
\sum_i 
\mathrm{tr}
[
C_\mathrm{q}^\frac{1}{2}
\varDelta_i 
C_\mathrm{k}^\frac{1}{2}
C_\mathrm{k}^\frac{1}{2}
\varDelta_i^\top
C_\mathrm{q}^\frac{1}{2}
]
\\
&=
\sum_i 
\|
C_\mathrm{q}^\frac{1}{2}
\varDelta_i 
C_\mathrm{k}^\frac{1}{2}
\|^2
\\
&=
\sum_i 
\|
\underbrace{
C_\mathrm{q}^\frac{1}{2}
G_i
C_\mathrm{k}^\frac{1}{2}
}_{G_i'}
-
\underbrace{
C_\mathrm{q}^\frac{1}{2}
A_\mathrm{q}^\top
}_{A_\mathrm{q}'^\top}
H_i
\underbrace{
A_\mathrm{k}
C_\mathrm{k}^\frac{1}{2}
}_{A_\mathrm{k}'}
\|^2.
\end{align}
Hence, we can still solve it with HOSVD.

\subsection{Concatenative PE}

Another PE uses concatenation:
\begin{align}
    Q_i &= 
    \begin{bmatrix}
    W_{\mathrm{q},i}X \\
    E_{\mathrm{q},i}
    \end{bmatrix},
\\
    K_i &= 
    \begin{bmatrix}
    W_{\mathrm{k},i}X \\
    E_{\mathrm{k},i}
    \end{bmatrix}
    .
\end{align}
Then, the attention map will be
\begin{align}
M_i &= 
Q_i^\top K_i \\
&=
X^\top G_i X
+ 
E_{\mathrm{q},i}^\top E_{\mathrm{k},i},
\end{align}
which has just a bias term $E_\mathrm{q}^\top E_\mathrm{k}$ and there is no impact in loss function with low-rank approximation.

\subsection{Multiplicative PE}

Consider a multiplicative PE for token $X$:
\begin{align}
    X'=X \odot E,
\end{align}
where $\odot$ denotes Hadamard product.
We just need to replace the correlation with $C'=X'X'^\top$ to solve in a straightforward manner.

However, rotary PE (RoPE)~\cite{su2024roformer} uses multiplicative PE on query and key, not token $X$.
More precisely, we can represent per token:
\begin{align}
    q_{i,m} &= \Theta_{i,m} W_{\mathrm{q},i} x_m,
    \\
    k_{i,m} &= \Theta_{i,m} W_{\mathrm{k},i} x_m,
\end{align}
where $\Theta_{i,m}$ is a block diagonal rotation matrix for $i$th head and $m$th token, such that $\Theta_{i,m}^\top \Theta_{i,n} =\Theta_{i,n-m}$.

For example, Llama-2 uses the same RoPE for all heads with block rotation:
\begin{align}
    \Theta_{i,m} &=
    \begin{bmatrix}
        \cos(m\Phi) & - \sin(m\Phi) \\
        \sin(m\Phi) & \cos(m\Phi)
    \end{bmatrix}
    \in\mathbb{R}^{d/h\times d/h},
    \\
    \Phi &=
    \mathrm{diag}\big[
    \{\theta^{-2ih/d}\}_{i=0}^{d/2h-1}
    \big] 
    \in\mathbb{R}^{d/2h\times d/2h},
\end{align}
with a base rope theta of $\theta=10^4$.

We have the loss:
\begin{align}
    \mathcal{L} &=
    \mathbb{E}_{X}
    \sum_{i,m,n} \| q_{i,m}^\top k_{i,m} 
    -\hat{q}_{i,m}^\top \hat{k}_{i,m}
    \|^2
    \\
    &=
    \mathbb{E}_{X}
    \sum_{i,m,n} 
    \|
    x_m^\top 
    \underbrace{
    (W_{\mathrm{q},i}^\top
    \Theta_{i,n-m} W_{\mathrm{k},i}
    - 
    A_\mathrm{q}^\top B_{\mathrm{q},i}^\top
    \Theta_{i,n-m} 
    B_{\mathrm{k},i}
    A_\mathrm{k}
    )
    }_{\varDelta_{i,n-m}\in\mathbb{R}^{d\times d}}
    x_n
    \|^2
    \\
    &=
    \mathbb{E}_{X}
    \sum_{i,m,n} 
    \mathrm{tr}[
    \varDelta_{i,n-m}
    x_n
    x_n^\top
    \varDelta_{i,n-m}^\top
    x_m 
    x_m^\top 
    ]
    \\
    &=
    \sum_{i,m,n} 
    \mathrm{tr}[
    \varDelta_{i,n-m}
    C
    \varDelta_{i,n-m}^\top
    C 
    ]
    \\
    &=
    \sum_{i,m,n} 
    \| 
    C^\frac{1}{2}
    \varDelta_{i,n-m}
    C^\frac{1}{2}
    \|^2
    \\
    &=
    \sum_{i,m,n} 
    \| 
    \underbrace{
    C^\frac{1}{2}
    W_{\mathrm{q},i}^\top
    \Theta_{i,n-m} W_{\mathrm{k},i}
    C^\frac{1}{2}
    }_{W_{i,n-m}'}
    - 
    \underbrace{
    C^\frac{1}{2}
    A_\mathrm{q}^\top
    }_{A_\mathrm{q}'^\top}
    \underbrace{
    B_{\mathrm{q},i}^\top
    \Theta_{i,n-m} 
    B_{\mathrm{k},i}
    }_{H_{i,n-m}}
    \underbrace{
    A_\mathrm{k}
    C^\frac{1}{2}
    }_{A_\mathrm{k}'}
    \|^2
\end{align}
where we assumed $x_m$ and $x_n$ are independent. 
Then, we can solve it with HOSVD.
However, considering all token lengths over $m$ and $n$ is not practical, and we may need to consider attention windows such as $|n-m|\leq5$ to optimize.
When a causal mask is used, we do not need to sum over $m>n$ but only $m\geq n$.

NOTE: we can generalize RoPE with other unitary rotations.

\begin{figure}
    \centering
    \includegraphics[width=0.8\linewidth]{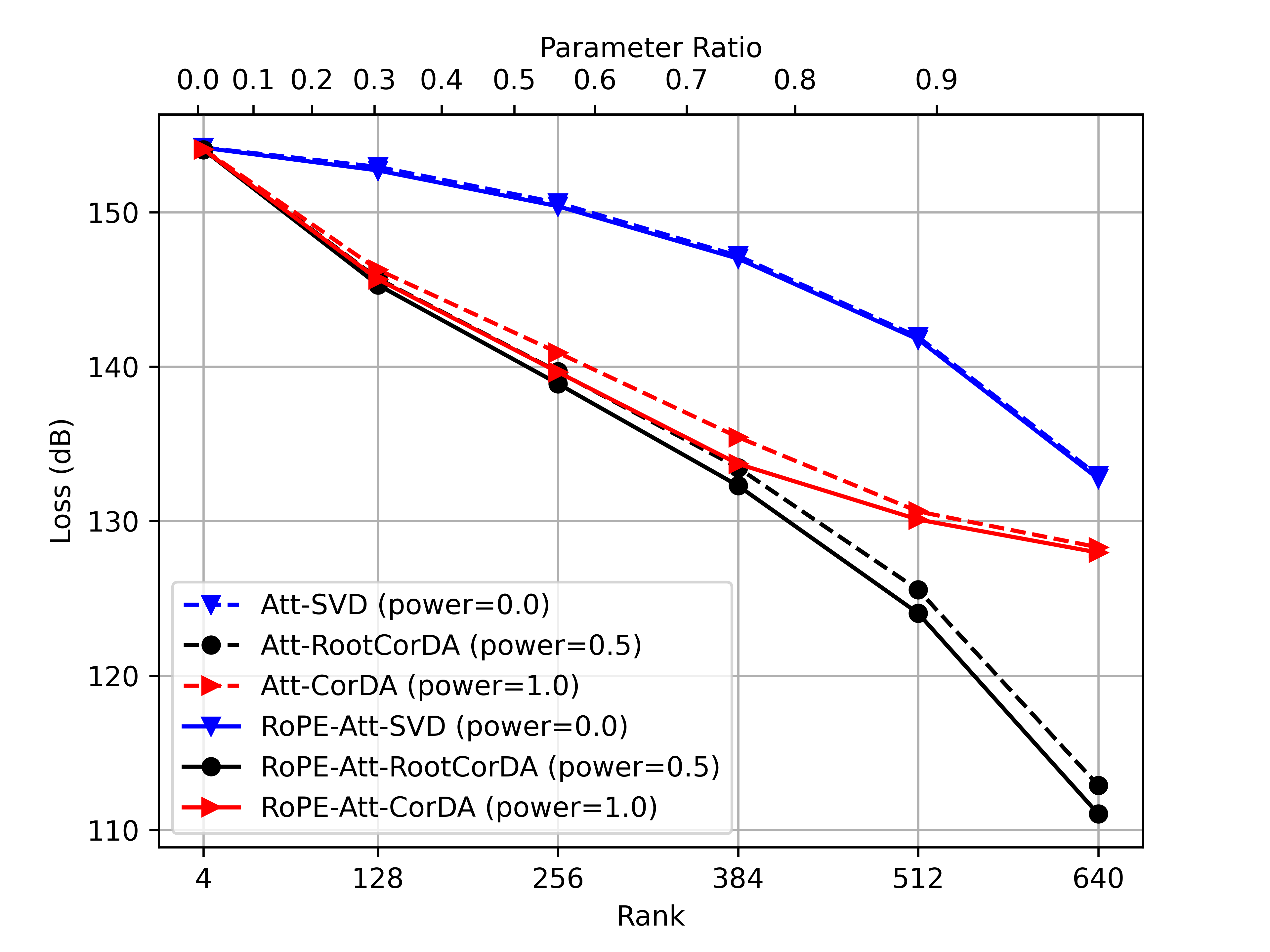}
    \caption{RoPE-Attention-Aware Distillation: 10-token window.}
    \label{fig:rope}
\end{figure}

Fig.~\ref{fig:rope} shows the result of HOSVD with/without RoPE consideration.
The loss is calculated over 10-token window, with RoPE base theta of $10^4$, used in Llama-2, while the hiiden size is still $768$.
HOSVD without RoPE consideration was already a good approximation as it is optimal at diagonal token.
RoPE-aware HOSVD offers additional 1--2 dB gain.

\section{Joint Value-Output Compression}
\label{sec:vo}

Many LLMs have output projection after QKV attention.
The attention output will be
\begin{align}
    Y= \sum_i W_{\mathrm{o},i} 
    W_{\mathrm{v},i} X \sigma(M_i^\top),
\end{align}
where $W_{\mathrm{o},i}\in\mathbb{R}^{d\times d/h}$ is the $i$th head output projection, and $Y\in\mathbb{R}^{d\times l}$ is the attention output.
This motivates us to optimize value projection and output projection jointly.

NOTE: $W_{\mathrm{o},i} W_{\mathrm{v},i}$ can be triangularized by LU factorization to save the number of parameters from $2d^2$ to $2d^2-d^2/h$ without any performance loss.

As $\sigma(M_i)$ is just weighting $X$, we may assume that the statistics still holds as $\mathbb{E}[X\sigma(M_i)\sigma(M_i)^\top X^\top]=C$ for uncorrelated tokens.
Hence, we end up with optimizing
\begin{align}
    \mathcal{L} &=
    \| W_\mathrm{o} W_\mathrm{v} C^\frac{1}{2} -B_\mathrm{o}
    \underbrace{A_\mathrm{o}B_\mathrm{v}}_{H} A_\mathrm{v} C^\frac{1}{2}\|^2.
\end{align}
Hence, both projections can be combined together.

Nevertheless, when we consider minimizing individual head projection loss for arbitrary attention weights:
\begin{align}
    \mathcal{L} &=
    \sum_i \| 
    \underbrace{
    W_{\mathrm{o},i} W_{\mathrm{v},i} C^\frac{1}{2}
    }_{G_i\in\mathbb{R}^{d\times d}}
    -
    B_\mathrm{o}
    \underbrace{
    A_{\mathrm{o},i} B_{\mathrm{v},i}
    }_{H_i\in\mathbb{R}^{r_\mathrm{o}\times r_\mathrm{v}}}
    \underbrace{
    A_\mathrm{v}
    C^\frac{1}{2}
    }_{A_\mathrm{v}'}
    \|^2.
\end{align}
Then the solution is HOSVD:
\begin{align}
    B_\mathrm{o} &=
    \mathrm{RightSingular}_{r_\mathrm{o}}[
    \sum_i G_i A_\mathrm{v}'^\top 
    A_\mathrm{v}' G_i^\top
    ]
    ,
    \\
    A_\mathrm{v}' &=
    \mathrm{RightSingular}_{r_\mathrm{v}}[
    \sum_i G_i^\top B_\mathrm{o} 
    B_\mathrm{o}^\top G_i
    ]
    ,
    \\
    A_{\mathrm{o},i} &=
    B_{\mathrm{o}}^\top W_{\mathrm{o},i}
    J_i
    \\
    B_{\mathrm{v},i} &=
    J_i^+
    W_{\mathrm{v},i}
    A_{\mathrm{v}}'^\top     ,
\end{align}
for arbitrary full-rank matrix $J_i\in\mathbb{R}^{d/h\times d/h}$.
Selecting $J_i$ can save the number of parameters by up to $d/h\times d/h$.

\subsection{Bias Update}
Some LLMs such as OPT uses bias in QKVO.
Let's consider bias impact.
The attention output will be:
\begin{align}
    Y &=
    \sum_i W_{\mathrm{o},i} ( W_{\mathrm{v},i} X + b_{\mathrm{v},i} 1^\top)
    \sigma(M_i)
    +
    b_{\mathrm{o},i} 1^\top
    \\
    &=
    \sum_i
    W_{\mathrm{o},i}W_{\mathrm{v},i}X\sigma(M_i)
    +
    W_{\mathrm{o},i}b_{\mathrm{v},i}1^\top
    \sigma(M_i)
    +
    b_{\mathrm{o},i}1^\top.
\end{align}

Considering any arbitrary attention map $M_i$, we may want to optimize:
\begin{align}
    \mathcal{L} &=
    \sum_i
    \|     
    W_{\mathrm{o},i}
    (
    W_{\mathrm{v},i}X
    +
    b_{\mathrm{v},i}1^\top
    )
    +b_{\mathrm{o},i}1^\top
    -
    \hat{W}_{\mathrm{o},i}
    (
     \hat{W}_{\mathrm{v},i}X
    +
    \hat{b}_{\mathrm{v},i}1^\top
    )
    -\hat{b}_{\mathrm{o},i}1^\top
    \|^2
    .
\end{align}
The gradient with respective to $\hat{b}_{o}$ is given
\begin{align}
    -
    \big(
    W_{\mathrm{o},i}
    (
    W_{\mathrm{v},i}X
    +
    b_{\mathrm{v},i}1^\top
    )
    +b_{\mathrm{o},i}1^\top
    -
    \hat{W}_{\mathrm{o},i}
    (
     \hat{W}_{\mathrm{v},i}X
    +
    \hat{b}_{\mathrm{v},i}1^\top
    )
    -\hat{b}_{\mathrm{o},i}1^\top
    \big)
    1.
\end{align}
Thus the KKT condition gives:
\begin{align}
    \hat{b}_{\mathrm{o},i} &=
    b_{\mathrm{o},i} +
    W_{\mathrm{o},i}(W_{\mathrm{v},i}\mu+b_{\mathrm{v},i})
    -
    \hat{W}_{\mathrm{o},i}(\hat{W}_{\mathrm{v},i}\mu+\hat{b}_{\mathrm{v},i}).
\end{align}

Plugging into the loss gives:
\begin{align}
    \mathcal{L} &=
    \sum_i \|
    W_{\mathrm{o},i}
    W_{\mathrm{v},i}(X-\mu 1^\top)
    -
    \hat{W}_{\mathrm{o},i}
     \hat{W}_{\mathrm{v},i}
     (X-\mu 1^\top)
    \|^2
    \\
    &=
    \sum_i \|
    \underbrace{
    W_{\mathrm{o},i}
    W_{\mathrm{v},i}
    }_{G_i\in\mathbb{R}^{d\times d}}
    C_0^\frac{1}{2}
    -
    {B}_{\mathrm{o}}
    \underbrace{
    {A}_{\mathrm{o},i}
    {B}_{\mathrm{v},i}
    }_{H_i\in\mathbb{R}^{r_\mathrm{o}\times r_\mathrm{v}}}
    {A}_{\mathrm{v}}
     C_0^\frac{1}{2}
    \|^2
    .
\end{align}
Here, $C_0 = (X-\mu 1^\top)(X-\mu 1^\top)^\top$ is centered covariance (it can be normalized).
Hence, this is solved by HOSVD:
\begin{align}
    B_\mathrm{o} &= 
    \mathsf{RightSingular}_{r_\mathrm{o}}
    \big[ 
    \sum_i G_i C_0 A_\mathrm{v}^\top A_\mathrm{v} C_0 G_i^\top
    \big],
    \\
    A_\mathrm{v}C_0^\frac{1}{2} &= 
    \mathsf{RightSingular}_{r_\mathrm{o}}
    \big[ 
    \sum_i C_0^\frac{1}{2} 
    G_i^\top B_\mathrm{o} B_\mathrm{o}^\top 
    G_i C_0^\frac{1}{2}
    \big].
\end{align}
Note that $\hat{b}_{\mathrm{v}}$ has no impact as it can be absorbed by $\hat{b}_\mathrm{o}$.
Hence, we can keep the original bias or changed to zero bias.

\subsection{Attention-Aware Joint VO Compression}

The output projection module takes the input token:
\begin{align}
X_{\mathrm{o},i} &=
W_{\mathrm{v},i}X\sigma(M_i^\top).
\end{align}
The covariance of the token is
\begin{align}
    C_{\mathrm{o},i} &=
    X_{\mathrm{o},i} X_{\mathrm{o},i}^\top
    \\
    &=
    W_{\mathrm{v},i}X\sigma(M_i^\top)\sigma(M_i)W_{\mathrm{v},i}^\top.
\end{align}
Over the heads, we have cross-correation terms:
\begin{align}
    X_\mathrm{o} &=
    \begin{bmatrix}
        W_{\mathrm{v},1} X \sigma(M_1^\top) \\
        W_{\mathrm{v},2} X \sigma(M_2^\top) 
        \\
        \vdots\\
        W_{\mathrm{v},h} X \sigma(M_h^\top) 
    \end{bmatrix}
    \\
    &=
    \underbrace{
    \mathsf{diag}\big[
    W_{\mathrm{v},1},
    W_{\mathrm{v},2},
    \ldots,
    W_{\mathrm{v},h}
    \big]
    }_{W_\mathrm{v}'\in\mathbb{R}^{hd_\mathrm{h}\times hd}}
    \underbrace{
    (I_h \otimes X)
    \begin{bmatrix}
        \sigma(M_1^\top) \\
        \sigma(M_2^\top)\\
        \vdots \\
        \sigma(M_h^\top)
    \end{bmatrix}
    }_{X'\in\mathbb{R}^{hd\times l}}
    .
\end{align}
We consider using the covariance of output projection module not value projection module instead.
The covariance of the output projection $C_\mathrm{o}\in\mathbb{R}^{hd_\mathrm{h}\times hd_\mathrm{h}}$ is given as
\begin{align}
    C_\mathrm{o} &= X_\mathrm{o} X_\mathrm{o}^\top
    \\
    &=
    W_\mathrm{v}' \underbrace{X'X'^\top}_{C_\mathrm{v}\in\mathbb{R}^{hd\times hd}} W_\mathrm{v}'^\top
    .
\end{align}
Using this attention-aware token statistics $C_\mathrm{v}$ can be more accurate to optimize, rather than simple token statistics $C$.

The value projection module takes the input token $X$ typically.
However, there is no impact when we instead take the attention-weighted token for each head before value projection: $X'$.
Even though we have no statistics on this, we can predict it from $C_\mathrm{o}$ as $C_o = W_\mathrm{v}' C_\mathrm{v} W_\mathrm{v}'^\top$:
\begin{align}
    C_\mathrm{v} &=     W_\mathrm{v}'^+ C_\mathrm{o} [W_\mathrm{v}'^+]^\top.
\end{align}
Note that this is at most the rank of $hd_\mathrm{h}$.

The loss will be
\begin{align}
    \mathcal{L} &=
    \|
    \sum_i
    W_{\mathrm{o},i} W_{\mathrm{v},i} X
    \sigma(M_i^\top)
    -
    \hat{W}_{\mathrm{o},i} \hat{W}_{\mathrm{v},i} X
    \sigma(M_i^\top)    
    \|^2
    \\
    &=
    \|
    W_\mathrm{o}
    W_\mathrm{v}'
    X'
    -
    \hat{W}_\mathrm{o}
    \hat{W}_\mathrm{v}'
    X'
    \|^2
    \\
    &=
    \|
    W_\mathrm{o}
    W_\mathrm{v}'
    C_\mathrm{v}^\frac{1}{2}
    -
    \hat{W}_\mathrm{o}
    \hat{W}_\mathrm{v}'
    C_\mathrm{v}^\frac{1}{2}
    \|^2
    \\
    &=
    \|
    W_\mathrm{o}
    C_\mathrm{o}^\frac{1}{2}
    -
    \hat{W}_\mathrm{o}
    \hat{W}_\mathrm{v}'
    {W}_\mathrm{v}'^+
    C_\mathrm{o}^\frac{1}{2}
    \|^2
    .
\end{align}
Here we have
\begin{align}
    \hat{W}_\mathrm{v}' W_\mathrm{v}'^+
    &=
    \mathsf{diag}[
    \hat{W}_{\mathrm{v},1}W_{\mathrm{v},1}^\top
    (
    W_{\mathrm{v},1}
    W_{\mathrm{v},1}^\top
    )^+
    ,
    \hat{W}_{\mathrm{v},2}W_{\mathrm{v},2}^\top
    (
    W_{\mathrm{v},2}
    W_{\mathrm{v},2}^\top
    )^+
    \ldots,
    \hat{W}_{\mathrm{v},h}W_{\mathrm{v},h}^\top
    (
    W_{\mathrm{v},h}
    W_{\mathrm{v},h}^\top
    )^+
    ]    
    \in\mathbb{R}^{hd_\mathrm{h}\times hd_\mathrm{h}}.
\end{align}

We write:
\begin{align}
    \mathcal{L}
    &=
    \big\| 
    \sum_i W_{\mathrm{o},i}
    [C_{\mathrm{o}}^\frac{1}{2}]_i
    -
    B_{\mathrm{o}}
    \underbrace{
    A_{\mathrm{o},i}
    B_{\mathrm{v},i}
    }_{H_i\in\mathbb{R}^{r_\mathrm{o}\times r_\mathrm{v}}}
    A_{\mathrm{v}}
    W_{\mathrm{v},i}^+
    [C_0^\frac{1}{2}]_i
    \big\|^2
    \\
    &=
    \big\| 
    W_{\mathrm{o}}
    C_{\mathrm{o}}^\frac{1}{2}
    -
    B_{\mathrm{o}}
    \sum_i
    {H_i}
    A_{\mathrm{v}}
    W_{\mathrm{v},i}^+
    [C_0^\frac{1}{2}]_i
    \big\|^2
    \\
    &=
    \big\| 
    W_{\mathrm{o}}
    C_{\mathrm{o}}^\frac{1}{2}
    -
    B_{\mathrm{o}}
    \underbrace{
    \begin{bmatrix}
        H_1 & \cdots & H_h
    \end{bmatrix}
    }_{H\in\mathbb{R}^{r_\mathrm{o}\times hr_\mathrm{v}}}
    (I_h \otimes A_\mathrm{v})
    \mathsf{diag}[
    W_{\mathrm{v},1}^+ 
    , 
    \ldots
    W_{\mathrm{v},h}^+ 
    ]
    C_0^\frac{1}{2}
    \big\|^2
    .
\end{align}
Note that $H_i$ is of rank up to $\min(r_\mathrm{o}, r_\mathrm{v},d_\mathrm{h})$.

Gradient:
\begin{align}
    \nabla_{H} \mathcal{L}
    &=
    -
    \Big(
    B_\mathrm{o}
    \Big)^\top
    \Big(
    W_\mathrm{o} C_\mathrm{o}^\frac{1}{2}
    -
    B_{\mathrm{o}}
    H
    (I_h \otimes A_{\mathrm{v}})
    W_{\mathrm{v}}'^+
    C_0^\frac{1}{2}
    \Big)
    \Big(
    (I_h \otimes A_{\mathrm{v}})
    W_{\mathrm{v}}'^+
    C_0^\frac{1}{2}
    \Big)^\top
    ,
    \\
    \nabla_{A_{\mathrm{o},j}} \mathcal{L}
    &=
    -
    \Big(
    B_\mathrm{o}
    \Big)^\top
    \Big(
    W_\mathrm{o} C_\mathrm{o}^\frac{1}{2}
    -
    B_{\mathrm{o}}
    H
    (I_h \otimes A_{\mathrm{v}})
    W_{\mathrm{v}}'^+
    C_0^\frac{1}{2}
    \Big)
    \Big(
    B_{\mathrm{v},j}
    A_{\mathrm{v}}
    W_{\mathrm{v},j}^+
    [C_0^\frac{1}{2}]_j
    \Big)^\top
    ,
    \\
    \nabla_{B_{\mathrm{v},j}} \mathcal{L}
    &=
    -
    \Big(
    B_\mathrm{o}
    A_{\mathrm{o},j}
    \Big)^\top
    \Big(
    W_\mathrm{o} C_\mathrm{o}^\frac{1}{2}
    -
    B_{\mathrm{o}}
    H
    (I_h \otimes A_{\mathrm{v}})
    W_{\mathrm{v}}'^+
    C_0^\frac{1}{2}
    \Big)
    \Big(
    A_{\mathrm{v}}
    W_{\mathrm{v},j}^+
    [C_0^\frac{1}{2}]_j
    \Big)^\top
    ,
    \\
    \nabla_{B_\mathrm{o}} \mathcal{L}
    &=
    -
    \Big(
    W_\mathrm{o} C_\mathrm{o}^\frac{1}{2}
    -
    B_{\mathrm{o}}
    H
    (I_h \otimes A_{\mathrm{v}})
    W_{\mathrm{v}}'^+
    C_0^\frac{1}{2}
    \Big)
    \Big(
    H
    (I_h \otimes A_{\mathrm{v}})
    W_{\mathrm{v}}'^+
    C_0^\frac{1}{2}
    \Big)^\top,
    \\
    \nabla_{A_\mathrm{v}} \mathcal{L}
    &=
    -
    \sum_j
    \Big(
    B_\mathrm{o}
    H_j
    \Big)^\top
    \Big(
    W_\mathrm{o} C_\mathrm{o}^\frac{1}{2}
    -
    B_{\mathrm{o}}
    \sum_i
    H_i
    A_{\mathrm{v}}
    W_{\mathrm{v},i}^+
    [C_0^\frac{1}{2}]_i
    \Big)
    \Big(
    W_{\mathrm{v},j}^+
    [C_0^\frac{1}{2}]_j
    \Big)^\top
    .
\end{align}
The optimal $B_\mathrm{o}$ is the left-singular of $W_{\mathrm{o}} C_0^\frac{1}{2}$, having unitary condition: $B_\mathrm{o}^\top B_\mathrm{o}=I_{r_\mathrm{o}}$.

From the first KKT, we have a linear system to solve for $H$:
\begin{align}
    H
    \begin{bmatrix}
    A_\mathrm{v}W_{\mathrm{v},1}^+[C_0^\frac{1}{2}]_1\\
    \vdots\\
    A_\mathrm{v}W_{\mathrm{v},h}^+[C_0^\frac{1}{2}]_h        
    \end{bmatrix}
    \begin{bmatrix}
    A_\mathrm{v}W_{\mathrm{v},1}^+[C_0^\frac{1}{2}]_1\\
    \vdots\\
    A_\mathrm{v}W_{\mathrm{v},h}^+[C_0^\frac{1}{2}]_h        
    \end{bmatrix}^\top
    &=
    B_\mathrm{o}^\top W_\mathrm{o} C_\mathrm{o}^\frac{1}{2}
    \begin{bmatrix}
    A_\mathrm{v}W_{\mathrm{v},1}^+[C_0^\frac{1}{2}]_1\\
    \vdots\\
    A_\mathrm{v}W_{\mathrm{v},h}^+[C_0^\frac{1}{2}]_h        
    \end{bmatrix}^\top
    .    
\end{align}
Hence, we have
\begin{align}
    H &=
    B_\mathrm{o}^\top
    W_\mathrm{o}
    C_0
    [W_\mathrm{v}'^+]^\top
    (I_h \otimes A_\mathrm{v}^\top)
    \Big(
    (I_h \otimes A_\mathrm{v})
    W_\mathrm{v}'^+
    C_0
    [W_\mathrm{v}'^+]^\top
    (I_h \otimes A_\mathrm{v}^\top)
    \Big)^+
    .
\end{align}
Plugging into the loss, we have
\begin{align}
    \mathcal{L} &=
    \|W_\mathrm{o}C_0^\frac{1}{2}\|^2
    -
    \Big\|
    B_\mathrm{o}^\top
    W_\mathrm{o}C_0
    [W_\mathrm{v}'^+]^\top
    (I_h\otimes A_\mathrm{v})^\top
    \Big(
    (I_h\otimes A_\mathrm{v})
    W_\mathrm{v}'^+
    C_0
    [W_\mathrm{v}'^+]^\top
    (I_h\otimes A_\mathrm{v})^\top
    \Big)^{-\frac{1}{2}}
    \Big\|^2
\end{align}

The last KKT condition requires solving in vectorization:
\begin{align}
    \sum_{i,j}(G_j G_i^\top \otimes H_j^\top H_i)
    \mathsf{vec}[A_\mathrm{v}]
    &=
    \sum_j
    \mathsf{vec}[
    H_j B_\mathrm{o}^\top W_\mathrm{o}
    C^\frac{1}{2} G_j^\top
    ],
\end{align}
where $G_i\in\mathbb{R}^{d\times hd_\mathrm{h}}$ is defined:
\begin{align}
    G_i &= W_{\mathrm{v},i}^+[C_0^\frac{1}{2}]_i.
\end{align}

\section{MLP-Aware Joint Compression}
\label{sec:mlp}

SparseLLM~\cite{bai2024sparsellm} proposed the way to sparsify MLP layer in LLM models as it consumes two thirds of trainable parameters.
The key idea is to minimize the MLP loss, not local loss.
LLM uses typically 2-layer MLP:
\begin{align}
    z &= W_1 x + b_1,\\
    a &= \sigma(z),\\
    y &= W_2 a + b_2.
\end{align}
The first linear layer typically upsamples by a factor of four, and then the second linear layer downsamples to the same dimension. 
Activation-aware low-rank approximation can minimize loss individually for $z$ given $x$ and $y$ given $a$, but not the MLP output $y$ given $x$.

SparseLLM uses the closed-form solution to minimize:
\begin{align}
    \mathcal{L} &=
    \alpha \| W_1x +b_1 - z \|^2
    +
    \beta \| a-\sigma(z) \|^2
    +
    \gamma \| W_2 a + b_2 - y\|^2
    ,
\end{align}
for auxiliary variables $a$ and $z$, given pre-trained input $x$ and output $y$.

Optimizing $a$ can be obtained by ridge regression:
\begin{align}
    a^\star &=
    (\gamma W_2^\top W_2 + \beta I)^+ (\beta \sigma(z) + \gamma W_2^\top (y-b_2)).
\end{align}
Optimal $z$ can be also obtained closed-form way with case for ReLU:
\begin{align}
    z_- &=
    W_1 x + b_1,\\
    z_+ &=
    \frac{1}{\alpha+\beta} 
    (\alpha z_- 
    +\beta a),
\end{align}
depending on $[z]_i$'s sign.

The same approach can be used for low-rank approximation.
Given $z$, we can optimize low-rank matrix $\hat{W}_1=B_1 A_1$ by SVD of $(z-b_1)x^+ C_x^\frac{1}{2}$, where $(z-b_1)x^+=(z-b_1)x^\top C_x^+$ corresponds to the effective weight matrix to map $x$ onto $z$.
Given $a$, we approximate $\hat{W}_2=B_2A_2$ by SVD of $(y-b_2)a^+ C_a^\frac{1}{2}=(y-b_2)a^\top C_a^\frac{-1}{2}$, given correlation $C_a=a a^\top$.

\section{Sparse Matrix}
\label{sec:spa}

Consider low-rank plus sparse decomposition:
\begin{align}
    \hat{W} &=
    BA + D,
\end{align}
where $D\in\mathbb{R}^{d'\times d}$ is a sparse matrix such that $\|D\|_0 \leq \kappa$.
As discussed so far, given a $D$ matrix, the best low-rank matrices are SVD of $(W-D)C^\frac{1}{2}$.
Given $BA$, finding sparse $D$ is an NP-hard problem, and often it is solved by greedy or relaxed methods such as matching pursuit and proximal gradient.
Considering the $\ell_1$ relaxation, we have
\begin{align}
\mathcal{L}' &=
\| (D + BA-W)C^\frac{1}{2} \|^2
+
\lambda
(
\|D\|_1 - \kappa).
\end{align}
Fast iterative shrinkage-threshold algorithm (FISTA) uses iterations with Nesterov's accelerating technique:
\begin{align}
    D_k &=
    \mathcal{T}_{\lambda\mu_k}[
    D_{k-1} - 2\mu_k (D_{k-1} + BA - W) C
    ]
    ,
    \\
    \mu_{k+1} &=
    \frac{1}{2}
    (1 + \sqrt{1+4\mu_k^2})
    ,\\
    D_k &\leftarrow D_k +
    \frac{\mu_k - 1}{\mu_{k+1}}
    (D_k - D_{k-1}),
\end{align}
for iterations $k= 1,2,\ldots$ with a stepsize $\mu_1=1$.
$\mathcal{T}_\alpha$ is a soft shrinkage operator:
\begin{align}
    \mathcal{T}_\alpha[x] &=
        \mathrm{sign}[x](x - \alpha)_+ 
.
\end{align}
We may iterate SVD and FISTA.
The choice of $\lambda$ is crucial to have a target sparsity.
It is not easy to adjust $\lambda$ such that the target sparsity is achieved beforehand.

\begin{figure}[t]
\centering
\includegraphics[width=0.8\linewidth]{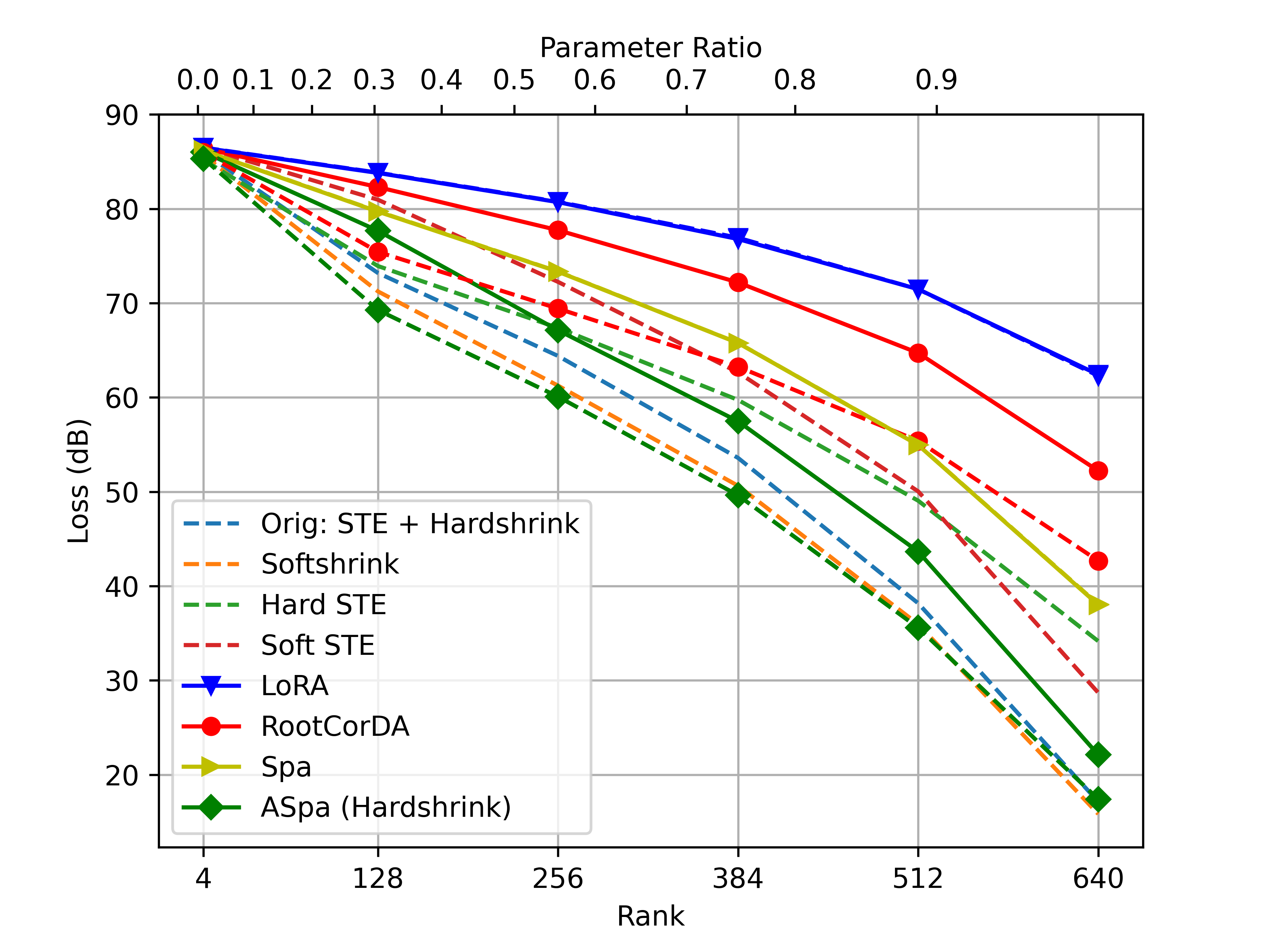}
\caption{Random weight approximation with/without correlation. 
Correlation is sampled from Wishart distribution with covariance of identity or off-diagonal decaying of $0.9$ factor.
Weight is normal distributed.
}
\label{fig:spa}
\end{figure}

Alternatively, we use a regular gradient method with straight-through estimator (STE):
\begin{align}
    D &= \underbrace{D - D.\mathrm{detach}}_{\mathrm{STE\ Trick}} + \mathcal{S}_\kappa[D.\mathrm{detach}],
\end{align}
where $\mathcal{S}_\kappa[\cdot]$ is a hard shrinkage operator, i.e., sparcification operator passing only $\kappa$ elements having largest magnitude.
This STE method has a benefit over FISTA: i) the sparsity can be specified; ii) any other loss function including the final downstream task loss can be incorporated; and iii) the quantization-aware training can be readily integrated in the STE projection.
Nevertheless, soft shrink and hard shrink are actually differentiable, and we may not need to use STE. 
Fig.~\ref{fig:spa} shows the comparison of STE and Hard/Softshrink. 
In this experiment, Hardshrink works best.

\begin{figure}[t]
\centering
\includegraphics[width=0.8\linewidth]{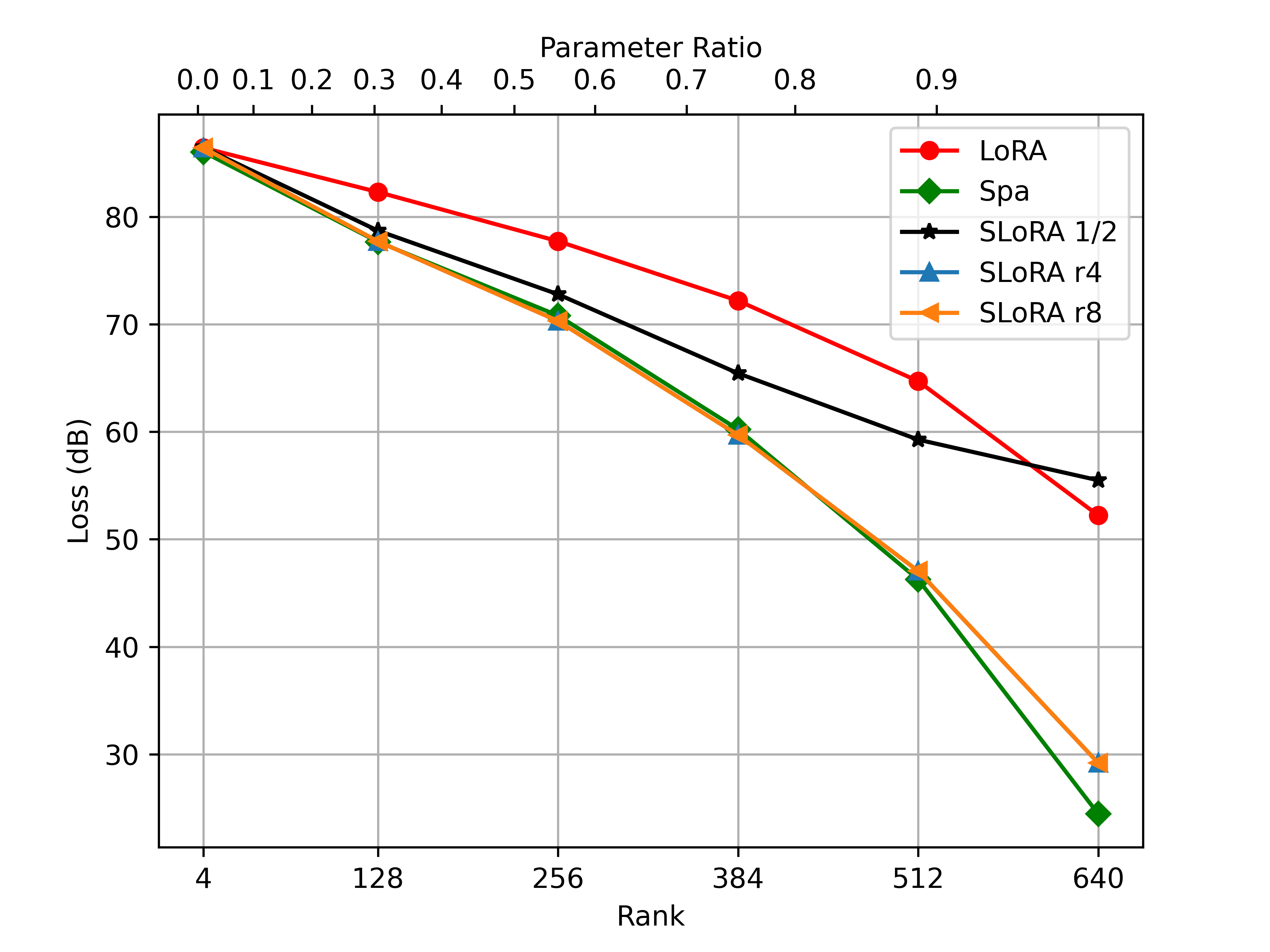}
\caption{Low-rank plus sparse approximation does not outperform sparse-alone approximation.}
\label{fig:slora}
\end{figure}

We also notice that sparse approximation can be better than low-rank approximation.
And, also joint low-rank plus sparse approximation did not work well as shown in Fig.~\ref{fig:slora}.

\begin{figure}[t]
\centering
\includegraphics[width=0.8\linewidth]{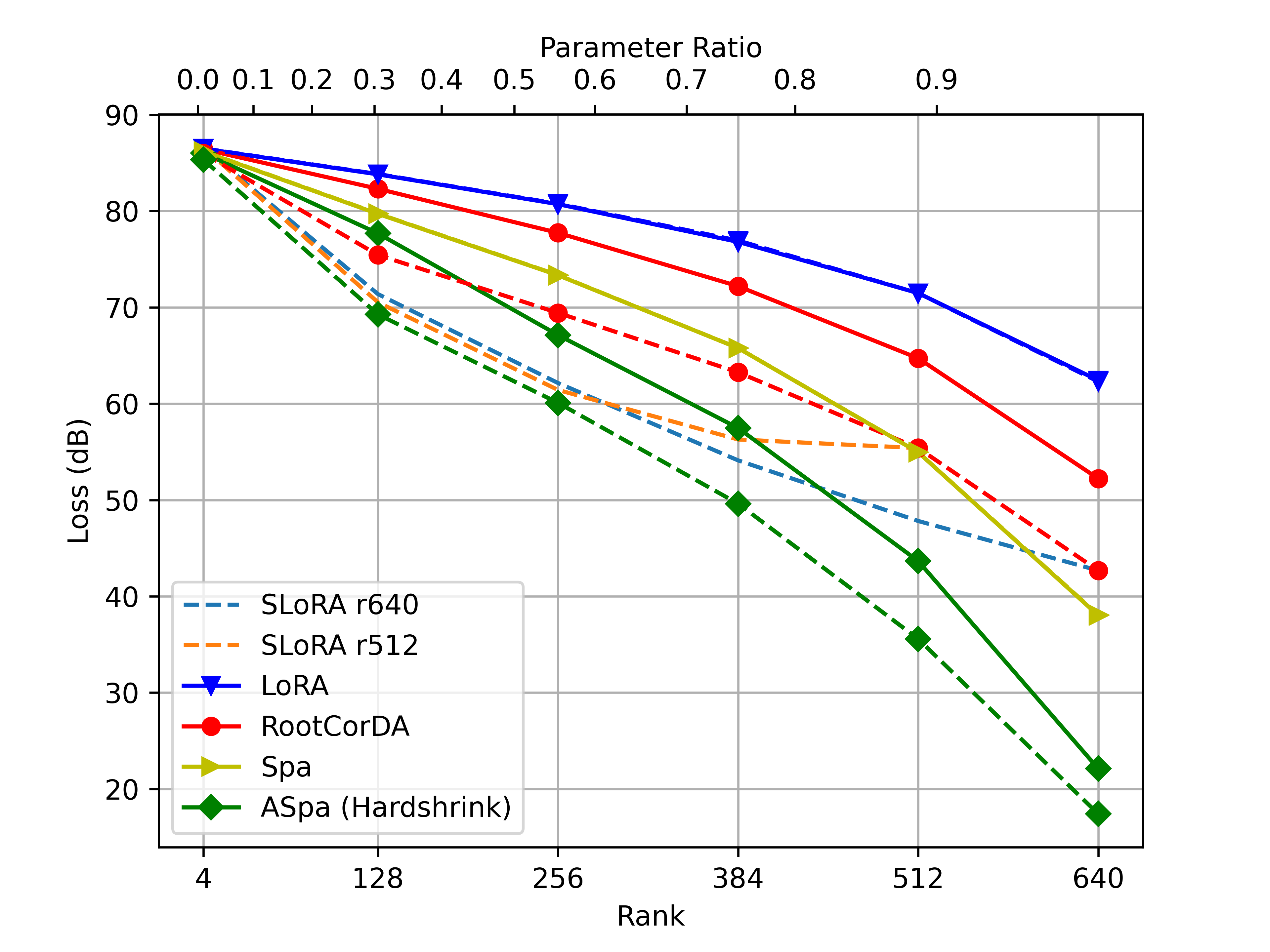}
\caption{Sparsification of $B$ and $A$ low-rank matrices.}
\label{fig:spa2}
\end{figure}

However, unstructured sparse matrix may require index storage to memorize the non-zero entry locations.
When we use a mask, it requires $d' d$ binary memory as well as non-zero values in $D$. 
When the sparsity is small, keeping index will be more efficient, i.e., keeping $\log_2(dd')\tau$.
Fig.~\ref{fig:spa2} shows the case with sparsification for low-rank adapter $B,A$, starting RootCorDA of rank 640 and 512.
Although sparsified low-rank approximation has a benefit, it does not outperform sparse approximation alone.

Another possibility using sparse approximation is to sparsify LoRA matrices $B$ and $A$.
However, doing so may be poor because the product of two sparse matrices can be much more sparse: e.g., 50\% sparse $B$ and $A$ will result in 25\% sparse $BA$.
Hence, using sparse matrices for $B$ and $A$ may be a bad solution.
Similarly, using sparse $W_\mathrm{q}$ and $W_\mathrm{k}$ may be a poor combination for attention map approximation.

\begin{figure}[t]
\centering
\includegraphics[width=0.8\linewidth]{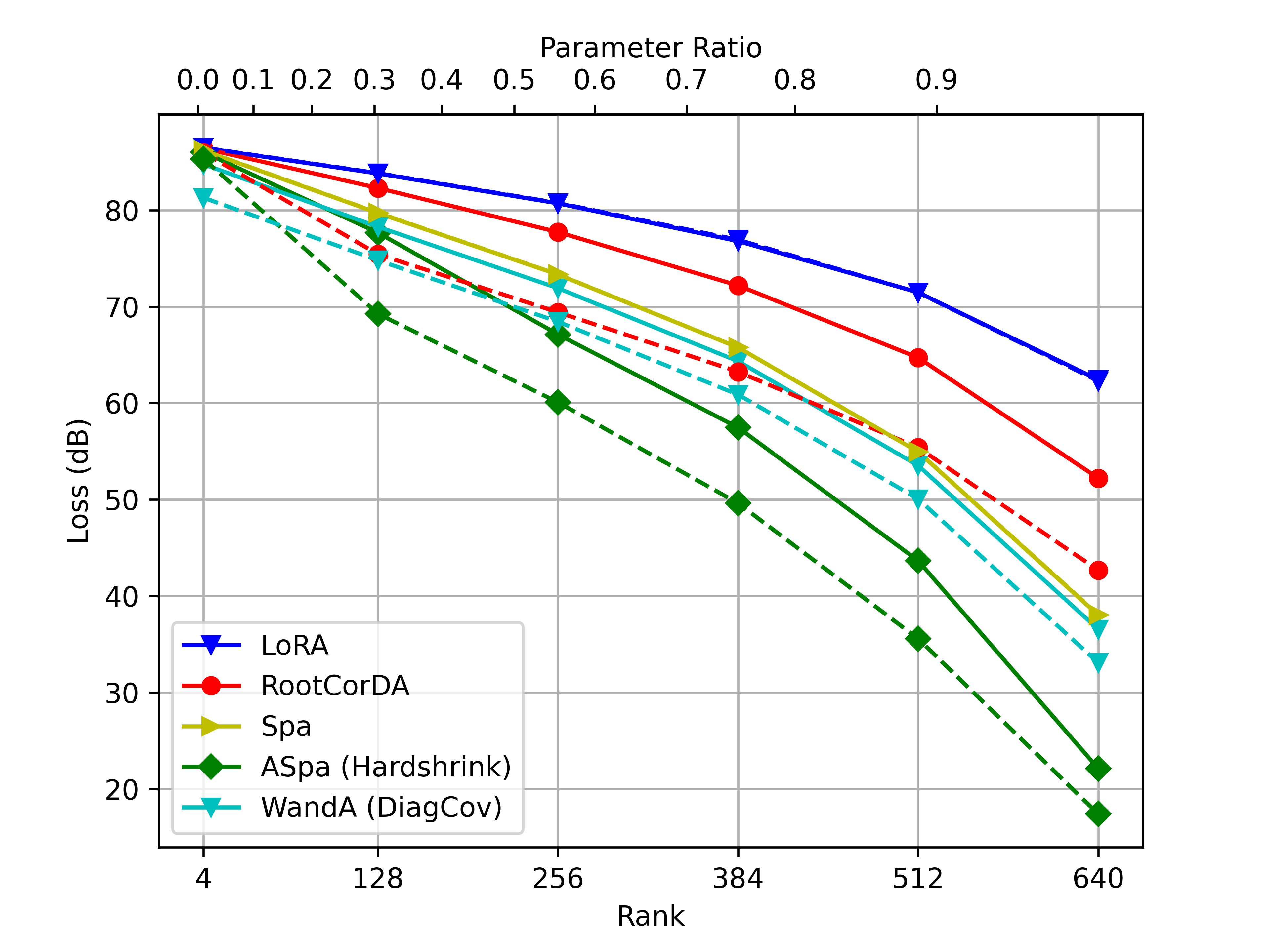}
\caption{Comparison with WandA.}
\label{fig:spa3}
\end{figure}

WandA~\cite{sun2023simple}, SparseGPT~\cite{frantar2023sparsegpt} and SparseLLM~\cite{bai2024sparsellm} use non-iterative solutions by considering only diagonal covariance:
\begin{align}
    C &\rightarrow C \odot I_d.
\end{align}
This does not require iterative compressed sensing. 
However, the diagonal approximation has a degraded performance as in Fig.~\ref{fig:spa3}.

\subsection{Quantization-Aware Distillation}
\label{sec:quant}

We can use STE for quantization-aware distillation in a straightforward manner.
Whatever the loss, we can use STE for the trainable parameters, e.g., for $B$ and $A$ low-rank matrices:
\begin{align}
    B &\leftarrow B - B.\mathrm{detach} + \mathcal{Q}[B.\mathrm{detach}],\\
    A &\leftarrow A - A.\mathrm{detach} + \mathcal{Q}[A.\mathrm{detach}],
\end{align}
where we may consider a simple chunk-wise $q$-bit uniform quantization:
\begin{align}
    x' &=
    \mathcal{Q}[x]
    \\
    &=
    \mathrm{round}\big[
    (x-x_{\min})\cdot \frac{2^q - 1}{x_{\max} - x_{\min}} 
    \big]
    \cdot 
    \frac{x_{\max} - x_{\min}}{2^q-1}
    +  x_{\min},
\end{align}
where $x_{\min}$ and $x_{\max}$ are determined from a chunk of $x$.

\section{LLM Models}
\label{sec:model}

\begin{table}[t]
\centering
\caption{OPT Models~\cite{zhang2022opt}}
\label{tab:opt}
\vskip 0.15in
\small
\begin{tabular}{rrrrrrl}
  \toprule
  Models & \# layers $L$& \# heads $h$ & hidden size $d$ & head dim $d_\mathrm{h}$ & $d_\mathrm{i}=4d$ & Huggingface ID \\
  \midrule
  125M   & 12 & 12 & 768 & 64 & 3072 & facebook/opt-125m \\
  350M   & 24 & 16 & 1024 & 64 & 4096 & facebook/opt-350m \\
  1.3B   & 24 & 32 & 2048 & 64 &8192 & facebook/opt-1.3b\\
  2.7B   & 32 & 32 & 2560 & 80 &10240 & facebook/opt-2.7b\\
  6.7B   & 32 & 32 & 4096 & 128 &16384 & facebook/opt-6.7b\\
  13B   & 40 & 40 & 5120 & 128 &20480 & facebook/opt-13b \\
  30B   & 48 & 56 & 7168 & 128&28672 & facebook/opt-30b \\
  66B   & 64 & 72 & 9216 & 128&36864 & facebook/opt-66b \\
  175B   & 96 & 96 & 12288 & 128 &49152 & \\
  \bottomrule
\end{tabular}
\vskip -0.1in
\end{table}

\begin{table}[t]
\centering
\caption{Qwen2 Models}
\label{tab:qwen}
\begin{tabular}{ccccccc}
  \hline
  Models & \# layers $L$& \# heads $h$& \# KV heads $h_\mathrm{kv}$& hidden size $d$& head dim $d_\mathrm{h}$ & $d_\mathrm{i}$\\
  \hline
  0.5B   & 24 & 14 & 2 & 896 & 64 & 4864\\
  1.5B   & 28 & 12 & 2 & 1536 & 128 & 8960 \\
  3B   & 36 & 16 & 2 & 2048 & 128 & 11008 \\
  7B   & 28 & 28 & 4 & 3584 & 128 & 18944  \\
  14B   & 40 & 40 & 4 & 5120 & 128 & 27392\\
  32B   & 60 & 56 & 8 & 7168 & 128 & 28672\\
  72B   & 32 & 32 & 32 & 4096 & 128 & 22016 \\
  \hline
\end{tabular}
\end{table}

Parameters for some major transformer models are listed in Table~\ref{tab:model}.
OPT model variants are listed in Table~\ref{tab:opt}.

For LLM models, we used LLaVa: \texttt{liuhaotian/llava-v1.6-vicuna-7b}. 
It has Vicuna-7B model for LLM and ViT based on CLIP for the vision encoder.
The Vicuna is an instruction-tuned version of LLaMa-2, having 32 transformer layers.
CLIP ViT has 24 transformer layers.

\begin{table}[t]
\small
\centering
\caption{Transformer Models}
\label{tab:model}
\begin{tabular}{cccc}
    \hline
     & ViT-16/B & Llama-2-7B & Llama-3.2-1B \\
     \hline
ID  & google/vit-base-patch16-224 & meta-llama/Llama-2-7b-hf & meta-llama/Llama-3.2-1B-Instruct \\ 
hidden size $d$ & 768 & 4096 & 2048 \\
hidden act & gelu & silu & silu \\
intermediate size $d_\mathrm{i}$& 3072 ($4d$) & 11008 ($2.68d$)& 8192 ($4d$)\\
head dim $d_\mathrm{h}=d/h$ & 64 & 128 & 64 \\
num attention heads $h$ & 12 & 32 & 32\\
num key value heads $h_\mathrm{kv}$ & 12 & 32 & 8 \\
num hidden layers $L$ & 12 & 32 & 16 \\
qkv bias & True & False & False\\
mlp bias & True & False & False \\
rope theta $\theta$& --- & 1e4 & 5e5 \\
max position embeddings & 197 & 4096 &  131072 \\
\hline
    \hline
     & OPT-350M & BLOOM-560M & Qwen2-0.5B \\
     \hline
ID  & facebook/opt-350m & bigscience/bloom-560m & Qwen/Qwen2-0.5B \\ 
hidden size & 1024 & 1024 & 896 \\
hidden act & relu & gelu & silu \\
intermediate size & 4096 & 4096 & 4864 \\
head dim & 64 & 64 & 64 \\
num attention heads & 16 & 16 & 14\\
num key value heads & 16 & 16 & 2 \\
num hidden layers & 24 & 24 & 24 \\
qkv bias & True & True & True \\
mlp bias & True & True & False \\
rope theta & --- & --- & 1e6 \\
max position embeddings & 2048 & 2048 &  131072 \\
\hline
    \hline
     & RoBERTa-350M & Phi-3.5 mini & Gemma-2B \\
     \hline
ID  & FacebookAI/roberta-base & microsoft/Phi-3.5-mini-instruct & google/gemma-2b \\ 
hidden size & 768 & 3072 & 2048 \\
hidden act & gelu & silu & gelu \\
intermediate size & 3072 ($4d$) & 8192 & 16384 \\
head dim & 64 & 96 & 256 \\
num attention heads & 12 & 32 & 8\\
num key value heads & 12 & 32 & 1 \\
num hidden layers & 12 & 32 & 18 \\
qkv bias & True & False & False \\
mlp bias & True & False & False \\
rope theta & --- & 1e4 & 1e4 \\
max position embeddings & 514 & 131072 &  8192 \\
\hline
\end{tabular}
\end{table}

\end{document}